\documentclass{article}
\usepackage{booktabs}
\usepackage{framed,multirow}
\usepackage{url}
\usepackage{hyperref}
\usepackage{amsmath}
\usepackage{bm}
\usepackage[font={footnotesize}]{caption} 		% style of figure captions
\usepackage{fp}									% high-precision arithmetic operations
\usepackage[final]{pdfpages} 							% inclusion of PDF pages (must be put BEFORE pgfplots, tikz and externalize stuff)
\usepackage{subcaption}
\usepackage{pgfplots}								% pgfplots support
\usepackage{tikz}									% tikz support
\usepgfplotslibrary{external} 							% converts each picture as a separate pdf
\pgfplotsset{compat=1.16}								% require a specific version of pgfplots					
\tikzsetexternalprefix{ext/}	
\usepackage{xcolor}									% driver-independant colors extensions
\usepackage{color} 									% foreground and background color control 

\definecolor{black}	{RGB}{0,0,0}					% black
\definecolor{white}	{RGB}{255,255,255}					% black
\definecolor{blue1}	{RGB}{0,177,234}				% deep cyan {0,177,234}
\definecolor{blue2}	{RGB}{76,200,239}					% lighter cyan 
\definecolor{blue3}	{RGB}{127,215,244}					% lighter cyan
\definecolor{blue4}	{RGB}{178,231,248}					% lighter cyan
\definecolor{bluegray1}{RGB}{0,127,167}					% deep cyan/gray mix {0,127,167}
\definecolor{bluegray2}{RGB}{76,165,193}				% lighter cyan/gray mix 
\definecolor{bluegray3}{RGB}{127,191,211}				% lighter cyan/gray mix
\definecolor{bluegray4}{RGB}{178,216,228}				% lighter cyan/gray mix
\definecolor{gray1}	{RGB}{76,84,93}					% deep gray
\definecolor{gray2}	{RGB}{129,135,141}					% lighter gray
\definecolor{gray3}	{RGB}{165,169,174}					% lighter gray
\definecolor{gray4}	{RGB}{201,203,206}					% lighter gray
\definecolor{orange1}	{RGB}{255,126,46}			% deep orange {255,126,46}
\definecolor{orange2}	{RGB}{255,164,108}				% lighter orange
\definecolor{orange3}	{RGB}{255,190,150}				% lighter orange
\definecolor{orange4}	{RGB}{255,216,192}				% lighter orange
\definecolor{purple1}{RGB}{89,89,171}
\definecolor{purple4}{RGB}{189,189,231}
\definecolor{brown}	{RGB}{205,133,63}					% brown
\definecolor{emerald}{rgb}{0.31, 0.78, 0.47}							% put externally compiled figures in ext/	
\usetikzlibrary{%
arrows,%
calc,%
fit,%
patterns,%
plotmarks,%
shapes.geometric,%
shapes.misc,%
shapes.symbols,%
shapes.arrows,%
shapes.callouts,%
backgrounds,%
chains,%
topaths,%
trees,%
petri,%
mindmap,%
matrix,%
folding,%
fadings,%
through,%
positioning,%
scopes,%
decorations.fractals,%
decorations.shapes,%
decorations.text,%
decorations.pathmorphing,%
decorations.pathreplacing,%
decorations.footprints,%
decorations.markings,%
calligraphy,
shadows,
3d}
\newcommand*{\shifttext}[2]{%
  \settowidth{\@tempdima}{#2}%
  \makebox[\@tempdima]{\hspace*{#1}#2}%
}

\usepackage{enumitem}
\usepackage{microtype}
\usepackage{graphicx}
\usepackage{subcaption}
\usepackage{booktabs} % for professional tables

\usepackage{hyperref}

\usepackage[accepted]{icml2024}

\usepackage{amsmath}
\usepackage{amssymb}
\usepackage{mathtools}
\usepackage{amsthm}

\usepackage[capitalize,noabbrev]{cleveref}

\theoremstyle{plain}
\newtheorem{theorem}{Theorem}[section]

\theoremstyle{definition}

\theoremstyle{remark}
\newtheorem{remark}[theorem]{Remark}

\usepackage[textsize=tiny]{todonotes}

\allowdisplaybreaks

\icmltitlerunning{PiT: Position-induced Transformer for Operator Learning}

\begin{document}

\twocolumn[
\icmltitle{Positional Knowledge is All You Need:\\ Position-induced Transformer (PiT) for Operator Learning}

%PiT: An Efficient Position-induced Transformer for \\
%Operator Learning in Partial Differential Equations

% It is OKAY to include author information, even for blind
% submissions: the style file will automatically remove it for you
% unless you've provided the [accepted] option to the icml2024
% package.

% List of affiliations: The first argument should be a (short)
% identifier you will use later to specify author affiliations
% Academic affiliations should list Department, University, City, Region, Country
% Industry affiliations should list Company, City, Region, Country

% You can specify symbols, otherwise they are numbered in order.
% Ideally, you should not use this facility. Affiliations will be numbered
% in order of appearance and this is the preferred way.
\begin{icmlauthorlist}
\icmlauthor{Junfeng Chen}{xxx}
\icmlauthor{Kailiang Wu}{xxx,yyy,zzz}
\end{icmlauthorlist}

\icmlaffiliation{xxx}{Department of Mathematics, Southern University of Science and Technology, Shenzhen 518055, China}
\icmlaffiliation{yyy}{Shenzhen International Center for Mathematics, Southern University of Science and Technology, Shenzhen 518055, China}
\icmlaffiliation{zzz}{National Center for Applied Mathematics Shenzhen (NCAMS), Shenzhen 518055, China}
\icmlcorrespondingauthor{Kailiang Wu}{wukl@sustech.edu.cn}

% You may provide any keywords that you
% find helpful for describing your paper; these are used to populate
% the "keywords" metadata in the PDF but will not be shown in the document
\icmlkeywords{Machine Learning, ICML}

\vskip 0.3in
]

% this must go after the closing bracket ] following \twocolumn[ ...

% This command actually creates the footnote in the first column
% listing the affiliations and the copyright notice.
% The command takes one argument, which is text to display at the start of the footnote.
% The \icmlEqualContribution command is standard text for equal contribution.
% Remove it (just {}) if you do not need this facility.

%\printAffiliationsAndNotice{}  % leave blank if no need to mention equal contribution
\printAffiliationsAndNotice{} % otherwise use the standard text.

\begin{abstract}

Operator learning for Partial Differential Equations (PDEs) is rapidly emerging as a promising approach for surrogate modeling of intricate systems. Transformers with the self-attention mechanism---a powerful tool originally designed for natural language processing---have recently been adapted for operator learning. However, they confront challenges, including high computational demands and limited interpretability. This raises a critical question: {\em Is there a more efficient attention mechanism for Transformer-based operator learning?} This paper proposes the Position-induced Transformer (PiT), built on an innovative position-attention mechanism, which demonstrates significant advantages over the classical self-attention in operator learning. Position-attention draws inspiration from numerical methods for PDEs. 
Different from self-attention, position-attention is induced by only the spatial interrelations of sampling positions for input functions of the operators, and 
 does not rely on the input function values themselves, thereby greatly boosting efficiency. PiT  exhibits superior performance over current state-of-the-art neural operators in a variety of complex operator learning tasks across diverse PDE benchmarks. Additionally, PiT possesses an enhanced discretization convergence feature, compared to the widely-used Fourier neural operator.

\end{abstract}
\section{Introduction}

Partial Differential Equations (PDEs) are essential in modeling a vast array of phenomena across various fields including physics, engineering, biology, and finance. They are the foundation for predicting and understanding many complex dynamics in natural and engineered systems. 
Over the past century, traditional numerical methods, such as finite element, finite difference, and spectral methods, have been well established for solving many PDEs. However, 
these methods often face challenges for complex nonlinear systems with complicated geometries or in high dimensions.  

The advent of machine learning has shifted the paradigm in addressing these challenges in PDEs. 
Key developments include but are not limited to PINN \cite{raissi2019physics}, Deep-Galerkin \cite{sirignano2018dgm}, and 
Deep-Ritz \cite{yu2018deep}. 
These methods are typically solution-learners, i.e., learning a solution of PDEs. 
 They closely resemble traditional 
 approaches such as finite elements, 
 replacing local basis functions 
 with neural networks. 
 While advantageous for high-dimensional problems and complex geometries, solution-learners are usually limited to a single instance, i.e., solving one  solution of the PDEs with a fixed initial/boundary condition. To get solution for every new condition, retraining a new neural network is required, which can be very costly.

Solution-learners usually focus on a given PDE. However, for many complex systems  
the governing equations remain unclear due to uncertain mechanisms, yet identifying underlying
PDEs is very challenging without
sufficient domain knowledge. Data-driven approaches have emerged as a powerful tool for discovering unknown PDEs or surrogate modeling of known yet complex PDEs. 
 Early strategies include sparse-promoting regression \cite{rudy2017data,schaeffer2017learning}; however, even after learning the underlying PDEs, numerical solving is still necessary to obtain their solutions.
Other techniques, such as 
PDE-Net \cite{long2018pde,long2019pde}, can learn both the underlying PDE and its solution dynamics.

Another data-driven approach is Flow Map Learning (FML) \cite{qin2019data,wu2020data,chen2022deep,churchill2023flow}.  
Unlike solution-learners, FML is an operator-learner, which approximates the evolution operator of time-dependent systems with varying initial conditions. 
Once learned, the flow map or evolution operator can be recursively utilized to predict long-term solution behaviors of the equations for any new initial condition without the need for retraining.

Recently, more versatile operator-learners have been systematically developed for learning mappings between infinite-dimensional function spaces.  
Existing frameworks include, but are not limited to, the neural operators  \cite{anandkumar2019neural,li2020multipole,li2021fourier,kovachki2023neural}, DeepONets  \cite{lu2021learning,jin2022mionet,lanthaler2023nonlinear}, principal component analysis-based methods \cite{bhattacharya2021model}, and attention-based methods \cite{cao2021choose,kissas2022learning,hao2023gnot}, etc. These operator-learners are applicable to learn the solution operators of parametric PDEs, including mappings from initial conditions to solutions at specific future times, or from boundary conditions, source terms, and model parameters to steady-state solutions.

Transformers, a powerful tool initially designed for natural language processing \cite{vaswani2017attention}, have also been adapted for learning operators in PDEs, e.g., \citet{liu2022ht,hao2023gnot,xiao2023improved}. The core of Transformers is the self-attention mechanism.  However, conventional self-attention lacks positional awareness, which is found crucial in natural language processing \cite{vaswani2017attention,shaw2018self} and graph representation \cite{dwivedi2020generalization}, thus sparking significant research interest \cite{dai2019transformer,dufter2022position}. 
In PDE operator learning, there exist few studies on integrating positional knowledge with self-attention. \citet{cao2021choose,lee2022mesh} concatenate the coordinates of sampling points with input function values, while \citet{li2023transformer} adopt the rotary position embedding technique \cite{su2024roformer} to enhance self-attention. 
Self-attention in operator learning is content-based and relies heavily on the values of input functions. This necessitates distinct attention calculations for each training batch instance, resulting in significant memory usage and high computational costs, especially when compared to the neural operators in \citet{kovachki2023neural}. This raises critical questions: {\em Is self-attention indispensable for Transformer-based operator learning? What key positional information is necessary, and how can it be efficiently encoded into Transformer-based neural operators?}

To overcome the challenges of self-attention in operator learning, we propose a novel attention mechanism, termed  \emph{position-attention}, from a numerical mathematics perspective. 
This mechanism is induced by only spatial relations without relying on input function values, marking a significant difference from classical content-based self-attention. Position-attention greatly enhances computational efficiency and effectively integrates positional knowledge. It also resonates with the principles of numerically solving PDEs, offering an interpretable approach to operator learning. 
Building upon position-attention and its variants, we develop a novel deep learning architecture, termed \emph{Position-induced Transformer} (PiT), for operator learning. Compared to current state-of-the-art neural operators, PiT exhibits superior performance across various benchmarks from elliptic to hyperbolic PDEs, even in challenging cases where the solutions contain discontinuities. Like many neural operators \cite{azizzadenesheli2024neural}, PiT features a remarkable discretization convergence property (also called discretization/mesh invariance in the literature \cite{kovachki2023neural,li2021fourier}), enabling effective generalization to new meshes which are unseen during training. 

The main contributions of this work include: 
\begin{itemize}[leftmargin=*]
	\item We find the importance of positional knowledge, specifically the spatial interrelations of the nodal points where the input functions are sampled, in operator learning. We propose the novel  position-attention mechanism and its two  variants to effectively incorporate such positional knowledge. Compared to self-attention, position-attention is interpretable from a numerical mathematics perspective and is more efficient for operator learning.
	\item  Based on position-attention and its two variants, we construct PiT, a lightweight Transformer whose training time scales only sub-linearly with the sampling mesh resolution of input/output functions. Moreover, PiT is discretization-convergent, offering consistent and convergent predictions as the testing meshes are refined.
	\item We conduct numerical experiments on various PDE benchmarks, showcasing the remarkable performance of PiT, and demonstrate its greater robustness in discretization convergence (with 48\% smaller prediction error for the Darcy2D benchmark)  compared to the Fourier neural operator (FNO). {Our code is accessible at \url{github.com/junfeng-chen/position_induced_transformer}.}
\end{itemize}

\section{Approach}
\subsection{Preliminaries}
\textbf{Operator Learning}.  
Consider generic parametric PDEs: 
\begin{equation}
    \label{eq:gpde}
    \mathcal{L}_au=f, 
\end{equation}
where $a\in\mathcal{A}(\Omega_a;\mathbb{R}^{d_a}), u\in\mathcal{U}(\Omega_u;\mathbb{R}^{d_u})$ are functions defined on the bounded domains $\Omega_a$ and $\Omega_u$, respectively; $\mathcal{L}_a: \mathcal{U} \rightarrow {\mathcal F}$ is a partial differential operator; $f\in {\mathcal F}$;  
$\mathcal{A}$ and $\mathcal{U}$ are Banach spaces of functions over $\Omega_a$ and $\Omega_u$, respectively. 
As in \citet{li2021fourier,anandkumar2019neural,li2020multipole,kovachki2023neural}, we assume $\Omega_a=\Omega_u=\Omega\subset\mathbb{R}^{d}$ in this paper. 
Denote the operator that maps $a$ to $u$ by $\Phi$, which can be the evolution operator that maps the initial condition to the solution at a specific future time, or the solution operator that maps the source term or model parameters to the steady-state solution. 
Operator learning aims to construct a neural operator  $\Phi_{\theta}$, as surrogate  model of $\Phi$, from sampling data pairs $\{a^j,u^j\}_{j=1}^J$. 
The data are usually sampled on two (possibly different) meshes  $X_a=\{x_i\}_{i=1}^{N_a}\subset\Omega$ and $X_u=\{\hat x_i\}_{i=1}^{N_u}\subset\Omega$:
$$a^j = \left\{a^j(x_i)\right\}_{i=1}^{N_a}, ~~~ u^j = \left\{u^j(\hat x_i)\right\}_{i=1}^{N_u}, ~~~ j=1,2,...,J.$$
Assume that the input data $\{a^j\}_{j=1}^J$ are drawn from a probability measure  $\mu_{\mathcal{A}}$ supported on $\mathcal{A}$, and the sampling points $X_a$ are i.i.d.~drawn from a measure $\mu_\Omega$ on $\Omega$, denoted as $X_a\sim\mu_{\Omega}$. 
After training the parameters $\theta$ on the data pairs $\{a^j,u^j\}_{j=1}^J$, we expect that the trained neural operator $\Phi_{\bm \theta}$ exhibits small generalization error defined as
\begin{equation}
    \label{eq:no}
    \mathbb{E}_{a\sim\mu_{\mathcal{A}}}\left(\|u-\Phi_{\bm \theta}(a|_{X})\|^2_{\mathcal{U}}\right), \quad\forall X\sim\mu_{\Omega},
\end{equation}
where the norm $\|\cdot\|_{\mathcal{U}}$ is in practice replaced with a vector norm of the output function values queried on a new mesh $X_{\tt new}$. The formulation \eqref{eq:no} indicates that the learned operator $\Phi_{\bm \theta}$ can accept any mesh points in the domain of $a$ and predict $u$ at any queried mesh $X_{\tt new}$.

\begin{figure}[!t]
	\setlength{\fboxsep}{0pt}%
	\setlength{\fboxrule}{0.2pt}
	\captionsetup[subfigure]{labelformat=empty}
	\centering
	\begin{subfigure}[t]{.22\textwidth}
		\centering
		\fbox{\includegraphics[width=0.7\linewidth]{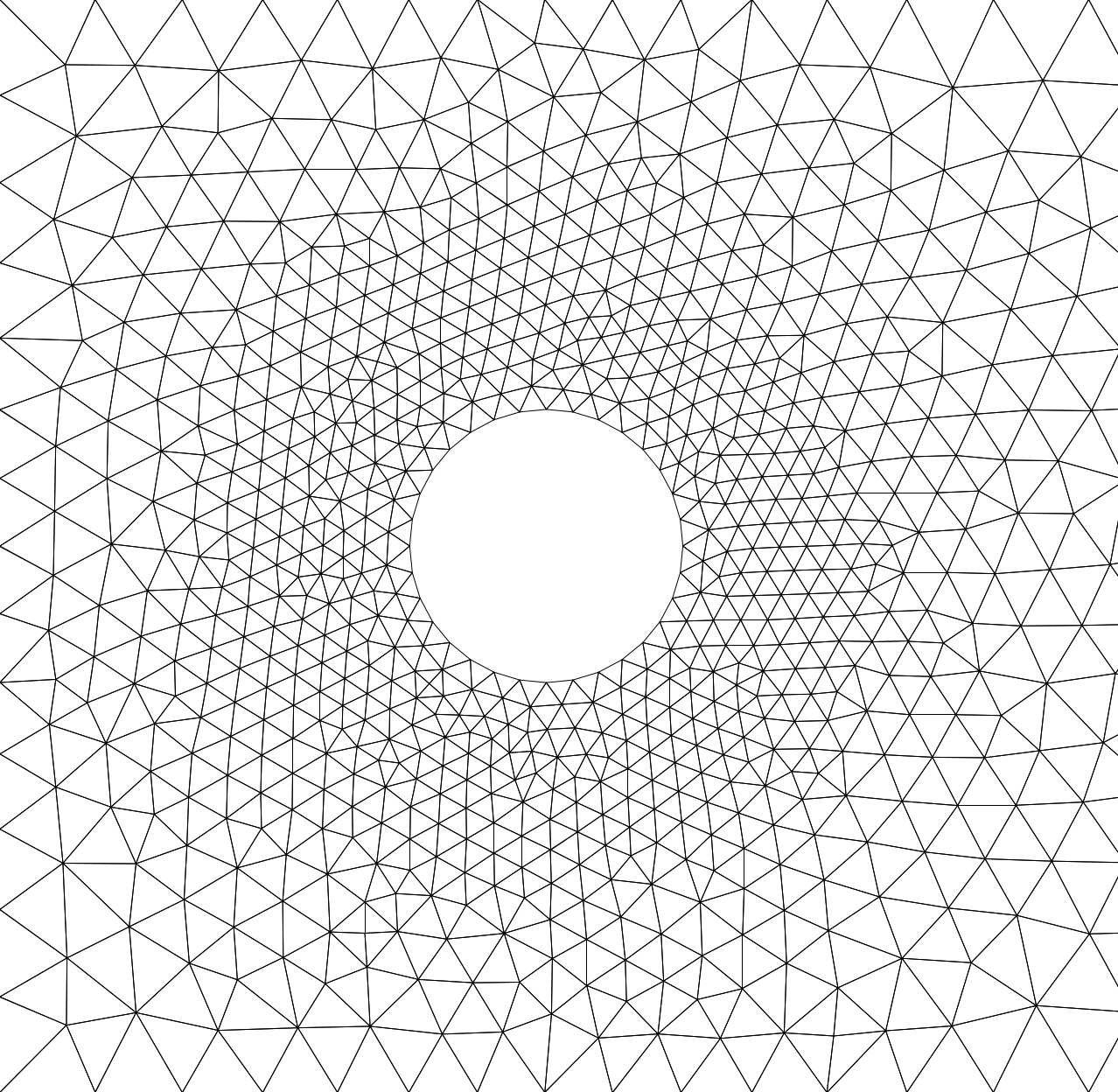}}
		\caption{Training mesh resolution}
	\end{subfigure}%
	\begin{subfigure}[t]{.22\textwidth}
		\centering
		\fbox{\includegraphics[width=0.7\linewidth]{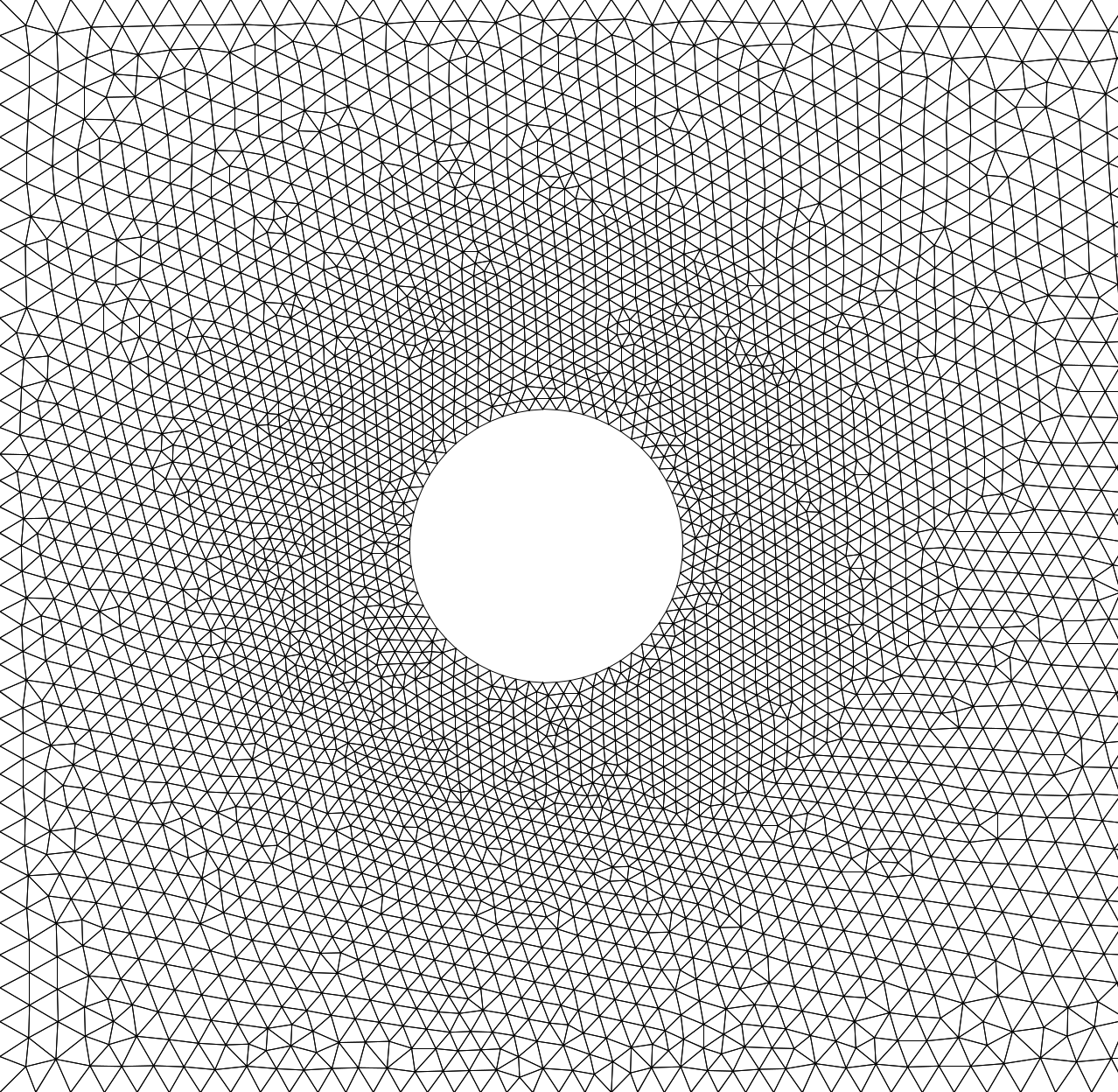}}
		\caption{Testing mesh resolution}
	\end{subfigure}
	\caption{Discretization convergence test for neural operators. }
	\label{fig:mesh_invariance}
\end{figure}

It is often desirable to achieve a reliable neural operator that is trained with merely inexpensive data on a coarse mesh and yet generalizes well to finer meshes without the need for retraining, as illustrated in Figure \ref{fig:mesh_invariance}.
In particular, one expects consistent and convergent predictions as the testing meshes are refined \cite{azizzadenesheli2024neural}.
A common method for assessing this is the \emph{zero-shot super-resolution} evaluation.

A neural operator typically adopts an Encoder-Processor-Decoder architecture:
$$\Phi_{\theta}=\text{Decoder}\circ \text{LAYER}_L\circ\cdots\circ \text{LAYER}_1\circ \text{Encoder},$$
where the Encoder lifts the input function from $\mathbb{R}^{d_a}$ to a higher-dimensional feature space $\mathbb{R}^{d_1}$, and the Decoder projects the hidden features from $\mathbb{R}^{d_L}$ to $\mathbb{R}^{d_u}$.
The Encoder and Decoder are usually implemented by linear layers, and can include nonlinearities if necessary. 
In the Processor, let the function $v_\ell:\Omega\rightarrow\mathbb{R}^{d_\ell}$ be the continuum of the feature map $U_\ell$ of $\text{LAYER}_\ell$. Then, the forward pass $U_{\ell+1}=\text{LAYER}_{\ell+1}(U_\ell)$ approximates the transform
\begin{equation*}
    %\label{eq:integral_operator}
    \resizebox{1\hsize}{!}{ $ \displaystyle v_{\ell+1}(x)=
         \sigma\left(\int_{\Omega}\kappa_\ell(x,y,v_\ell(x),v_\ell(y))\,v_\ell(y)\,dy+ v_\ell(x) W_\ell \right),$}
    %\end{aligned}
\end{equation*}
where the integral kernel $\kappa_\ell$ needs to be parametrized and trained, $W_\ell\in\mathbb{R}^{d_\ell \times d_{\ell+1} }$ is a trainable matrix, and $\sigma$ is a nonlinear activation function. Various parametrization methods for $\kappa_\ell$ have been explored, including but not limited to message passing on graphs \cite{anandkumar2019neural,li2020multipole}, Fourier transform \cite{li2021fourier}, and multiwavelet transform \cite{gupta2021multiwavelet}.

\textbf{Transformer and Self-attention}.
Transformers, proposed by \citet{vaswani2017attention}, are fundamental in natural language processing and form the basis of major advanced language models including GPT and BERT. Their essence lies in the self-attention mechanism.  
Recently, \citet{kovachki2023neural} observed the connections between attention 
and neural operators, highlighting the potential of Transformers in operator learning. 
Various specialized and effective Transformers have been developed for operator learning, using Galerkin-type attention \cite{cao2021choose}, hierarchical attention \cite{liu2022ht}, cross-attention \cite{lee2022mesh,li2023transformer}, mixture of experts \cite{hao2023gnot}, and orthogonal regularization \cite{xiao2023improved}.

Consider $U\in\mathbb{R}^{N_v\times d_\ell }$ as the input sequence comprising $N_v$ elements, each represented by a $d_\ell$-dimensional feature vector.  Self-attention can be expressed as 
\begin{equation}
	\label{eq:selfatt}
	\text{SelfAtt}(U)=\text{Softmax}\left(\frac{UW^Q(UW^K)^T}{\sqrt{d_{\ell+1}}}\right)UW^V,
\end{equation}
where $W^Q, W^K, W^V\in \mathbb{R}^{d_\ell \times d_{\ell+1}}$ are trainable matrices. 
Self-attention is content-based, and it heavily depends on input function values in operator learning. 
This demands separate attention computations for each training batch instance, 
leading to intensive memory and computational costs, 
compared to the neural operators in \citet{kovachki2023neural}.

%%%%%%%%%%%%%%%%%%%%%%%%%%%%%%%%%%%%%%%%%%%%%%%%
\subsection{Novel Position-attention and Its Variants}
\label{sec:posatt}
We find the positional knowledge, specifically the spatial interrelations of the sampling points, is essential for operator learning. We propose the \emph{position-attention} mechanism and its two variants, which effectively incorporate such positional knowledge. 
In contrast to classical self-attention, 
position-attention does not rely on the input function values themselves, thereby greatly boosting efficiency. 
Furthermore, position-attention is consistent with the changes of mesh resolution and converges as the meshes are refined. 
%%%%%%%%%%%%%%%%%%%%% 
\begin{figure*}[htbp]
    \centering
    \begin{subfigure}[c]{.3\textwidth}
	\centering
	\hspace{10mm}\includegraphics[width=.99\linewidth]{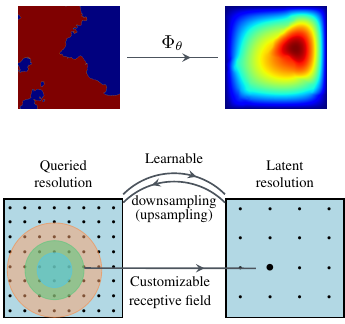}
\end{subfigure}
\begin{subfigure}[c]{.6\textwidth}
	\centering
	\includegraphics[width=.9\linewidth]{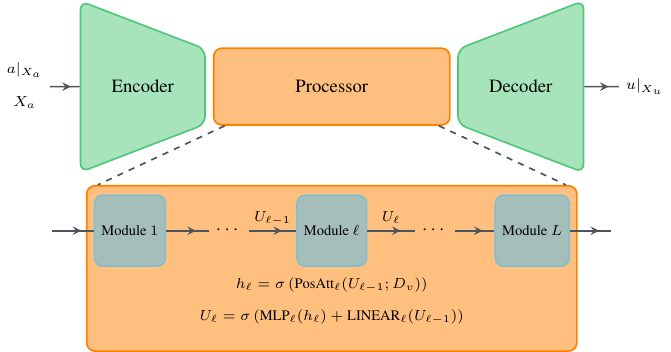}
\end{subfigure}
    \caption{Overview of Position-induced Transformer for operator learning. Top left: A trained neural operator can serve as a surrogate model to specific parametric PDEs. Bottom left: Cross position-attention provides learnable downsampling/unsampling between meshes at different resolutions, and local position-attention supports customizable receptive field. Right: The Encoder-Processor-Decoder architecture of PiT.}
    \label{fig:architecture}
\end{figure*}

\textbf{Position-attention}. 
Let $U\in\mathbb{R}^{N_v\times d_\ell}$ be the values of a generic function $v$ sampled on a mesh $X_v$ of $N_v$ nodal points. Define the pairwise-distance matrix $D\in\mathbb{R}^{N_v\times N_v}$:
\begin{equation}
    \label{eq:pairwise}
    \begin{aligned}
        & D_{ij} = \|x_i-x_j\|_2^2.\\
    \end{aligned}
\end{equation}
Let $\lambda>0$ and {$W^V\in\mathbb{R}^{d_{\ell}\times d_{\ell +1}}$} be trainable parameters. The position-attention mechanism is defined by
\begin{equation}
    \label{eq:PosAtt}
    \text{PosAtt}(U;D) :=\text{Softmax}(-\lambda D)UW^V, 
\end{equation}
which is linear with respect to $U$. Here, $\text{Softmax}(-\lambda D)U$ can be understood as a global linear convolution, with the kernel weights adjusted according to the relative distances between sampling points. This design is motivated by the concept of \emph{domain of dependence} in PDEs, reflecting how the solution at a point is influenced by the local neighboring information. Specifically, the $i$th row of the output is
\begin{equation}
    \label{eq:PosAtt2}
    \text{PosAtt}(U;D)_i = \sum_{k=1}^{N_v}\frac{\exp(-\lambda D_{ik})}{\sum_{j=1}^{N_v}\exp(-\lambda D_{ij})}(UW^V)_k.
\end{equation}
\begin{theorem}
    \label{prop:pa}
	Let $\{X_n\}_{n=1}^{+\infty}$ be a sequence of refined meshes on $\Omega$ with $X_n\sim\mu_{\Omega}$. Denote by $D^n$ the pairwise-distance matrix \eqref{eq:pairwise} corresponding to $X_n$.  Assume that $v(x)$ is bounded on $\Omega$, and denote by $U^n$ the function values of $v$ on $X_n$. As $n\to +\infty$, the position-attention \eqref{eq:PosAtt} converges to an integral operator: specifically, for any $\varepsilon > 0$, 
	\begin{equation*}
	\lim_{n\to+\infty}  {\rm Pr} \left\{ 	\frac{1}{|X_n|} \Big\| \text{PosAtt}(U^n;D^n) - {\mathcal F}   |_{X_n} \Big\| \le \varepsilon \right\} = 1, 
	\end{equation*}
	where $|X_n|$ denotes the number of nodal points in $X_n$, 
	\begin{equation}\label{eq:2}
	   {\mathcal F}(x) :=  \int_{\Omega}\kappa^{\lambda}\left(x-y\right)v(y)W^V\,d\mu_{\Omega}(y), 
	\end{equation}
and  
\begin{equation}
    \label{eq:PosAtt_kernel}
    \kappa^{\lambda}\left(x-y\right) = \frac{\exp(-\lambda\|x-y\|_2^2)}{\int_{\Omega}\exp(-\lambda\|x-y'\|_2^2)\,d\mu_{\Omega}(y')}
\end{equation}
is the integral kernel induced by position-attention. 
\end{theorem}

The proof of Theorem \ref{prop:pa} is put in Appendix \ref{app:proof}.
The fixed measure $\mu_{\Omega}$ and the row-wise Softmax normalization play crucial roles in the convergence, which implies that position-attention is discretization-convergent, eliminating the need for nested mesh refinement as required in \citet{kovachki2023neural}. If one replaces the Softmax normalization with element-wise exponentiation, then $\kappa^{\lambda}$ becomes a Gaussian kernel, which is, however, sensitive to mesh resolution.

The kernel (\ref{eq:PosAtt_kernel}) induced by position-attention is independent of the input functions. This is a notable difference from classical self-attention, which is content-based and heavily relies on the input function values themselves. Indeed, position-attention draws inspiration from numerical schemes solving PDEs. For instance, consider the upwind scheme for the advection equation $v_t+ s v_x=0$ with a constant speed $s$:  
\begin{equation*}
	\resizebox{1\hsize}{!}{$
	\begin{aligned}
	v_j^{n+1} &= v_j^n-\frac{c}{2} ( v_{j+1}^n-v_{j-1}^n )+\frac{|c|}2 ( v_{j+1}^n-2 v_{j}^n + v_{j-1}^n  )
	\\ &=: H_{c}( v_{j-1}^n, v_{j}^n,  v_{j+1}^n ),
\end{aligned}$}
\end{equation*}
where $v_j^n$ is the numerical solution at the $j$th grid point and time $t_n$. Here, the operator $H_c$ is discretization-convergent, depending only on a fixed Courant--Friedrichs--Lewy (CFL) number $c:=s\Delta t/\Delta x$, and is independent of the input function values $\{v_{j-1}^n,v_j^n,v_{j+1}^n\}$. 
This scheme can be interpreted as a local linear convolution. Position-attention shares a similar concept but employs a global linear convolution, with the kernel reflecting a stronger dependence on local neighboring regions. Indeed, the value of $\lambda$ in position-attention is interpretable, as most attention at a queried point $x$ is directed towards points $y$ with the distance to $x$ smaller than $1/\sqrt{\lambda}$; see Appendix \ref{section:interpretabilit} for detailed discussions.

\textbf{Cross Position-attention}. 
We further propose a novel interpretable variant, {\em cross position-attention}, for 
downsampling/upsampling unstructured data. It interpolates $U$ from a mesh $X_1$ onto another mesh $X_2$ by
\begin{equation}
    \label{eq:cross_PosAtt}
    \text{CroPosAtt}(U;D) :=
    \text{Softmax}(-\lambda D_{1\rightarrow 2}^T)UW^V,
\end{equation}
where $D_{1\rightarrow 2}$ is the pairwise-distance matrix between $X_1$ and $X_2$. As $X_1$ is refined, cross position-attention also approximates the integral operator defined in equation (\ref{eq:2}). This property allows us to construct a discretization-convergent Encoder that downsamples the input function values on any mesh $X_a$ to a pre-fixed coarser \emph{latent mesh} $X_v$, on which the Processor is inexpensive. Analogously, a discretization-convergent Decoder can be constructed to upsample the processed features onto any output mesh $X_u$.
\begin{remark}
    While $X_a,X_v \sim \mu_\Omega$ are important for discretization convergence, we do not require any structure in the output mesh $X_u$. The output function values can be queried at any point in the domain $\Omega$. The whole model architecture and computational complexity will be detailed in Section \ref{section:topoformer}. 
\end{remark}

\textbf{Local Position-attention}. 

The position-attention mechanism naturally captures global dependencies. However, local patterns are often crucial in the solutions of various PDEs, especially those of hyperbolic nature or dominated by convection. To address this, we introduce a local variant of position-attention:
\begin{equation}
    \label{eq:sparse_PosAtt}
    \resizebox{1\hsize}{!}{$ \displaystyle 
    \text{LocPosAtt}(U;D)_i=\sum_{D_{ik}\leq r_i^2}\frac{\exp(-\lambda D_{ik})}{\sum\limits_{D_{ij}\leq r_i^2}\exp(-\lambda D_{ij})}(UW^V)_k.$}\\
\end{equation}
Similar to Theorem \ref{prop:pa}, as the mesh is refined, local position-attention approximates the integral operator
$$\int_{B_{r_i}(x_i)}\kappa_{r_{x_i}}^{\lambda}\left(x_i-y\right)u(y)W^V\,\mu_{\Omega}(dy)$$
with the induced compact kernel
\begin{equation}
    \label{eq:sparse_PosAtt_kernel}
    k_{r_x}^{\lambda}\left(x-y\right)=\frac{\exp(-\lambda\|x-y\|_2^2)}{\int_{B_{r_x}(x)}\exp(-\lambda\|x-y'\|_2^2)\,d\mu_{\Omega}(y')},
\end{equation}
where $B_{r_x}(x)$, usually termed receptive field, is a ball with radius $r_x$ and center $x$. We take the radius $r_x$ as a quantile of the row values in the pairwise-distance matrix, adapting the receptive field to the local density of nodal points. This design enables local position-attention to effectively handle functions exhibiting multiscale features. The value of quantile is left as a hyperparameter; see Section \ref{section:meta_parameter}. 
\subsection{Position-induced Transformer}
\label{section:topoformer}

We now design our novel Transformer architecture, PiT, which is primarily composed of the proposed global, cross, and local position-attention mechanisms for mixing features over the domain $\Omega$. A sketch of the PiT architecture is depicted in Figure \ref{fig:architecture}.

\textbf{Encoder}. 
The Encoder comprises lifting and downsampling operations using both local and cross position-attention mechanisms:
$$\text{Encoder} = \sigma\circ\text{LocPosAtt}_{\text{in}}(\;\cdot\;;D_{a\rightarrow v}^T)\circ\sigma\circ\text{LINEAR},$$
where $D_{a\rightarrow v}$ is the pairwise-distance matrix between the input and latent meshes, $X_a$ and $X_v$; LINEAR refers to a fully connected layer applied row-wisely to the feature matrix. This design allows us to embed the inputs on a coarse mesh into a higher-dimensional feature space, while the local position-attention effectively extracts the local features of the inputs. 
The dimension $d_v$ of the lifted features is termed the \emph{encoding dimension}, which is an important hyperparameter for the model's expressive capacity. 

\textbf{Processor}. 
The Processor consists of a sequence of global position-attention modules. To address the nonlinearity in general operators, we propose the following module as the building block to construct the Processor:
\begin{equation*}
    \label{eq:PosAtt_module}
    \begin{aligned}
        & h_\ell = \sigma\left(\text{PosAtt}_\ell(U_{\ell-1};D_v)\right),\\
        & U_{\ell} = \sigma\left(\text{MLP}_\ell(h)+\text{LINEAR}_\ell(U_{\ell-1})\right),
    \end{aligned}
\end{equation*}
where $D_v$ is the pairwise-distance matrix of $X_v$; $U_0$ is the output of Encoder; MLP$_\ell$ refers to a multilayer perceptron applied row-wisely to $h_\ell$. 
Throughout our experiments, 
we stack  four global attention modules ($L=4$) in the Processor with two layers in MLP, and take $d_\ell = d_v$ for all $1\le \ell \le L$.

\textbf{Decoder}. 
In the Decoder, we firstly upsample the feature map $U_L$ from the Processor to the queried nodal points $X_u$, and then apply an MLP row-wisely to project the features back to the range space of the output functions.
$$\text{Decoder} = \text{MLP}\circ\sigma\circ\text{LocPosAtt}_{\text{out}}(\;\cdot\;;D_{u\rightarrow v}^T).$$

\textbf{High Efficiency: 
	Linear Computational Complexity}. 
Due to the kernel matrix multiplication, the forward computation of global position-attention has a quadratic complexity of $O(N_v^2)$. To accelerate large-scale operator learning tasks, we adopt a downsampling-processing-upsampling network architecture. Denote the numbers of nodal points in the meshes $X_a$, $X_v$, and $X_u$ by $N_a$, $N_v$, and $N_u$, respectively. We take $N_v$ relatively small for efficiency. 
The computational complexities in the Encoder and Decoder are $O(N_aN_vd_v+N_ad_v^2)$ and $O(N_uN_vd_v+N_ud_v^2)$, respectively, which are both linear to the numbers of input and output mesh points. This is confirmed by our experiments, where we observe that the training time of PiT scales only sub-linearly with $N_a$ and $N_u$ (see Appendix \ref{section:sublinearly}).

We treat $N_v$ as a hyperparameter of PiT for balancing computational efficiency and information retention on the coarse latent mesh; see Section \ref{section:meta_parameter}. For training data on structured meshes, we obtain a coarser latent mesh via pooling. For unstructured data, if the distribution is known, one can generate an appropriate latent mesh by sampling; if unknown, one can use farthest point sampling \cite{Zhou2018} to preserve spatial distribution in the latent mesh.

\section{Related Work}
\textbf{Operator Learning}. 
This area is actively researched with numerous related contributions.   
\citet{chen1995universal} established a universal approximation theorem for approximating nonlinear operators using neural networks. 
Motivated by this theorem, the DeepONet framework 
\cite{lu2021learning}, comprising a trunk-net and a branch-net, was proposed for operator learning. 
The branch-net inputs discretized function values, while the trunk-net
inputs coordinates in the domain of the output function.
They combine to predict the output function values at specified coordinates.
This framework has motivated various extensions, e.g., \citet{jin2022mionet}, \citet{seidman2022nomad}, \citet{lanthaler2023nonlinear}, \citet{lee2023hyperdeeponet}, and \citet{patel2024variationally}, etc.

Another pioneering framework is the neural operators based on
iterative kernel integration \cite{anandkumar2019neural,li2021fourier,kovachki2023neural}. 
Unlike the trunk-branch architecture in DeepONet, this framework typically relies on composing linear integral operators with nonlinear activation functions. 
The graph neural operator \cite{anandkumar2019neural} leverages message passing to approximate linear kernels in the form $\kappa(x,y)$. FNO \cite{li2021fourier} is related to a shift-invariant kernel $\kappa(x-y)$, facilitating operator learning in a frequency domain via discrete Fourier transform. 
This renders FNO efficient for problems with periodic features. 
As FNO is limited to uniformly distributed data,  various new variants have emerged to handle more complex data structures and geometries, e.g, Geo-FNO \cite{li2022fourier}, the non-equispaced Fourier solver \cite{lin2022non}, and the Vandermonde neural operator \cite{lingsch2023vandermonde}. 
Researchers have also developed other related approaches for learning operators in frequency or modal spaces with generalized Fourier projections \cite{wu2020data}, multi-wavelet basis \cite{gupta2021multiwavelet}, and Laplacian eigenfunctions \cite{chen2023laplace}.

\textbf{Transformer-based Neural Operators}. 
Recently, Transformers have been extended to operator learning, including Galerkin and Fourier Transformers \cite{cao2021choose}, HT-net \cite{liu2022ht}, MINO \cite{lee2022mesh}, OFormer \cite{li2023transformer}, GNOT \cite{hao2023gnot}, and ONO \cite{xiao2023improved}. These Transformers are built on self-attention, which relies on the input function values to compute attention weights. This results in distinct attention calculations for each training batch instance, making the Transformer-based neural operators computationally expensive. In contrast, position-attention only rely on the pre-defined pairwise-distance matrix of the sampling points and does not depend on the input function values. This new mechanism notably reduces memory usage and accelerates training.

Cross position-attention, which downsamples the input functions onto coarse latent meshes, also contributes to the high efficiency of PiT. Related ideas include content-based cross-attention used in MINO \cite{lee2022mesh}, and bilinear interpolation employed by Galerkin Transformer \cite{cao2021choose}. PiT combines the advantages of both, simultaneously possessing the applicability to irregular point clouds, similar to MINO, and the interpretability, akin to the interpolation in Galerkin Transformer. There are also many other efforts reducing the computational costs of Transformers, such as random feature approximation \cite{choromanski2020rethinking,peng2021random}, low-rank approximation \cite{lu2021soft,xiong2021nystromformer}, Softmax-free normalization \cite{cao2021choose}, linear cross-attention \cite{li2023transformer,hao2023gnot}, etc. These techniques may potentially be combined with position-attention to further enhance its efficiency.

\textbf{Positional/Structural Encoding in Transformers}.
Transformers, following their success in large language models, have found broad applications in fields such as imaging \cite{dosovitskiy2020image} and graph modeling \cite{velivckovic2018graph,yun2019graph}. In these models, self-attention is content-based and requires positional encoding. Position information typically falls into two categories: absolute and relative.  \citet{vaswani2017attention} used sinusoidal functions to encode the absolute positions of words in a sentence. In contrast, \citet{yang2018modeling} and \citet{guo2019gaussian} focused on the localness of text by adjusting self-attention scores based on word distances. Trainable relative positional encoding was proposed by \citet{shaw2018self,dai2019transformer}. For graph applications, topological information is as important as position. Structural and positional information in graphs is represented by the graph's Laplacian spectrum \cite{dwivedi2020generalization,kreuzer2021rethinking}, shortest-path distance \cite{ying2021transformers}, and kernel-based sub-graph \cite{mialon2021graphit,chen2022structure}, etc.

 \section{Numerical Experiments}
 This section presents the experimental results from a variety of PDE benchmarks, demonstrating the superior performance of PiT compared to many other operator learning methods. We also validate the discretization convergence property of PiT in Section \ref{section:zero_shot}. Section \ref{sec:ablation} presents rigorous comparative studies between self-attention and position-attention. In Section \ref{section:adaptivePIT}, we provide some insights on combining self-attention and position-attention. The impacts of hyperparameters are explored in Section \ref{section:meta_parameter}. More experimental results are presented in Appendix \ref{sec:more}.

\subsection{Benchmarks and Baselines}
Our tests encompass a diverse range of operator learning benchmarks: \textbf{InviscidBurgers} \cite{lanthaler2023nonlinear}, \textbf{ShockTube} \cite{lanthaler2023nonlinear}, \textbf{Darcy2D} \cite{li2021fourier}, \textbf{Vorticity} \cite{li2021fourier}, \textbf{Elasticity} \cite{li2022fourier},  and \textbf{NACA} \cite{li2022fourier}. 
These problems cover elliptic, parabolic, and hyperbolic PDEs, including challenging equations whose solutions exhibit discontinuities. 
 Data for these problems are collected on either structured meshes or irregular point clouds. Due to page limitations, we put the detailed setups of these problems in Appendix \ref{section:datasets}.

We compare PiT with various strong baselines in operator learning:  \textbf{DeepONet} \cite{lu2021learning}; \textbf{shift-DeepONet} \cite{lanthaler2023nonlinear}; \textbf{FNO} \cite{li2021fourier} and \textbf{FNO++}, the newest implementation \cite{NeuralOperator} of FNO using GELU activation and a two-layer MLP after each Fourier layer; \textbf{Geo-FNO} \cite{li2022fourier}; \textbf{Galerkin Transformer} \cite{cao2021choose}; \textbf{OFormer} \cite{li2023transformer}; \textbf{GNOT} \cite{hao2023gnot}; \textbf{ONO} \cite{xiao2023improved}. The latter four baselines are all Transformer-based neural operators. Besides the results presented in the following sections, we put more comprehensive comparisons between PiT and the baselines in the appendices; see parameter counts in Appendix \ref{sec:params_counts}; see training speed and memory usage in Appendix \ref{sec:runtime}.

\subsection{Main Results}
\label{section:main_results}

Table \ref{tab:main_result} presents the prediction errors of our method and the nine baselines for the six benchmarks. The results for the baselines are directly cited from those original papers, if applicable, and are marked as ``$-$" otherwise. The results of FNO++ are produced using the network hyper-parameters suggested in the references \cite{li2021fourier,lanthaler2023nonlinear}. Details about the network architectures and training configurations can be found in Appendix \ref{section:experimental_details}. 

\begin{table*}[t]
	\caption{Relative errors on the test sets of six benchmarks. The results of InviscidBurgers and ShockTube are reported with the relative $l_1$ errors. Other benchmarks are evaluated with the relative $l_2$ errors. {The best result of each task is \textbf{bolded}, and the second best result is \underline{underlined}.} The data of Darcy2D are represented on a $211\times 211$ uniform grid. The results of Galerkin Transformer and OFormer for Elasticity and NACA are cited from \citet{hao2023gnot}. The results of FNO for Elasticity and NACA are cited from \citet{li2022fourier}.}
	\label{tab:main_result}
	\vskip 0.15in
	\begin{center}
	\resizebox{0.8\linewidth}{!}{
		\begin{tabular}{lcccccc}
			\toprule
			\multicolumn{1}{c}{} &InviscidBurgers        &ShockTube             &Darcy2D                &Vorticity             &Elasticity             &NACA         \\
			\midrule
			DeepONet             &$0.285$                &$0.0422$              &$-$                    &$-$                   &$-$                    &$-$        \\
			shift-DeepONet       &$0.0783$               &$0.0276$              &$-$                    &$-$                   &$-$                    &$-$        \\
			FNO++                &$\mathbf{0.00995}$     &$0.0194$              &$\underline{0.00509}$  &$0.1315$              &$-$                    &$-$          \\
			FNO                  &$0.0157$               &$\underline{0.0156}$  &$0.0109$               &$0.1559$              &$0.0508$               &$0.0421$     \\
			OFormer              &$-$                    &$-$                   &$0.0128$               &$0.1755$              &$0.0183$               &$0.0183$     \\
			Galerkin Transformer &$-$                    &$-$                   &$0.00844$              &$0.1399$              &$0.0201$               &$0.0161$     \\
			GNOT                 &$-$                    &$-$                   &$-$                    &$0.138$               &$\underline{0.00865}$  &$0.00757$    \\
			ONO                  &$-$                    &$-$                   &$-$                    &$\underline{0.1195}$  &$0.0118$               &$\underline{0.0056}$     \\
			Geo-FNO              &$-$                    &$-$                   &$-$                    &$-$                   &$0.0229$               &$0.0138$     \\
			\midrule
			PiT                  &$\underline{0.0136}$   &$\mathbf{0.0122}$     &$\mathbf{0.00485}$     &$\mathbf{0.1140}$     &$\mathbf{{0.00649}}$   &$\mathbf{0.00480}$   \\
			\bottomrule
		\end{tabular}
    }
	\end{center}
	\vskip -0.1in
\end{table*}

In InviscidBurgers and ShockTube, {\em PiT exhibits  excellent performance, comparable to FNO/FNO++, and significant superiority over DeepONet and shift-DeepONet.} Both tasks pose notable challenges due to the discontinuous target functions in the solution operators (see Figures \ref{fig:Burgers} and  \ref{fig:ShockTube}), which are inherently difficult for neural networks to learn. As \citet{lanthaler2023nonlinear} pointed out, DeepONet fails to effectively address such difficulties, while shift-DeepONet enhances the performance by incorporating shift-net. 
PiT overcomes the challenges thanks to its nonlinear Transformer architecture with position-attention. 
In the InviscidBurgers benchmark, PiT's prediction error is remarkably lower, at just 17\% of shift-DeepONet's error and a mere 4.8\% of DeepONet's error. In the ShockTube benchmark, PiT's prediction error is about 45\% of shift-DeepONet's and 29\% of DeepONet's. 

Furthermore, in both the Darcy2D and Vorticity benchmarks, 
{\em PiT achieves the lowest prediction errors, outperforming all tested baselines.}  
In the Darcy2D task, PiT's prediction error is only 38\% to 57\% of those of OFormer, FNO, and Galerkin Transformer. 
PiT also demonstrates the best performance 
in the Vorticity benchmark, representing a challenging operator learning task due to data scarcity and the complex patterns of turbulent flow. 
These results 
indicate that leveraging spatial interrelations of mesh points is highly beneficial for the attention mechanism in learning these operators.

In the Elasticity and NACA tasks, the PDEs are defined in irregular domains with complex geometries, and the data are sampled on unstructured point clouds for Elasticity and on a deformed mesh for NACA. The complexity of the data and geometry presents significant challenges for operator learning. {\em Again, PiT, with its position-attention mechanism, exhibits superior accuracy over all the baselines}, including the other four Transformers (OFormer, Galerkin Transformer, GNOT, and ONO) based on self-attention. This suggests that position-attention is crucial, while self-attention might be unnecessary for Transformer-based operator learning.

In addition to its outstanding accuracy, PiT also exhibits high efficiency in terms of training costs. For example, Table \ref{tab:Darcy2Dmore} displays the training times of PiT with data on various mesh resolutions. These results validate PiT's sub-linear scaling of training time with mesh resolution, consistent with our computational complexity analysis in Section \ref{section:topoformer}.

%%%%%%%%%%%%%%%%%%%%%%%%5
\subsection{Discretization Convergence Tests}
\label{section:zero_shot}
\begin{figure}[ht!]
	\centering
	\shifttext{-6mm}{\includegraphics[width=0.70\linewidth]{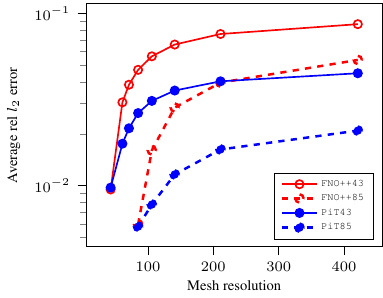}}
%    \begin{tikzpicture}[trim axis left, trim axis right]
%	     \begin{axis}[width=0.8\linewidth, transform shape,
%				     tick align=outside,
%		             xtick pos=bottom,ytick pos=left,ylabel={Average rel $l_2$ error},
%		             xticklabel style={/pgf/number format/fixed,/pgf/number format/precision=3, font=\scriptsize},
%		             scaled x ticks=false,
%		             yticklabel style={/pgf/number format/fixed,/pgf/number format/precision=3, font=\scriptsize},
%		             scaled y ticks=false,ymode=log,
%		             every major tick/.append style={major tick length=4.0pt, black},
%				     label style={font=\scriptsize},
%				     xlabel={Resolution},
%				     legend style={at={(0.99,0.68)}, nodes={scale=0.7, transform shape},font=\scriptsize},
%				     legend cell align={left}
%				     ]
%		     \legend{FNO++ 43,FNO++ 85,PiT 43,PiT 85};
%		     \addplot[thick, color=red, mark=o] table[x index=0,y index=1] {./Figures/zero_shot43.txt};
%		     \addplot[very thick, dashed, color=red, mark=o] table[x index=0,y index=1] {./Figures/zero_shot85.txt};
%		     \addplot[thick, color=blue, mark=*] table[x index=0,y index=2] {./Figures/zero_shot43.txt};
%		     \addplot[very thick, dashed, color=blue, mark=*] table[x index=0,y index=2] {./Figures/zero_shot85.txt};
%	     \end{axis}	
%    \end{tikzpicture}
	\caption{Discretization convergence tests on Darcy2D.}
	\label{fig:zero-shot}
\end{figure}

The Darcy2D dataset, originally collected on a $421^2$ Cartesian grid, is  downscaled onto a sequence of coarser meshes to serve as reference solutions. 
This enables the assessment of PiT's discretization convergence via zero-shot super-resolution evaluation. To this end, we train neural operators with data on a coarse mesh. After training, the learned operators are tested on a sequence of refined meshes:  $43^2$, 
$61^2$, $71^2$, $85^2$, $106^2$, $141^2$, $211^2$, $421^2$, respectively. 
The testing errors on these meshes are illustrated in Figure \ref{fig:zero-shot},  where PiT and FNO++ trained on the $43^2$ mesh (resp.~the $85^2$ mesh) are denoted as {\tt PiT43} and {\tt FNO++43} (resp.~{\tt PiT85} and {\tt FNO++85}). 

As seen from Figure \ref{fig:zero-shot}, for operator learning on the $43^2$ mesh, FNO++'s prediction error surges from $0.95\%$ to $8.67\%$ as the testing mesh resolution increases to $421^2$, while PiT's prediction error rises from $0.97\%$ to only $4.50\%$ (which is 48\% lower than that of FNO++). The greater robustness and accuracy of PiT compared to FNO++ are also observed in Figure \ref{fig:zero-shot} for the operators trained with $85^2$ mesh data. These results demonstrate PiT's superiority over FNO++ in terms of discretization convergence.
\subsection{Comparative Ablation Study}
\label{sec:ablation}
We have demonstrated that PiT with position-attention delivers superior performance compared to existing Transformer-based neural operators that utilize self-attention. This suggests that self-attention might not be necessary for operator learning. In this section, we provide a more rigorous ablation study to compare  position-attention and vanilla self-attention for operator learning in PDEs. We test two vanilla self-attention models:

\textbf{SelfAtt A}: All PosAtt layers in PiT are replaced with the vanilla self-attention. Three out of six benchmarks encounter an ``out of memory" (OOM) issue with a single 24GB RTX-3090 GPU.

\textbf{SelfAtt B}: Only PosAtt layers in Processor are replaced with self-attention, and this avoids the OOM issue.

\Cref{tab:ablation} presents the testing errors, parameter counts, and training time. These results validate that PiT is consistently more accurate than Transformers built upon vanilla self-attention, without trade-offs in efficiency.
%%%%%%%%%%%%%%%%%%%5
\begin{table*}[tbhp]
    \caption{Comparisons of position-attention and vanilla self-attention on all benchmark problems. The best result of each task is \textbf{bolded}.}
    \label{tab:ablation}
    \vskip 0.15in
    \begin{center}
    \resizebox{0.9\textwidth}{!}{
	    \begin{tabular}{llllllll}
		    \toprule 
		    \multicolumn{1}{c}{}                                                                    &Model            &InviscidBurgers    &ShockTube          &Darcy2D             &Vorticity             &Elasticity            &NACA  \\
		    \midrule
		    \multirow{3}{*}{Testing errors}                                                         &SelfAtt A        &$0.016$            &$0.0259$           &OOM                 &OOM                   &$0.0295$              &OOM\\
		                                                                                            &SelfAtt B        &$0.0235$           &$0.016$            &$0.0072$	           &$0.156$	              &$0.169$	             &$0.0164$\\
		                                                                                            &PiT              &$\mathbf{0.0136}$  &$\mathbf{0.0122}$  &$\mathbf{0.00485}$  &$\mathbf{0.114}$      &$\mathbf{0.00649}$    &$\mathbf{0.0048}$\\
		    \midrule
		    \multirow{3}{*}{Parameter counts}                                                       &SelfAtt A        &$152,833$          &$152,961$          &OOM                 &OOM                   &$9,732,609$           &OOM\\
		                                                                                            &SelfAtt B        &$128,263$          &$128,391$          &$444,677$	       &$1,776,387$	          &$8,684,049$	         &$1,774,341$\\
		                                                                                            &PiT              &$\mathbf{95,503}$  &$\mathbf{95,631}$  &$\mathbf{313,613}$  &$\mathbf{1,252,103}$  &$\mathbf{6,586,929}$  &$\mathbf{1,250,061}$\\
		    \midrule
		    \multirow{3}{*}{\begin{tabular}{@{}c@{}}Training time\\second/epoch\end{tabular}}       &SelfAtt A        &$1.73$             &$5.51$             &OOM                 &OOM                   &$\mathbf{7.13}$       &OOM\\ 
		                                                                                            &SelfAtt B        &$1.47$             &$1.73$             &$15.3$	           &$18.7$	              &$9.30$	             &$21.1$\\ 
		                                                                                            &PiT              &$\mathbf{0.938}$   &$\mathbf{1.04}$    &$\mathbf{14.7}$     &$\mathbf{16.3}$       &$7.69$                &$\mathbf{15.3}$\\  
		    \bottomrule
	    \end{tabular}
    }
    \end{center}
    \vskip -0.1in
\end{table*}

%%%%%%%%%%%%%%%%%%%%%%%%%%%%%%%%%%%%
\subsection{Can Self-attention Enhance PiT?}
\label{section:adaptivePIT}
In this section, we aim to answer to such a question: Does combining self-attention and position-attention enhance PiT? To address this, let us consider a Transformer, termed \emph{Self-PiT}, based on a combined attention mechanism:
\begin{equation}
    \begin{aligned}
        &\text{SelfPosAtt}(U;D)\\
        &=\text{Softmax}\left(-{\lambda} D+\frac{UW^Q(UW^K)^T}{\sqrt{d_{\ell+1}}}\right)UW^V.
    \end{aligned}
    \label{eq:adaptiveposatt}
\end{equation}
We have tested Self-PiT on the InviscidBurgers and ShockTube benchmarks, for which the prediction errors are $0.00816$ and $0.0179$, respectively. 
By comparing them with the results of PiT in Table \ref{tab:main_result}, we conclude that Self-PiT does not consistently outperform PiT in terms of accuracy, yet it requires more computational complexities.

\subsection{Hyperparameter Study}
\label{section:meta_parameter}
\begin{figure*}[ht!]
	\centering
	\begin{subfigure}[c]{0.32\textwidth}
	    \centering
	    \includegraphics{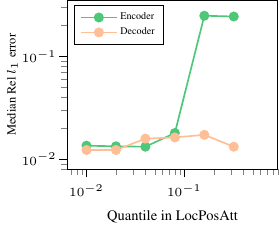}
    \label{fig:quantile_burgers}
    \end{subfigure}%
    \begin{subfigure}[c]{0.32\textwidth}
	    \centering
	    \includegraphics{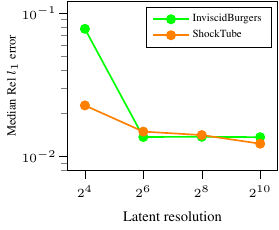}
    \label{fig:downsample}
    \end{subfigure}%
    \begin{subfigure}[c]{0.32\textwidth}
	    \centering
	    \includegraphics{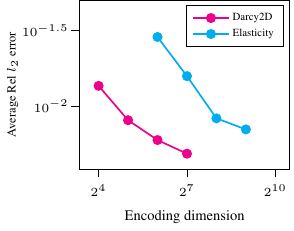}
	\label{fig:width}
    \end{subfigure}
\caption{Impacts of the three hyperparameters in PiT. }
\label{fig:hyper_parameter}
\end{figure*}

Figure \ref{fig:hyper_parameter} illustrates the impacts of the following three important hyperparameters in PiT. 

\textbf{Quantile in LocPosAtt}. 
A smaller quantile means a more compact receptive field in local position-attention, resulting in a stronger focus on local features. The results in Figure \ref{fig:hyper_parameter} indicate that PiT's performance on ShockTube is sensitive to the quantile in the Encoder but not sensitive to the quantile in the Decoder. Using a small quantile in the Encoder is critical; otherwise, PiT may yield a large prediction error.

\textbf{Latent Mesh Resolution $N_v$}. 
This hyperparameter balances computational efficiency and information retention on coarse latent meshes. Choosing a relatively large $N_v$ is crucial to retain essential information in PiT's Processor. Figure \ref{fig:hyper_parameter} shows that the prediction error decreases as $N_v$ increases, as expected. However, this benefit diminishes rapidly as $N_v$ reaches 64. On latent mesh of merely 64 points, PiT is efficient and sufficiently accurate for InviscidBurgers and ShockTube, even though the input data for these tasks is sampled on 1,024 and 2,048 grid points, respectively.

\textbf{Encoding Dimension $d_v$ (Network Width)}. 
$d_v$ affects the model's expressive capacity. As expected, Figure \ref{fig:hyper_parameter} shows a consistent decrease in relative errors as $d_v$ increases. For Darcy2D, the PiT model with $d_v=32$ has only $20,000$ trainable parameters, yet its prediction error is merely $0.00808$, which is already lower than the errors of FNO, OFormer, and Galerkin Transformer in Table \ref{tab:main_result}. While the latter three methods all have over $2,000,000$ trainable parameters, PiT achieves superior performance with only $20,000$ parameters, demonstrating its parsimonious nature.

%%%%%%%%%%%%%%%%%%%%%
\section{Conclusions}
Inspired by numerical mathematics, we propose position-attention (and two variants) and  Position-induced Transformer (PiT) for operator learning in PDEs. 
PiT exhibits outstanding performance across various PDE benchmarks, surpassing many operator learning baselines. Notably, PiT is discretization-convergent, enabling effective generalization to new meshes with different resolutions. We conclude that the position-attention mechanism is highly efficient for learning nonlinear operators, even in challenging hyperbolic PDEs with discontinuous solutions. Unlike classical self-attention, our position-attention is induced solely by the spatial interrelations of sampling points, without relying on input function values. Position-attention greatly enhances computational efficiency and effectively integrates positional knowledge. Our results demonstrate that position-attention is crucial for operator learning, and {\em positional knowledge is all you need}.

\section*{Impact Statement}

PiT emerges as a versatile operator learning framework, applicable to both the surrogate modeling of known parametric PDEs and the data-driven learning of unknown PDEs. It may broadly influence various PDE-related fields such as physics, engineering, biology, and finance, marking an important advancement in AI for science.

Incorporating human insights, which encompass physical and numerical knowledge, is recognized as pivotal in data-driven modeling. Our work not only highlights this integration but also facilitates the development of new operator learning frameworks and the enhancement of existing neural operators. This fosters growth in the realm of scientific machine learning.

One possible negative impact relates to the computational cost of PiT for high-dimensional, large-scale PDEs. To retain essential information of the operators, the latent mesh in PiT may require a large number of nodal points, which notably increases the computational cost. This directs future research endeavors toward further enhancing the current position-attention framework through sparse approximations, low-rank approximations, or Softmax-free variants.

\section*{Acknowledgements}
This work was partially supported by National Natural Science Foundation of China (Grant No.~92370108) and 
Shenzhen Science and Technology Program (Grant No.~RCJC20221008092757098).

%\nocite{langley00}

\bibliography{bib}
\bibliographystyle{icml2024}
\newpage
\appendix
\onecolumn
\section{Datasets and Setups of Benchmarks}
\label{section:datasets}
We briefly present the datasets of the benchmarks considered in this work.

The datasets for InviscidBurgers and ShockTube are obtained from \citet{lanthaler2023nonlinear} and are available for download at \url{https://zenodo.org/records/7118642}.

The datasets for Darcy2D and Vorticity are obtained from \citet{li2021fourier} and can be downloaded from  \url{https://drive.google.com/drive/folders/1UnbQh2WWc6knEHbLn-ZaXrKUZhp7pjt-}.

The datasets of Elasticity and NACA are obtained from \citet{li2022fourier} and are accessible for download at  \url{https://drive.google.com/drive/folders/1YBuaoTdOSr_qzaow-G-iwvbUI7fiUzu8}.
%%%%%%%%%%%%%%%%%%%%%%%%%%%%

More detailed descriptions about these datasets and setups are provided as follows. {\em All our codes can be found in the Supplementary Material.}

\subsection{Data and Setup for Benchmark 1: Inviscid Burgers}
In this benchmark, we consider a nonlinear hyperbolic PDE, namely, the 1D inviscid Burgers' equation \cite{lanthaler2023nonlinear}:
\begin{equation}
	\begin{alignedat}{2}
			\partial_t u +  \partial_x \left(\frac{u^2}{2}\right) &= 0, \quad && (x,t)\in [0,1]\times \mathbb{R}^+,\\
			u(\;\cdot\;,0) &= \bar{u}(x),\qquad && x\in[0,1],
	\end{alignedat}
	\label{equation:eg_ivbg}
\end{equation}
where the initial condition $\bar{u}(x)$ is sampled from a Gaussian random field. Our objective is to learn the operator that maps various initial conditions to the corresponding entropy solutions at $T=0.1$.
Due to the nonlinear hyperbolic nature of this PDE, its solution can develop discontinuities, even if the initial condition is smooth. This feature makes the task of learning the operator challenging.
The solution data at $T=0.1$ were obtained using a high-order finite volume scheme on a uniform mesh of 1,024 cells \cite{lanthaler2023nonlinear}. The full dataset in \citet{lanthaler2023nonlinear} comprises 1,024 input-output pairs for training and validation, and 128 pairs for testing.
Since we employ a training-testing setup, the validation set with 74 pairs is excluded from our experiments for PiT. In other words, we use only the 950 data pairs as our training set to ensure a fair comparison with the baselines.

\subsection{Data and Setup for Benchmark 2: ShockTube}
The ShockTube benchmark also involves a nonlinear hyperbolic system of PDEs, whose solutions contain discontinuities.
Specifically, we consider the shock tube problem of the 1D compressible Euler equations \cite{lanthaler2023nonlinear}:
\begin{equation}
	\partial_t \textbf{U} + \textbf{F}(\textbf{U})_x = 0,  \quad  (x,t)\in [-5,5]\times \mathbb{R}^+, 
	\label{equation:eg_euler}
\end{equation}
where the conservative vector and flux are respectively given by 
\begin{equation*}
    \textbf{U} = \begin{pmatrix} \rho \\ \rho u\\ E \end{pmatrix},\quad \textbf{F}(\textbf{U}) = \begin{pmatrix} \rho u \\ \rho u^2\\ (E+p)u \end{pmatrix}.
\end{equation*}
Here, $\rho$, $u$, and $p$ denote the fluid density, velocity, and pressure, respectively. The total energy, $E$, consists of the kinetic and internal energies, expressed as $E=\frac{1}{2}\rho u^2+\frac{p}{\gamma - 1}$, with the constant adiabatic index $\gamma=1.4$.

The objective is to learn the operator that maps the initial conservative function $\textbf{U}_0(x)$ to the total energy function $E(x,t)$ at $t=1.5$. The data for the target total energy function $E(x,t=1.5)$ are obtained on a uniform grid with 2,048 cells \cite{lanthaler2023nonlinear}. The full dataset in \citet{lanthaler2023nonlinear} consists of 1,024 and 128 input-output pairs for training and testing, respectively.

\subsection{Data and Setup for Benchmark 3: Darcy2D}
\label{section:darcy}
In the Darcy2D benchmark, we consider an elliptic equation with the Dirichlet boundary condition \cite{anandkumar2019neural,li2020multipole,li2021fourier}:
\begin{equation}
    \begin{alignedat}{2}
			    -\nabla\cdot (a\nabla u) &= f, \quad && x\in (0,1)^2,\\
			    u &= 0, && x\in \partial(0,1)^2,
    \end{alignedat}
\label{eq:eg_darcy}
\end{equation}
where $f=1$ is the forcing term. Our objective is to learn the operator that maps the permeability $a(x)$ to the pressure field $u(x)$. The dataset comprises 1,024 and 100 input-output pairs for training and testing, respectively. These pairs are obtained by solving the PDE \eqref{eq:eg_darcy} on a $421\times421$ grid. The permeabilities $a$ are random piecewise constant functions generated according to $a=\psi(\mu)$, where $\psi$ takes 12 for positive realizations of $\mu$ and 3 for negative ones, while $\mu$ itself is a Gaussian random field with the Neumann boundary condition. 
 
It is worth noting that the input functions of Darcy2D's solution operator, characterized by piecewise constants with jumps at interfaces (see Figure \ref{fig:Darcy2D}), pose difficulties in detecting discontinuities from relatively coarse data.

\subsection{Data and Setup for Benchmark 4: Vorticity}
This benchmark is related to the two-dimensional incompressible Navier–Stokes equations in the vorticity form \cite{li2021fourier}:
\begin{equation}
	\begin{alignedat}{2}
			\partial_t \omega  + \textbf{u}\cdot \nabla \omega  &= \nu \Delta \omega + f, \quad  && x\in (0,1)^2, t> 0,\\
			\nabla \cdot \textbf{u} &= 0, && x\in (0,1)^2, t>0,\\
			\omega (\;\cdot\;,t=0) &= \omega_0,&& x\in (0,1)^2,
	\end{alignedat}
	\label{equation:eg_ns}
\end{equation}
where $\textbf{u}(x,t)$ is the velocity field, $\omega = \nabla \times \textbf{u}$ is the vorticity, $\nu = 10^{-4}$ denotes the viscosity, and $f(x) = 0.1(\sin(2\pi(x_1 + x_2)) + \cos(2\pi(x_1 + x_2)))$ represents a periodic external force. The initial condition, $\omega_0(x)$, is generated by Gaussian random fields. All the data are generated on a $256 \times 256$ Cartesian grid and collected with a time lag $\Delta = 1$, then downsampled to a $64 \times 64$ grid. The objective is to learn the operator that maps the vorticity snapshots in the time period $t \in [1, 10]$ to the future vorticity snapshots up to $T = 30$. The dataset consists of 1,000 samples for training and 200 samples for testing. 

\subsection{Data and Setup for Benchmark 5: Elasticity}
We consider a hyper-elastic material in a unit cell with a cavity of random shape at the center \cite{li2022fourier}. The material is clamped at the bottom side and stretched with a force applied at the upper side. The displacement field for a solid body is governed by the following PDEs \cite{li2022fourier}:
\begin{equation}
    \rho\frac{\partial^2\textbf{u}}{\partial t^2} + \nabla\cdot\bm{\sigma} = 0
\end{equation}
with $\rho$ being the density, $\textbf{u}$ the displacement, and $\bm{\sigma}$ the stress tensor. This system is closed by constitutive models relating the strain and stress tensors.

The objective is to learn the solution operator that maps mesh point locations to the displacement field. The dataset in \citet{li2022fourier} consists of 1,000 samples for training and 200 samples for testing. Each sample is represented by a point cloud with 972 nodal points. The locations of these points vary across different samples, as the shapes of the cavities are different. See \citet{li2022fourier} for more details.

\subsection{Data and Setup for Benchmark 6: NACA}
This task is related to the transonic flow over airfoils described by the 2D compressible Euler equations \cite{li2022fourier}:
\begin{equation}
    \partial_t \textbf{U} + \nabla\cdot\textbf{F}(\textbf{U}) = 0,
\end{equation}
where the conservative vector and flux are respectively
\begin{equation*}
    \textbf{U} = \begin{pmatrix} \rho \\ \rho \textbf{v}\\ E \end{pmatrix},\quad \textbf{F}(\textbf{U}) = \begin{pmatrix} \rho v \\ \rho\textbf{v}\otimes\textbf{v}+p\textbf{I}\\ (E+p)\textbf{v} \end{pmatrix}.
\end{equation*}
Here, $\rho$ represents the density, $\textbf{v}$ denotes the velocity field, and $p$ is the pressure. The total energy is given by $E=\frac{1}{2}\rho|\textbf{v}|^2+\frac{p}{\gamma-1}$, with $\gamma=1.4$ being the constant adiabatic index.

The objective is to learn the operator that maps mesh point locations to the Mach number function defined on these mesh points. The dataset in \citet{li2022fourier} consists of 1,000 samples for training and 200 samples for testing. Each sample corresponds to a different airfoil shape and is represented on a C-grid mesh refined near the airfoil's surface. The dataset has been transformed onto a regular $221\times 51$ grid.

\section{Proof of Theorem \ref{prop:pa}} \label{app:proof}
In this section, we present the proof Theorem \ref{prop:pa}. 

\begin{proof}
	For any $x \in \Omega$, 
 define 
\begin{align*}
	{\mathcal G}_n(x) &: = \frac{1}{ |X_n| } \sum_{k=1}^{|X_n|} \exp( -\lambda \| x-x_k \|^2 ) (UW^V)_k,  
	\\
	{\mathcal H}_n(x) &:= \frac{1}{ |X_n| } \sum_{k=1}^{|X_n|} \exp( -\lambda \| x-x_k \|^2 ). 
\end{align*}
	Then we have 
		\begin{align*}
		\text{PosAtt}(U^n;D^n)_i &= \sum_{k=1}^{|X_n|}\frac{\exp(-\lambda D_{ik})}{\sum_{j=1}^{|X_n|}\exp(-\lambda D_{ij})}(UW^V)_k 
		\\
		&=  \frac{  \displaystyle \frac{1}{|X_n|} \sum_{k=1}^{|X_n|}{\exp(-\lambda D_{ik})}{}(UW^V)_k } { \displaystyle \frac{1}{|X_n|} \sum_{j=1}^{|X_n|}\exp(-\lambda D_{ij}) }  
		\\
		&=
		 \frac{ {\mathcal G}_n(x_i) }{   {\mathcal H}_n(x_i)  } =: {\mathcal F}_n(x_i).
	\end{align*}
	Note that ${\mathcal G}_n(x)$ and ${\mathcal H}_n(x)$ are 
	 the Monte--Carlo integration approximations to  
	 $$
	 		{\mathcal G}(x) := \int_{\Omega} \exp( -\lambda \| x - y \|_2^2) v(y)W^V d \mu_{\Omega}(y)
	 $$
	 and 
	 $$
	 	{\mathcal H}(x)  := \int_{\Omega} \exp( -\lambda \| x - y \|_2^2) d \mu_{\Omega}(y),
	 $$
	 respectively. 
	 By the strong law of large numbers, we know that 
	 $$
	 {\rm Pr} \left\{  \lim_{n \to \infty}  {\mathcal G}_n(x) = {\mathcal G}(x) \right\} = 1,
	 $$
	 and 
	 $$
	{\rm Pr} \left\{  \lim_{n \to \infty}  {\mathcal H}_n(x) = {\mathcal H}(x) \right\} = 1. 
	$$	 
	 It follows that 
\begin{equation}\label{eq:789}
		 {\mathcal F}_n(x)  -   {\mathcal F} (x) = \frac{ {\mathcal G}_n (x) }{  {\mathcal H}_n (x) }  - \frac{ {\mathcal G} (x) }{  {\mathcal H} (x) }  \xrightarrow[]{a.s.} 0. 
\end{equation}
	 Define 
	 $$
	 A_n := \int_{\Omega} \left| \frac{ {\mathcal G}_n(x) }{  {\mathcal H}_n(x) } - \frac{ {\mathcal G}(x) }{  {\mathcal H} (x) }   \right| d \mu_{\Omega} (x) =  \int_{\Omega} \left|   {\mathcal F}_n(x)  -   {\mathcal F} (x)    \right| d \mu_{\Omega} (x). 
	 $$
	 Then, for any $\varepsilon >0$, we have 
	 \begin{equation}\label{eq:345}
	 	{\rm Pr}  \left\{ \lim_{n \to \infty} A_n= 0 \right\} = 1, \qquad 
	 \lim_{n \to \infty}	{\rm Pr} \left\{  A_n \le \frac{ \varepsilon}2 \right\} = 1. 
	 \end{equation}
	 Using the Chebychev inequality and the  elementary inequality $(p-q)^2 \le p^2+q^2$ for non-negative $p$ and $q$, we have  
\begin{equation*}
		 	 \begin{aligned}
	 	 	{\rm Pr} \left\{ \left|	 \frac{1}{|X_n|} \sum_{i=1}^{|X_n|}  \big| {\mathcal F}_n(x_i)  -   {\mathcal F} (x_i)  \big| - A_n \right| > \frac{ \varepsilon}2   \right\} 
	 	 	 &\le \frac{4}{ |X_n| \varepsilon^2 } 
	 	\int_\Omega  \Big(  \left| {\mathcal F}_n(x)  -   {\mathcal F} (x)  \right| - A_n  \Big)^2 d \mu_\Omega(x) 
	 	\\
	 	& \le 
	 	\frac{4}{ |X_n| \varepsilon^2 } \left( A_n^2 + 
	 	\int_\Omega   \big(  {\mathcal F}_n(x)  -   {\mathcal F} (x)  \big)^2  d \mu_\Omega(x) 
	 	\right),
	 \end{aligned}
\end{equation*}
	 which further yields 
		 	 \begin{align*}
		{\rm Pr} \left\{ \left|	 \frac{1}{|X_n|} \sum_{i=1}^{|X_n|}  \big| {\mathcal F}_n(x_i)  -   {\mathcal F} (x_i)  \big| - A_n \right| \le  \frac{ \varepsilon}2   \right\} 
		\ge 1 - \frac{4}{ |X_n| \varepsilon^2 } \left( A_n^2 + 
		\int_\Omega   \big(  {\mathcal F}_n(x)  -   {\mathcal F} (x)  \big)^2  d \mu_\Omega(x) 
		\right). 
	\end{align*} 
	 Because 
	 $$
	 {\rm Pr} \left\{  	 \frac{1}{|X_n|} \sum_{i=1}^{|X_n|}  \big| {\mathcal F}_n(x_i)  -   {\mathcal F} (x_i)  \big|    \le  A_n + \frac{ \varepsilon}2   \right\} 
	 \ge 
	 {\rm Pr} \left\{ \left|	 \frac{1}{|X_n|} \sum_{i=1}^{|X_n|}  \big| {\mathcal F}_n(x_i)  -   {\mathcal F} (x_i)  \big| - A_n \right| \le  \frac{ \varepsilon}2   \right\},  
	 $$
	 we obtain 
		 	 \begin{align}\label{eq:456}
{\rm Pr} \left\{  	 \frac{1}{|X_n|} \sum_{i=1}^{|X_n|}  \big| {\mathcal F}_n(x_i)  -   {\mathcal F} (x_i)  \big|    \le  A_n + \frac{ \varepsilon}2   \right\} 
	\ge 1 - \frac{4}{ |X_n| \varepsilon^2 } \left( A_n^2 + 
	\int_\Omega   \big(  {\mathcal F}_n(x)  -   {\mathcal F} (x)  \big)^2  d \mu_\Omega(x) 
	\right). 
\end{align} 	 
	 Notice that if 
	 $$
	 \frac{1}{|X_n|} \sum_{i=1}^{|X_n|}  \big| {\mathcal F}_n(x_i)  -   {\mathcal F} (x_i)  \big|    \le  A_n + \frac{ \varepsilon}2 \quad \mbox{and} \quad  A_n \le \frac{ \varepsilon}2,
	 $$
	 then 
	 $$
	 \frac{1}{|X_n|} \sum_{i=1}^{|X_n|}  \big| {\mathcal F}_n(x_i)  -   {\mathcal F} (x_i)  \big| \le  \varepsilon. 
	 $$
	 This implies 
\begin{align*}
 {\rm Pr} \left\{ 	\frac{1}{|X_n|} \sum_{i=1}^{|X_n|}  \big| {\mathcal F}_n(x_i)  -   {\mathcal F} (x_i)  \big| \le  \varepsilon \right\} 
 & \ge {\rm Pr} \left\{   \frac{1}{|X_n|} \sum_{i=1}^{|X_n|}  \big| {\mathcal F}_n(x_i)  -   {\mathcal F} (x_i)  \big|    \le  A_n + \frac{ \varepsilon}2 ~~ \mbox{and} ~~ A_n \le \frac{ \varepsilon}2 \right\}
 \\
 & \ge {\rm Pr} \left\{   \frac{1}{|X_n|} \sum_{i=1}^{|X_n|}  \big| {\mathcal F}_n(x_i)  -   {\mathcal F} (x_i)  \big|    \le  A_n + \frac{ \varepsilon}2  \right\} + {\rm Pr} \left\{   A_n \le \frac{ \varepsilon}2 \right\} - 1, 
\end{align*}
where the second step follows from the probability inequality ${\rm Pr}(a \cap b) \ge {\rm Pr}(a) + {\rm Pr}( b) -1 $. 
Combing it with \eqref{eq:456}, we obtain 
\begin{align*}
	 {\rm Pr} \left\{ 	\frac{1}{|X_n|} \sum_{i=1}^{|X_n|}  \big| {\mathcal F}_n(x_i)  -   {\mathcal F} (x_i)  \big| \le  \varepsilon \right\} 
 \ge   {\rm Pr} \left\{   A_n \le \frac{ \varepsilon}2 \right\} - \frac{4}{ |X_n| \varepsilon^2 } \left( A_n^2 + 
	 \int_\Omega   \big(  {\mathcal F}_n(x)  -   {\mathcal F} (x)  \big)^2  d \mu_\Omega(x) 
	 \right). 
\end{align*}
Taking $n \to +\infty$ and using \eqref{eq:789}--\eqref{eq:345}, we obtain 
 $$
 1\ge \lim_{n\to+\infty} {\rm Pr} \left\{ 	\frac{1}{|X_n|} \sum_{i=1}^{|X_n|}  \big| {\mathcal F}_n(x_i)  -   {\mathcal F} (x_i)  \big| \le  \varepsilon \right\}  \ge 1. 
 $$
 Therefore, 
	 	\begin{equation*}
	 	\lim_{n\to+\infty}  {\rm Pr} \left\{ 	\frac{1}{|X_n|} \Big\| \text{PosAtt}(U^n;D^n) - {\mathcal F}   |_{X_n} \Big\| \le \varepsilon \right\} = 1, 
	 \end{equation*}
	 for any $\varepsilon > 0$. This means the position-attention \eqref{eq:PosAtt}  
	  converges in probability to the  integral operator \eqref{eq:2}. 
	The proof is completed. 
\end{proof}

\section{Technical Aspects of Position-attention}
\label{section:implementation}
\subsection{Ensuring Non-negativity of $\lambda$}

Ensuring the non-negativity of $\lambda$ is crucial for position-attention (as well as the cross and local variants) to be well-defined. There are several methods to maintain the non-negativity of during training, including training PiT via constrained optimization or transforming into a positive function of itself (e.g., $\lambda^2$). 
These methods can result in varying performance of the trained neural operator. While replacing $\lambda$ with $\lambda^2$ is an intuitive and straightforward choice, we have empirically observed higher prediction errors in PiT models constructed this way. Conversely, using $\tan(\lambda)$ and training PiTs under the constraint $0\leq\lambda<\frac{\pi}{2}$ has proven to be more advantageous in certain benchmark cases. This is demonstrated by the results in Table \ref{tab:tan}. 

\begin{table}[ht!]
	\caption{Prediction errors of PiT with two different methods for ensuring non-negativity of $\lambda$.}
	\label{tab:tan}
	\vskip 0.15in
	\begin{center}
		\resizebox{0.8\textwidth}{!}{
			\begin{tabular}{ccccccc}
				\toprule
				\multicolumn{1}{c}{}                    &InviscidBurgers &ShockTube  &Darcy2D  &Vorticity &Elasticity &NACA \\
				\midrule
			    \multicolumn{1}{c}{Replace $\lambda$ with $\lambda^2$}          &$0.0273$             &$\mathbf{0.0122}$        &$0.0102$      &$\mathbf{0.1169}$       &$\mathbf{{0.00649}}$       &$0.162$  \\
				\multicolumn{1}{c}{\begin{tabular}{c} Use $\tan(\lambda)$ and\\constrained optimization\end{tabular}} &$\mathbf{0.0136}$ &$0.0154$  &$\mathbf{0.00485}$    &$0.5757949$  &$0.00701$   &$\mathbf{0.00480}$ \\
				\bottomrule
			\end{tabular}
		}
	\end{center}
	\vskip -0.1in
\end{table}
\subsection{Multi-head Implementation for Position-attention}
In Transformers, the multi-head technique for self-attention is widely used to enhance performance. In parallel, we can formulate the multi-head implementation for position-attention as:
\begin{equation}\label{aaa1}
	\begin{aligned}
		\text{MultiHeadPosAtt}(U;D) &= \text{Concat}(\text{head}^1,\ldots,\text{head}^h),\\
		\text{where head}^i &=\text{Softmax}(-\lambda^iD)UW^i. \\
	\end{aligned}
\end{equation}
Here, $h$, which divides $d_v$ evenly, represents the number of heads; $\lambda^i>0$ and $W^i\in\mathbb{R}^{d_v\times\frac{d_v}{h}}$ are the training parameters for the $i$th head; the Concat operation in \eqref{aaa1} concatenates the outputs of all the heads, which are matrices in $\mathbb{R}^{N\times \frac{d_v}{h}}$, to produce the output of MultiHeadPosAtt in $\mathbb{R}^{N\times d_v}$. This implementation also applies to cross and local position-attention.

The multi-head implementation for position-attention defined above performs $h$ independent convolutions with trainable $\lambda^i$. With a larger $\lambda^i$, the attention decays more rapidly as the pairwise distance increases, indicating a stronger focus on local features. Therefore, multi-head position-attention is trained to provide a balanced view of both local and global aspects of the underlying operator, thereby enhancing the expressive capacity of PiT.

This is corroborated by the results presented in Table \ref{tab:multihead}. Generally, using $2$-head implementation in all position-attention layers leads to lower prediction errors than using single-head. Note that $2$-head implementation leads to a total of $4$ heads for the local position-attention layers in the Encoder and Decoder, and a total of $8$ heads for the global position-attention layers in the Processor. We find using more than $2$ heads in all position-attention layers hardly improve the performance of PiT.
\begin{table}[ht!]
	\caption{PiT's prediction errors in the six benchmarks when using different number of heads in position-attention and its variants.}
	\label{tab:multihead}
	\vskip 0.15in
	\begin{center}
		\resizebox{0.7\textwidth}{!}{
			\begin{tabular}{ccccccc}
				\toprule
				\multicolumn{1}{c}{}       &InviscidBurgers      &ShockTube         &Darcy2D             &Vorticity          &Elasticity         &NACA \\
				\midrule
				\multicolumn{1}{c}{$h=1$}  &$0.998$ 			 &$0.251$           &$0.00627$           &$\mathbf{0.1140}$  &$0.00794$          &${0.00553}$ \\
				\multicolumn{1}{c}{$h=2$}  &$\mathbf{0.0136}$    &$\mathbf{0.0122}$ &$\mathbf{0.00485}$  &$0.1169$           &${0.00691}$          &$\mathbf{0.00480}$  \\
				\multicolumn{1}{c}{$h=4$}  &$0.0162$             &$0.0161$          &$0.00502$           &$0.1205$           &$0.00708$          &$0.00507$  \\
				\multicolumn{1}{c}{$h=8$}  &$0.0157$             &$0.0158$          &$0.00530$           &$0.1186$           &$\mathbf{{0.00649}}$ &$0.00545$  \\
				\bottomrule  
			\end{tabular}
		}
	\end{center}
	\vskip -0.1in
\end{table}

\section{Interpretability of Position-attention}
\label{section:interpretabilit}
The position-attention mechanism is designed to be interpretable, drawing inspiration from the numerical methods for PDEs, as it has been claimed in Section \ref{sec:posatt}. Position-attention shares a similar concept with the upwind scheme and employs a linear convolution, with the kernel exhibiting a strong dependence on local neighbouring regions, resonating with the principle of \emph{domain of dependence} in PDEs and numerical methods. This insight greatly supports the interpretablility of our method from a theoretical point of view.

We take Darcy2D as an example, where $\tan(\lambda)$ has been used in position-attention as stated in \Cref{section:implementation}. \Cref{tab:interpretability} shows the values of $1/\sqrt{\tan(\lambda)}$ of all the attention heads (\textit{i.e.} all the convolutions) in the trained PiT model. These values are indeed interpretable, as most attention at a queried point $x$ is directed towards points $y$ with the distance to $x$ smaller than $1/\sqrt{\tan(\lambda)}$.
%%%%%%%%%%%%%%%%%%%5
\begin{table}[tbhp]
	\caption{Values of $1/\sqrt{\tan(\lambda)}$ in each attention head. These values are taken from the trained PiT model for the Darcy2D benchmark. This model adopts a 2-head implementation for all the attention layers.}
	\label{tab:interpretability}
	\vskip 0.15in
	\begin{center}
	\resizebox{0.4\textwidth}{!}{
		\begin{tabular}{lllll}
			\toprule
			&Layer       &Attention  &Head\;1       &Head\;2 \\
			\midrule
			&Encoder     &LocPosAtt  &$0.0483$    &$0.0155$\\
			&Processor\;1  &PosAtt     &$2.44$      &$0.840$\\
			&Processor\;2  &PosAtt     &$0.232$     &$0.827$\\
			&Processor\;3  &PosAtt     &$0.382$     &$0.0752$\\
			&Processor\;4  &PosAtt     &$0.588$     &$0.167$\\
			&Decoder     &LocPosAtt  &$0.0206$    &$0.0498$\\
			\bottomrule
		\end{tabular}
    }
	\end{center}
	\vskip -0.1in
\end{table}
%%%%%%%%%%%%%%%%%%%%%%
\section{Details of Numerical Experiments}
\label{section:experimental_details}

In this section, we outline the training configurations and the neural network architectures to enable easy reproduction of the reported results by readers. All codes and datasets are available through our GitHub repository.
\subsection{Training Configurations}
As in \citet{li2021fourier}, our experiments are conducted in a train-test setting. We use the mean of the relative $l_2$ error to train and evaluate models on Darcy2D, Vorticity, Elasticity, and NACA, as in \citet{li2021fourier, li2022fourier}. For InviscidBurgers and ShockTube, the mean of the relative $l_1$ error is used for training, while performance is assessed using the median of the relative $l_1$ error on the testing set, following \citet{lanthaler2023nonlinear}. The models are trained using the Adam optimizer and a cosine annealing learning rate scheduler \cite{loshchilov2016sgdr}. The initial learning rate is set to $0.001$, and the training lasts for $500$ epochs. The batch sizes adopted in the experiments are shown in \Cref{tab:batch_size}. All the experiments are performed on an NVIDIA GTX 3090 GPU card.
%%%%%%%%%%%%%%%%%%%%%%%%%%%%%%%%55
\begin{table}[tbhp]
	\caption{Batch size.}
	\label{tab:batch_size}
	\vskip 0.15in
	\begin{center}
	\resizebox{0.8\textwidth}{!}{
		\begin{tabular}{ccccccc}
			\toprule
			\multicolumn{1}{c}{}                    &InviscidBurgers &ShockTube  &Darcy2D  &Vorticity &Elasticity &NACA \\
			\midrule
			\multicolumn{1}{c}{\# Training samples} &$950$ &$1,024$  &$1,024$    &$1,000$  &$1,000$   &$1,000$ \\
			\multicolumn{1}{c}{Batch size}          &$5$             &$8$        &$8$      &$8$       &$10$       &$8$  \\
			\bottomrule
		\end{tabular}
    }
	\end{center}
	\vskip -0.1in
\end{table}
%%%%%%%%%%%%%%%%%%%%%%%%%%%%%%%%
PiT employs an Encoder-Processor-Decoder architecture. Detailed network architectures for the six benchmark problems are presented in Table \ref{tab:PiTdetails}. We begin by introducing some abbreviations representing the basic computational layers:
\begin{itemize}
    \item PosAtt$(w,h)$: A global position-attention layer followed by the GELU activation, where $w$  is the encoding dimension and $h$ is the number of attention heads.
    \item LocPosAtt$(w,h, \text{down (or up)})$: A local position-attention layer followed by the GELU activation, with $w$ as the encoding dimension and $h$ as the number of attention heads. This layer's locality parameter is indicated by a quantile value, defining the compactness of the receptive field. Downsampling or upsampling is integrated with the local position-attention. The quantile value and the latent resolution within PiT are detailed in Table \ref{tab:quantile_all} and Table \ref{tab:downsample}, respectively. 
    \item LINEAR$(w, \text{activate})$: A fully connected layer with $w$ neurons. This layer applies pointwisely to feature vectors. It can be optionally activated by the GELU function.
    \item MLP$(w_1,w_2)$: two stacked Linear layers with respectively $w_1$ and $w_2$ neurons. The first layer is activated by the GELU function. 
\end{itemize}
%%%%%%%%%%%%%%%%%%%%%%%%%%%
\begin{table}[tbhp]
	\caption{Quantile value for each benchmark task. In local position-attention, smaller value means more compact recptive field.}
	\label{tab:quantile_all}
	\vskip 0.15in
	\begin{center}
	\resizebox{0.8\textwidth}{!}{
		\begin{tabular}{ccccccc}
			\toprule
			\multicolumn{1}{c}{}                     &InviscidBurgers &ShockTube &Darcy2D    &Vorticity   &Elasticity &NACA            \\
			\midrule
			\multicolumn{1}{c}{Quantile in Encoder}  &$1\%$             &$4\%$   &$2\%$      &$1\%$       &$2\%$       &$0.5\%$  \\
			\multicolumn{1}{c}{Quantile in Decoder}  &$8\%$             &$2\%$   &$5\%$      &$8\%$       &$2\%$       &$2\%$  \\
			\bottomrule
		\end{tabular}
    }
	\end{center}
	\vskip -0.1in
\end{table}
\begin{table}[tbhp]
	\caption{Input and latent mesh resolutions for each task. (*) The data of Elasticity are sampled on irregular point clouds.}
	\label{tab:downsample}
	\vskip 0.15in
	\begin{center}
	\resizebox{0.8\textwidth}{!}{
		\begin{tabular}{ccccccc}
			\toprule
			\multicolumn{1}{c}{}                    &InviscidBurgers &ShockTube       &Darcy2D  &Vorticity &Elasticity* &NACA \\
			\midrule
			\multicolumn{1}{c}{Input resolution}  &$1,024\times 1$ &$2,048\times 1$ &$211\times 211$      &$64\times 64$       &$972$       &$221\times 51$  \\
			\multicolumn{1}{c}{Latent resolution}   &$1,024\times 1$ &$1,024\times 1$ &$32\times 32$      &$16\times 16$       &$972$       &$111\times 26$  \\
			\bottomrule
		\end{tabular}
    }
	\end{center}
	\vskip -0.1in
\end{table}
%%%%%%%%%%%%%%%%%%%%%%%%%
\begin{table}[tbhp]
	\caption{Details of the PiT architectures for all six benchmarks.}
	\label{tab:PiTdetails}
	\vskip 0.15in
	\begin{center}
	\resizebox{1.0\textwidth}{!}{
		\begin{tabular}{lllllll}
			\toprule
			\multicolumn{1}{c}{}        &InviscidBurgers &ShockTube  &Darcy2D  &Vorticity &Elasticity &NACA \\
			\midrule
			\multicolumn{1}{c}{Encoder} &\begin{tabular}{@{}l@{}}Linear$(64,\text{activate})$\\LocPosAtt$(64,2,\text{down})$\end{tabular} &\begin{tabular}{@{}l@{}}Linear$(64,\text{activate})$\\LocPosAtt$(64,2,\text{down})$\end{tabular} &\begin{tabular}{@{}l@{}}Linear$(128,\text{activate})$\\LocPosAtt$(128,2,\text{down})$\end{tabular}  &\begin{tabular}{@{}l@{}}Linear$(256,\text{activate})$\\LocPosAtt$(256,1,\text{down})$\end{tabular}  &\begin{tabular}{@{}l@{}}Linear$(512)$\\LocPosAtt$(512,{8})$\end{tabular} &\begin{tabular}{@{}l@{}}Linear$(256,\text{activate})$\\LocPosAtt$(256,2,\text{down})$\end{tabular} \\
			\midrule
			\multicolumn{1}{c}{Processor} &\begin{tabular}{@{}l@{}l@{}l@{}l@{}l@{}}$4\times \big [$\\PosAtt$(64,2)$ \\ MLP$(64,64)$\\LINEAR$(64)$\\GELU\\$\big ]$\end{tabular} &\begin{tabular}{@{}l@{}l@{}l@{}l@{}l@{}}$4\times \big [$\\PosAtt$(64,2)$ \\ MLP$(64,64)$\\LINEAR$(64)$\\GELU\\$\big ]$\end{tabular} &\begin{tabular}{@{}l@{}l@{}l@{}l@{}l@{}}$4\times \big [$\\PosAtt$(128,2)$ \\ MLP$(128,128)$\\LINEAR$(128)$\\GELU\\$\big ]$\end{tabular} &\begin{tabular}{@{}l@{}l@{}l@{}l@{}l@{}}$4\times\big [$\\PosAtt$(256,1)$ \\ MLP$(256,256)$\\Linear$(256)$\\GELU\\$\big ]$\end{tabular} &\begin{tabular}{@{}l@{}l@{}l@{}l@{}l@{}}$4\times \big [$\\PosAtt$(512,{8})$ \\ MLP$(512,512)$\\LINEAR$(512)$\\GELU\\$\big ]$\end{tabular} &\begin{tabular}{@{}l@{}l@{}l@{}l@{}l@{}}$4\times \big [$\\PosAtt$(256,2)$ \\ MLP$(256,256)$\\LINEAR$(256)$\\GELU\\$\big ]$\end{tabular} \\
			\midrule
			\multicolumn{1}{c}{Decoder}  &\begin{tabular}{@{}l@{}l@{}@{}l@{}l@{}}LocPosAtt$(64,2,\text{up})$\\\big[PosAtt$(64,2)$\\MLP$(64,64)$\\LINEAR(64)\\GELU\big]\\MLP(64,1)\end{tabular} &\begin{tabular}{@{}l@{}l@{}@{}l@{}l@{}}LocPosAtt$(64,2,\text{up})$\\\big[PosAtt$(64,2)$\\MLP$(64,64)$\\LINEAR(64)\\GELU\big]\\MLP(64,1)\end{tabular} &\begin{tabular}{@{}l@{}}LocPosAtt$(128,2,\text{up})$\\MLP$(128,1)$\end{tabular}  &\begin{tabular}{@{}l@{}}LocPosAtt$(256,1,\text{up})$\\MLP$(256,1)$\end{tabular}  &\begin{tabular}{@{}l@{}}LocPosAtt$(512,{8})$\\MLP$(512,1)$\end{tabular} &\begin{tabular}{@{}l@{}}LocPosAtt$(256,2,\text{up})$\\MLP$(256,1)$\end{tabular} \\
			\midrule
			\multicolumn{1}{c}{\# Parameters} &$95,503$ &$95,631$ &$313,613$ &$1,252,103$ &${6,586,929}$ &$1,250,061$ \\
			\midrule
			\multicolumn{1}{c}{\begin{tabular}{@{}l@{}}Training time\\ (seconds/epoch)\end{tabular}} &$0.938$ &$1.04$ &$14.7$ &$16.3$ &${7.69}$ &$15.3$ \\
			\bottomrule
		\end{tabular}
    }
	\end{center}
	\vskip -0.1in
\end{table}
%%%%%%%%%%%%%%%%%%%%
We present the details of FNO++\footnote{https://github.com/neuraloperator/neuraloperator/tree/master} in Table \ref{tab:FNOdetails}. For datasets on regular grids, FNO++ shows outstanding training speed thanks to the fast discrete Fourier transform.
%%%%%%%%%%%%%%%%%%%%%%%%%%%%
\begin{table}[ht!]
	\caption{Architecture details and training times of FNO++.}
	\label{tab:FNOdetails}
	\vskip 0.15in
	\begin{center}
	\resizebox{0.55\textwidth}{!}{
		\begin{tabular}{ccccc}
			\toprule
			\multicolumn{1}{c}{}        &InviscidBurgers &ShockTube  &Darcy2D  &Vorticity \\
			\midrule
			\multicolumn{1}{c}{Modes}   &$19$ &$7$ &$12$ &$12$ \\
			\multicolumn{1}{c}{Width}   &$32$ &$32$ &$32$ &$20$ \\
			\midrule
			\multicolumn{1}{c}{\# Parameters} &$170,593$ &$72,353$ &$2,376,449$ &$928,661$ \\
			\midrule
			\multicolumn{1}{c}{\begin{tabular}{c}Training time\\(seconds/epoch)\end{tabular}} &$0.434$ &$0.671$ &$5.64$ &$11.5$ \\
			\bottomrule
		\end{tabular}
    }
	\end{center}
	\vskip -0.1in
\end{table}
%%%%%%%%%%%%%%%%%%%%%%%%
\subsection{Insights on Hyper-parameter Calibration}
Tuning the hyper-parameters (quantile, latent resolution $N_v$, and encoding dimension $d_v$) in PiT models is not a difficult process. We recommend beginning with a small quantile value (for instance, 1\%) for the Encoder and Decoder, using a coarse latent mesh, and setting the encoding dimension to 64. These initial settings typically allow a PiT model to deliver comparable performance to baseline models for all our tested cases. 

Should there be a need for further refinement to attain higher accuracy, we proceed with localized tuning. This involves increasing the encoding dimension $d_v$ first. If the desired accuracy is not met through this adjustment, we then refine the latent resolution $N_v$ and, as a final step, modify the quantile. This stepwise approach helps in efficiently reaching the optimal performance without exhaustive search.
%%%%%%%%%%%%%%%%%%%%%%%%%
\subsection{Comparison of Parameter Counts with Baselines Models}
\label{sec:params_counts}
For further benchmark purpose, we provide the parameter count for each of the models in \Cref{tab:main_result}.
\begin{itemize}
    \item For the InviscidBurgers and ShockTube benchmarks, we have gathered parameter counts of DeepONet, shift-DeepONet, FNO and FNO++ in \Cref{tab:count1}:
    %%%%%%%%%%%%%%%%%%%5
    \begin{table}[tbhp]
	    \caption{Parameter counts of PiT and the baseline models in the InviscidBurgers and ShockTube benchmarks. The smallest model in each task is \textbf{bolded}, and the second smallest model is \underline{underlined}.}
	    \label{tab:count1}
	    \vskip 0.15in
	    \begin{center}
	    \resizebox{0.4\textwidth}{!}{
		    \begin{tabular}{llll}
			    \toprule
			    &Model           &InviscidBurgers       &ShockTube  \\
			    \midrule
			    &DeepONet        &$618,085$             &$3,190,673$\\
			    &shift-DeepONet  &$1,835,297$           &$6,047,633$\\
			    &FNO             &$\mathbf{90,593}$     &$\mathbf{41,505}$\\
			    &FNO++           &$170,593$             &$\underline{72,353}$\\
			    \midrule
			    &PiT             &$\underline{95,503}$     &$95,631$\\
			    \bottomrule
		    \end{tabular}
        }
	    \end{center}
	    \vskip -0.1in
    \end{table}
    %%%%%%%%%%%%%%%%%%%%%%
    \item For the Darcy2D and Vorticity benchmarks, we have gathered parameter counts for Galerkin Transformer, OFormer, FNO, and FNO++ in \Cref{tab:count2}:
    %%%%%%%%%%%%%%%%%%%5
    \begin{table}[tbhp]
	    \caption{Parameter counts of PiT and the baseline models in the Darcy2D and Vorticity benchmarks.}
	    \label{tab:count2}
	    \vskip 0.15in
	    \begin{center}
	    \resizebox{0.425\textwidth}{!}{
		    \begin{tabular}{llll}
			    \toprule
			    &Model                  &Darcy2D                            &Vorticity  \\
			    \midrule
			    &Galerkin Transformer   &$\underline{2.22\;\text{Million}}$   &$1.56$ Million\\
			    &OFormer                &$2.51$ Million                     &$1.85$ Million\\
			    &FNO                    &$2,368,001$                        &$\mathbf{926,517}$\\
			    &FNO++                  &$2,376,449$                        &$\underline{928,661}$\\
			    \midrule
			    &PiT                    &$\mathbf{313,613}$                 &$1,252,103$\\
			    \bottomrule
		    \end{tabular}
        }
	    \end{center}
	    \vskip -0.1in
    \end{table}
    %%%%%%%%%%%%%%%%%%%%%%
    \item For the Elasticity and NACA benchmarks, we have gathered the parameter counts for FNO and Geo-FNO in \Cref{tab:count3}. It is worth noting that, for the Elasticity benchmark, we can reduce the encoding dimension of PiT from $512$ to $256$, yielding a model with only $\mathbf{1,655,053}$ parameters (less than those of FNO and Geo-FNO). While this adjustment increases the testing error of PiT from $0.00649$ to $0.00829$ (as shown in \Cref{tab:main_result} and \Cref{fig:hyper_parameter}), it still remains lower than the testing errors of all baseline models. This further demonstrates the efficiency and accuracy of PiT.
    %%%%%%%%%%%%%%%%%%%5
    \begin{table}[tbhp]
	    \caption{Parameter counts of PiT and the baseline models in the Elasticity and NACA benchmarks.}
	    \label{tab:count3}
	    \vskip 0.15in
	    \begin{center}
	    \resizebox{0.345\textwidth}{!}{
		    \begin{tabular}{llll}
			    \toprule
			    &Model                  &Elasticity                  &NACA  \\
			    \midrule
			    &FNO                    &$\mathbf{2,368,001}$        &$\underline{2,368,001}$\\
			    &Geo-FNO                &$\underline{3,020,963}$        &$4,727,329$\\
			    \midrule
			    &PiT                    &$6,586,929$                 &$\mathbf{1,250,061}$\\
			    \bottomrule
		    \end{tabular}
        }
	    \end{center}
	    \vskip -0.1in
    \end{table}
    %%%%%%%%%%%%%%%%%%%%%%
\end{itemize}
%%%%%%%%%%%%%%%%%%%%%%%%%
\subsection{Comparison of Training Costs with Baseline Transformer-based Neural Operators}
\label{sec:runtime}
To validate the superior efficiency of our PiT model over other Transformer architectures, we have acquired the following runtime and memory results:
\begin{itemize}
    \item In the Darcy2D benchmark, \citet{cao2021choose} reported a training time of $0.61$ hours for $100$ epochs using the $211\times 211$ dataset, equating to $22$ seconds/epoch or $5.83$ iterations/second (with $128$ iterations/epoch at a batch size of $8$), with a Galerkin Transformer comprising $2.22$ million parameters. In contrast, our PiT model required only $14.7$ seconds/epoch, with a significantly lower parameter count of $0.31$ million, using the same dataset and GPU.
    \item For the Vorticity test case with $\nu=10^{-4}$ employing $10,000$ training samples, \citet{li2023transformer} detailed the training costs for OFomer and the Galerkin Transformer. Our experiments with PiT, adhering to the same batch size and GPU, demonstrated significantly greater efficiency over these two Transformer architectures, as shown in \Cref{tab:runtime}
    %%%%%%%%%%%%%%%%%%%5
    \begin{table}[tbhp]
	    \caption{Iteration per second and memory usage during model training. The best results are \textbf{bolded}, and the second best results are \underline{underlined}.}
	    \label{tab:runtime}
	    \vskip 0.15in
	    \begin{center}
	    \resizebox{0.6\textwidth}{!}{
		    \begin{tabular}{lllll}
			    \toprule
			    &Model                    &Iters/sec        &Memory (GB)          &params \#(Million)  \\
			    \midrule
			    &Galerkin Transformer     &$1.79$           &$16.65$              &$\underline{1.56}$\\
			    &OFormer                  &$\underline{1.89}$  &$\underline{15.93}$     &$1.85$  \\
			    \midrule
			    &PiT                      &$\mathbf{6.71}$  &$\mathbf{4.4}$       &$\mathbf{1.25}$\\
			    \bottomrule
		    \end{tabular}
        }
	    \end{center}
	    \vskip -0.1in
    \end{table}
    %%%%%%%%%%%%%%%%%%%%%%
    
\end{itemize}

\section{Additional Experimental Results}\label{sec:more}

In this section, we present additional experimental results to support our findings and demonstrate the effectiveness of the proposed PiT for complex operator learning tasks.

\subsection{Additional Experimental Results on Benchmark 1: InviscidBurgers}

\begin{figure}[ht!]
\captionsetup[subfigure]{labelformat=empty}
\centering

\begin{subfigure}[t]{.44\textwidth}
	\centering
	\includegraphics[width=.9\linewidth]{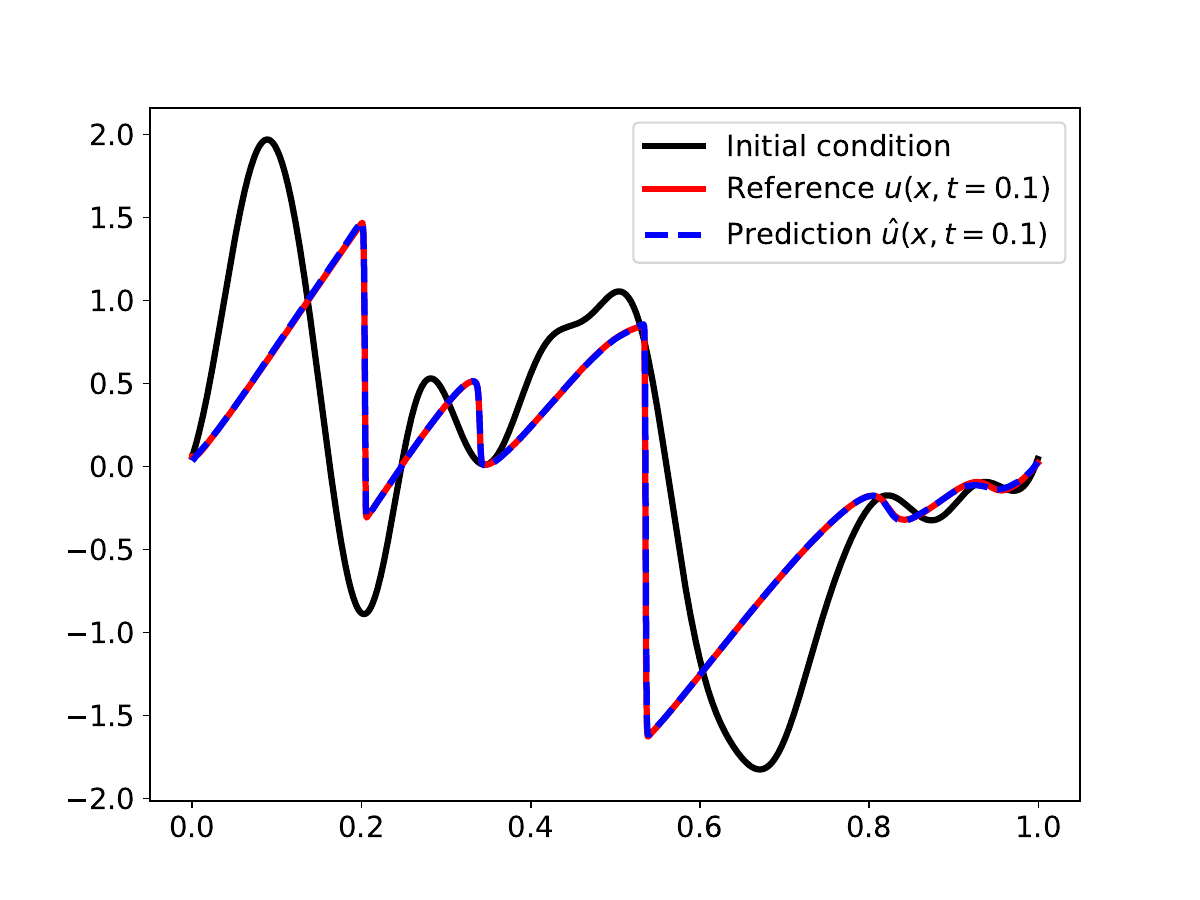}
\end{subfigure}
\begin{subfigure}[t]{.44\textwidth}
	\centering
	\includegraphics[width=.9\linewidth]{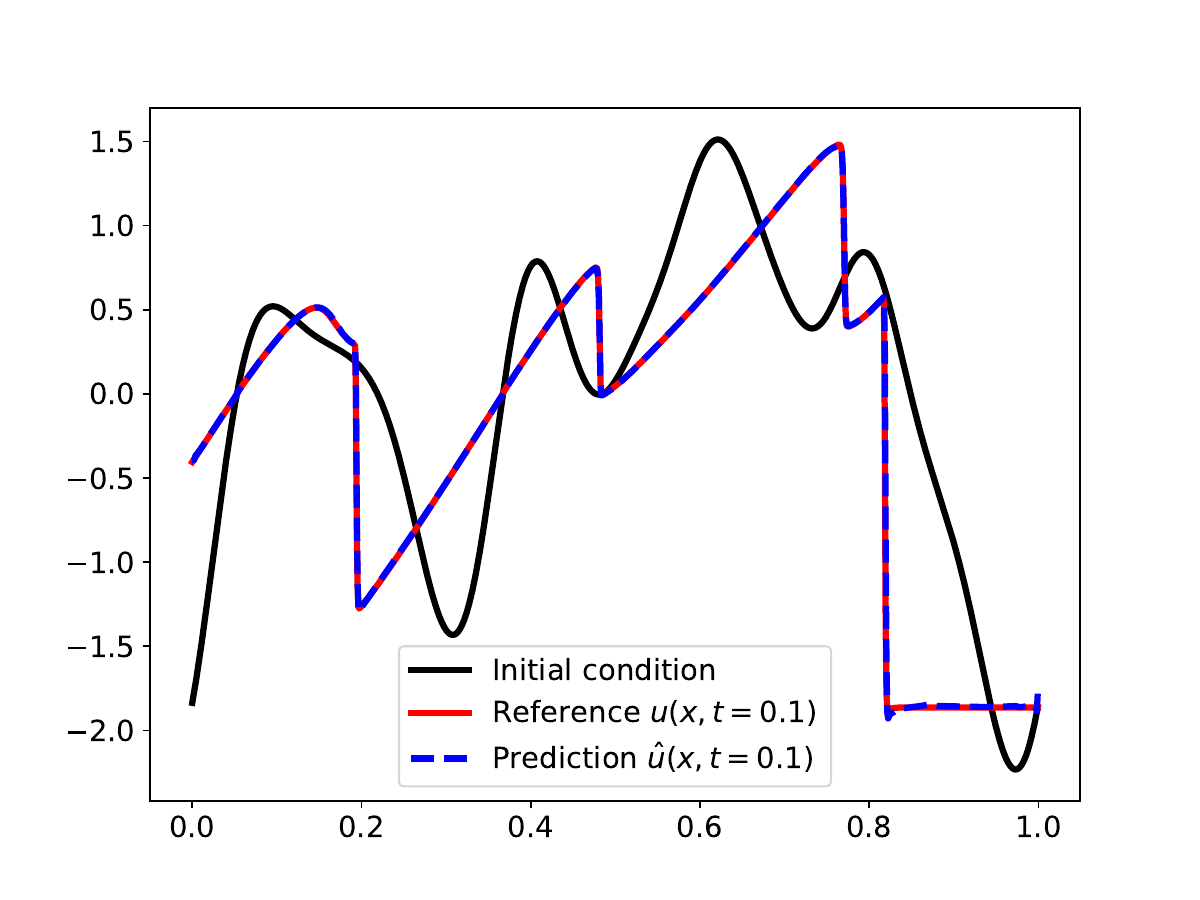}
\end{subfigure}

\smallskip

\begin{subfigure}[t]{.44\textwidth}
	\centering
	\includegraphics[width=.9\linewidth]{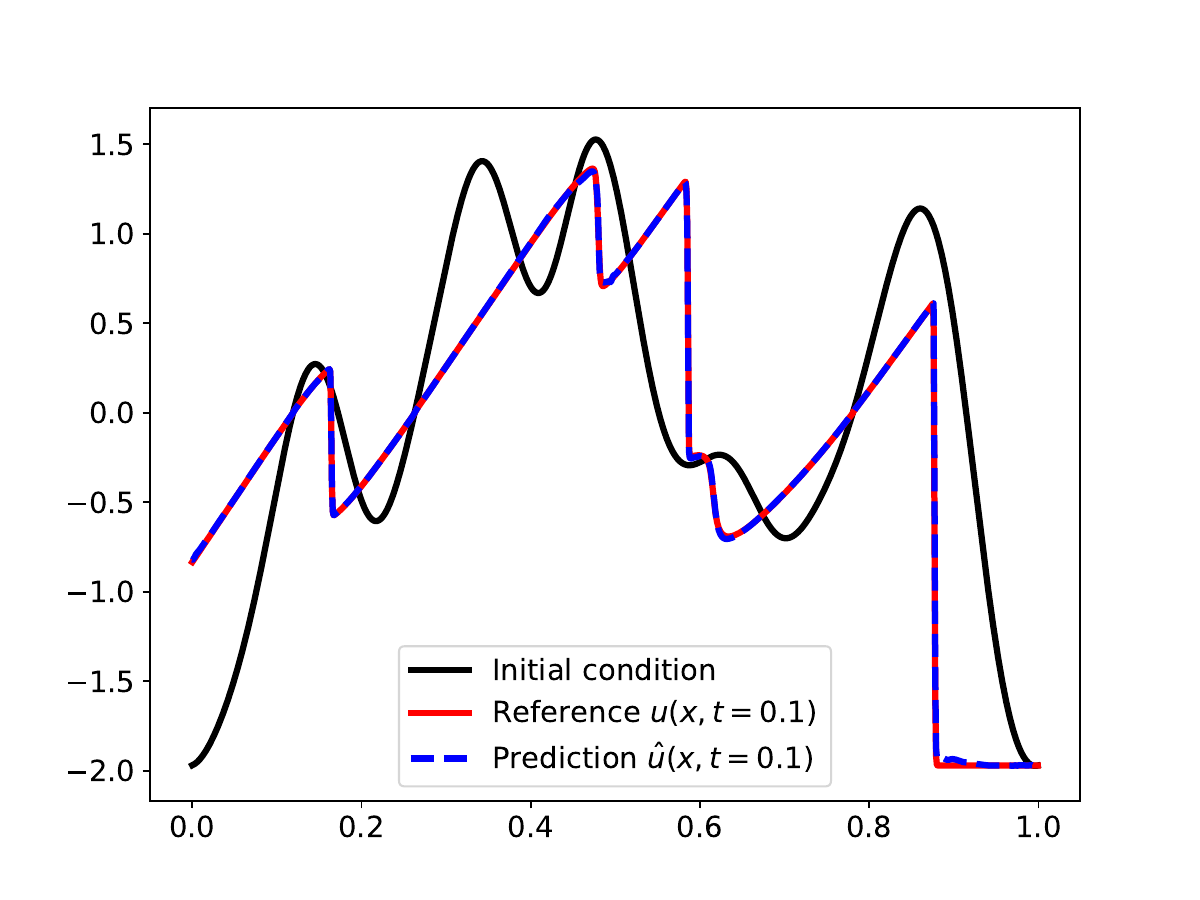}
\end{subfigure}
\begin{subfigure}[t]{.44\textwidth}
	\centering
	\includegraphics[width=.9\linewidth]{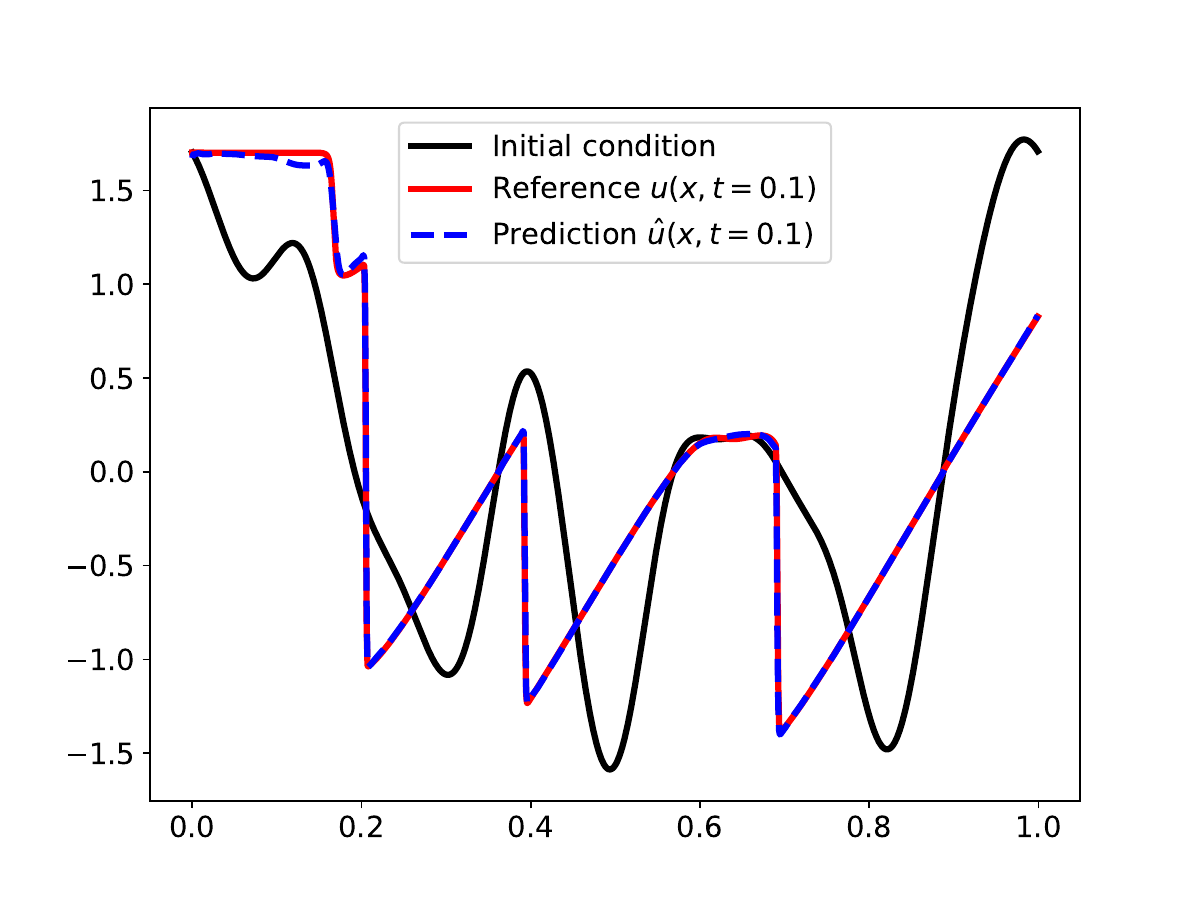}
\end{subfigure}

\caption{InviscidBurgers: Predictions given by Self-PiT for four different input functions. } 
\label{fig:Burgers}
\end{figure}

In Figure \ref{fig:Burgers}, we present some predicted solutions of the inviscid Burgers' equation obtained using the Self-PiT. Due to the nonlinear hyperbolic nature of the PDE, many discontinuities are developed in the solutions, even though the initial conditions are smooth.  
We observe an excellent agreement between the predicted and reference solutions.

In Table \ref{tab:2575}, we display the $25\%$ and $75\%$ quantiles of the relative $l_1$ errors computed over the testing data for PiT compared with the baselines. It is evident that PiT and Self-PiT achieve outstanding results compared to the baselines, although Self-PiT does not consistently outperform PiT. 

\begin{table}[ht!]
	\caption{$25\%$ and $75\%$ quantiles of relative $l_1$ errors ($\times 10^{-2}$) on the testing dataset.}
	\label{tab:2575}
	\vskip 0.15in
	\begin{center}
	\resizebox{0.8\textwidth}{!}{
		\begin{tabular}{ccccccc}
			\toprule
			\multicolumn{1}{c}{}                    &DeepONet       &shift-DeepONet  &FNO         &FNO++            &PiT                    &Self-PiT \\
			\midrule
			\multicolumn{1}{c}{InviscidBurgers}     &$25.4-32.4$    &$6.7-9.6$       &$1.3-1.9$   &$0.842-1.26$    &$1.1-1.61$            &$\mathbf{0.651-0.974}$\\
			\midrule
			\multicolumn{1}{c}{ShockTube}           &$3.4-5.4$      &$2.0-3.75$      &$1.2-2.1$   &$1.48-2.68$     &$\mathbf{0.9-1.82}$  &$1.4-2.64$ \\
			\bottomrule
		\end{tabular}
    }
	\end{center}
	\vskip -0.1in
\end{table}
%%%%%%%%%%%%
%%%%%%%%%%%%%%%%%%%%%%%%%%%%%%%%%%%%%
\subsection{Additional Experimental Results on Benchmark 2: ShockTube}
\begin{figure}[ht!]
\centering
\begin{subfigure}[c]{.44\textwidth}
	\centering
	\includegraphics[width=1.0\linewidth]{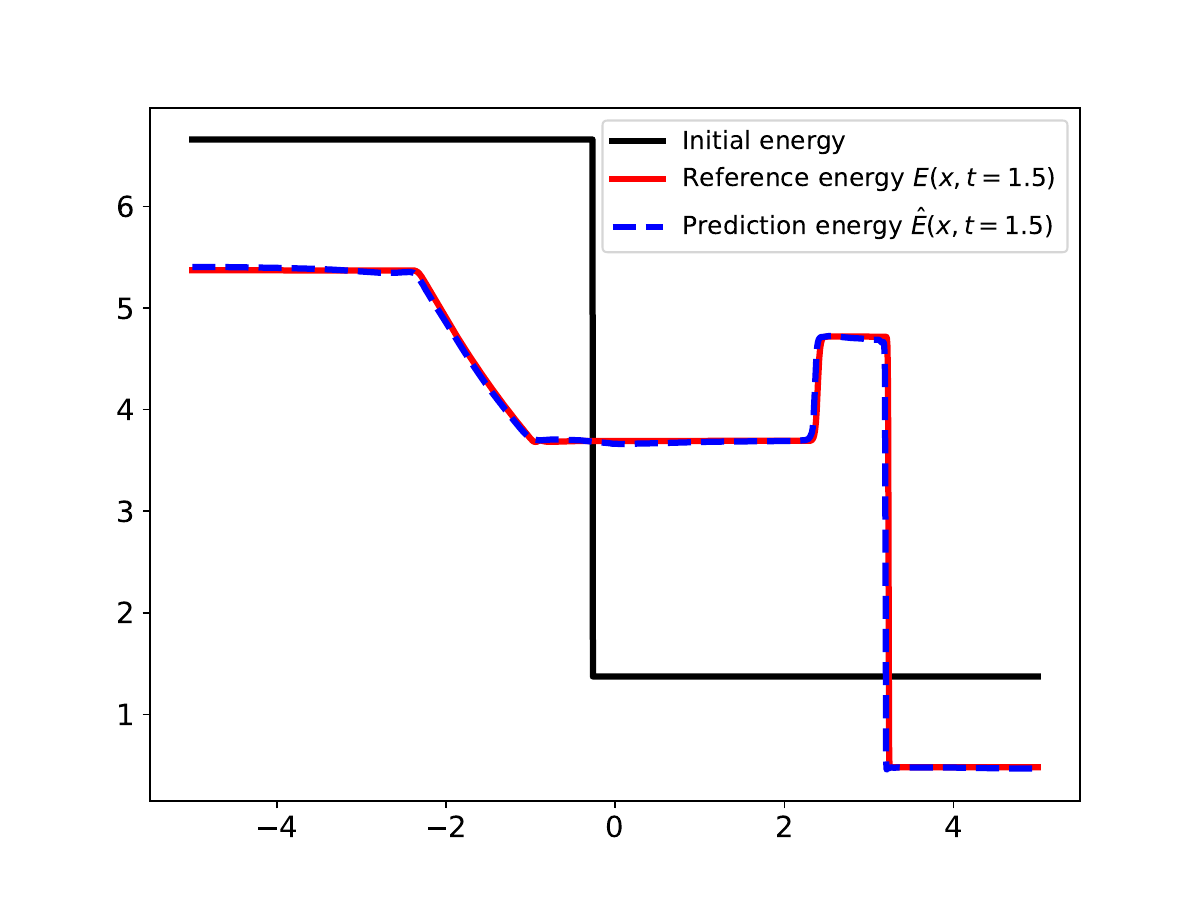}
\end{subfigure}
\begin{subfigure}[c]{.44\textwidth}
	\centering
	\includegraphics[width=1.0\linewidth]{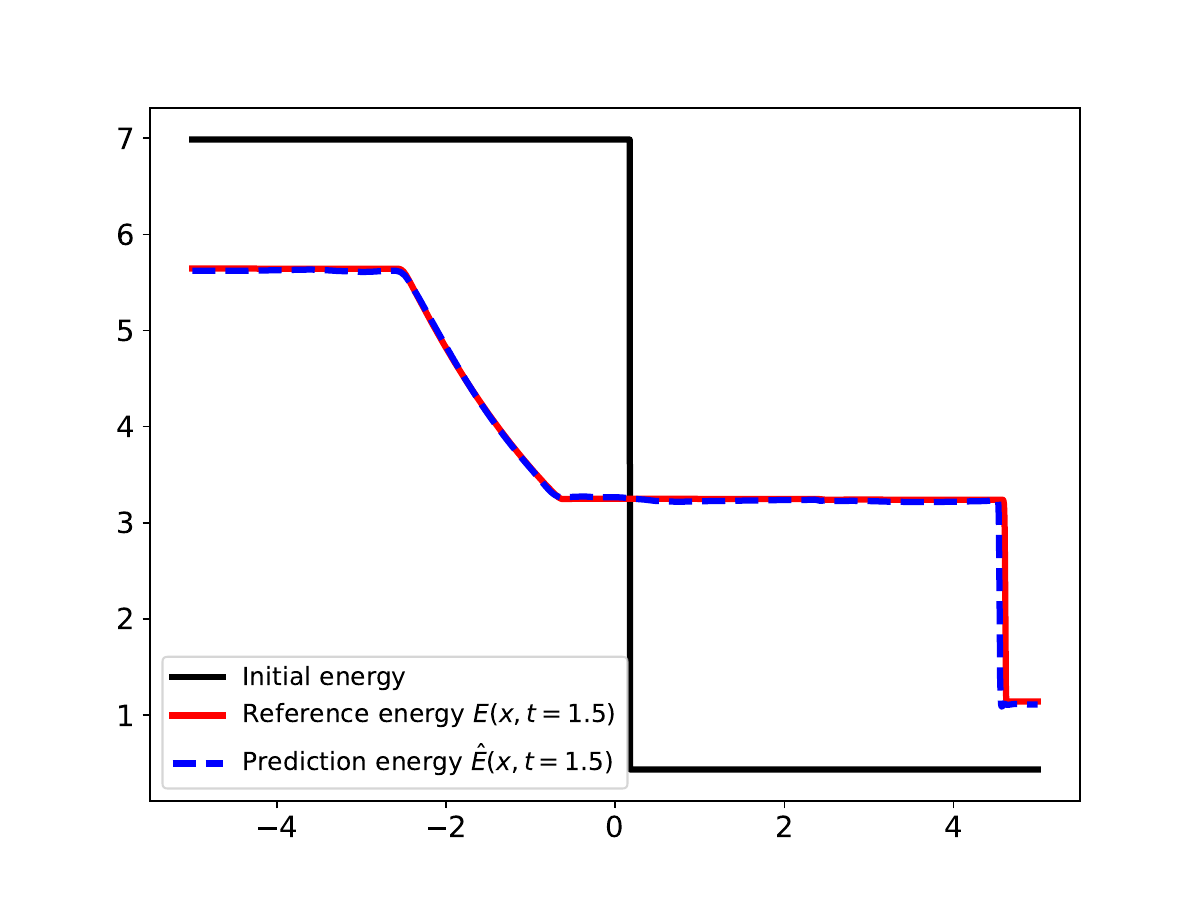}
\end{subfigure}
\smallskip

\begin{subfigure}[c]{.44\textwidth}
	\centering
	\includegraphics[width=1.0\linewidth]{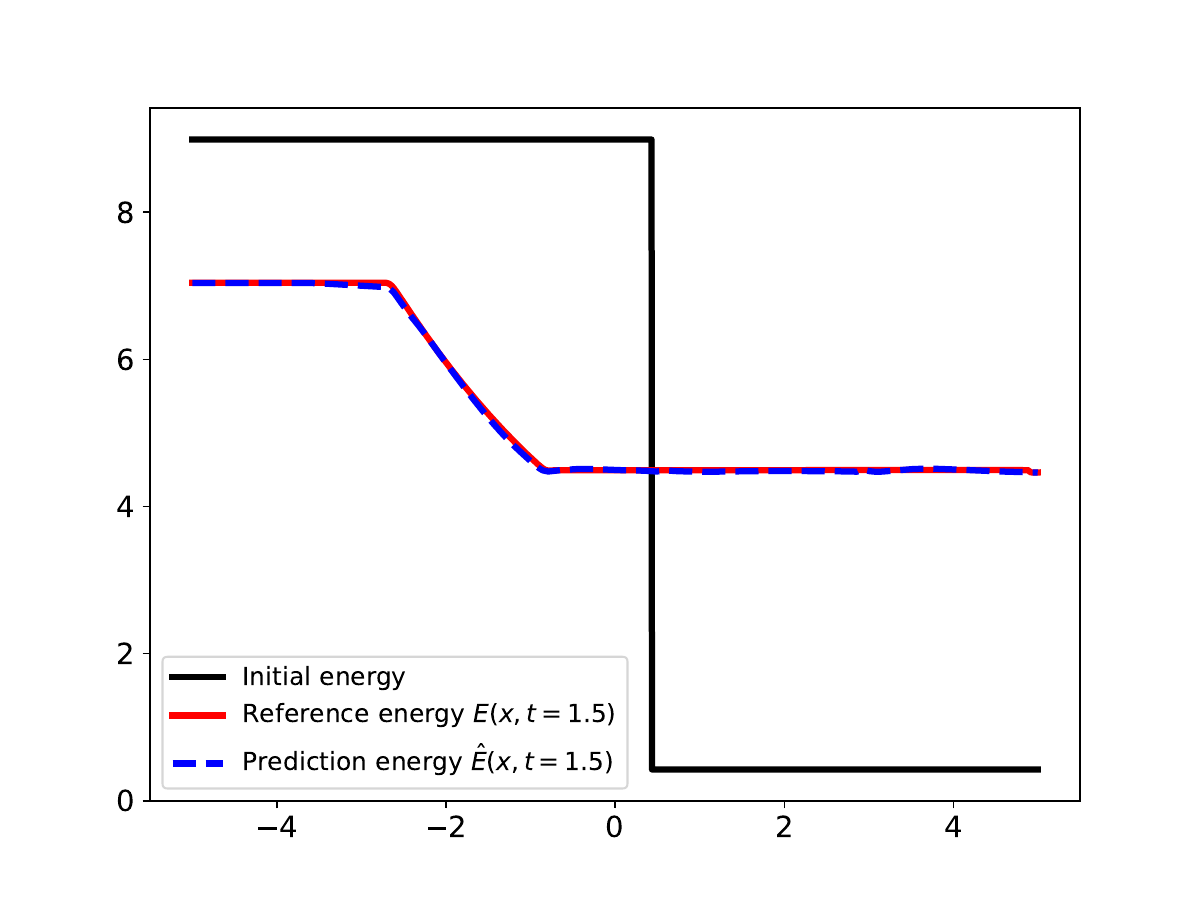}
\end{subfigure}
\begin{subfigure}[c]{.44\textwidth}
	\centering
	\includegraphics[width=1.0\linewidth]{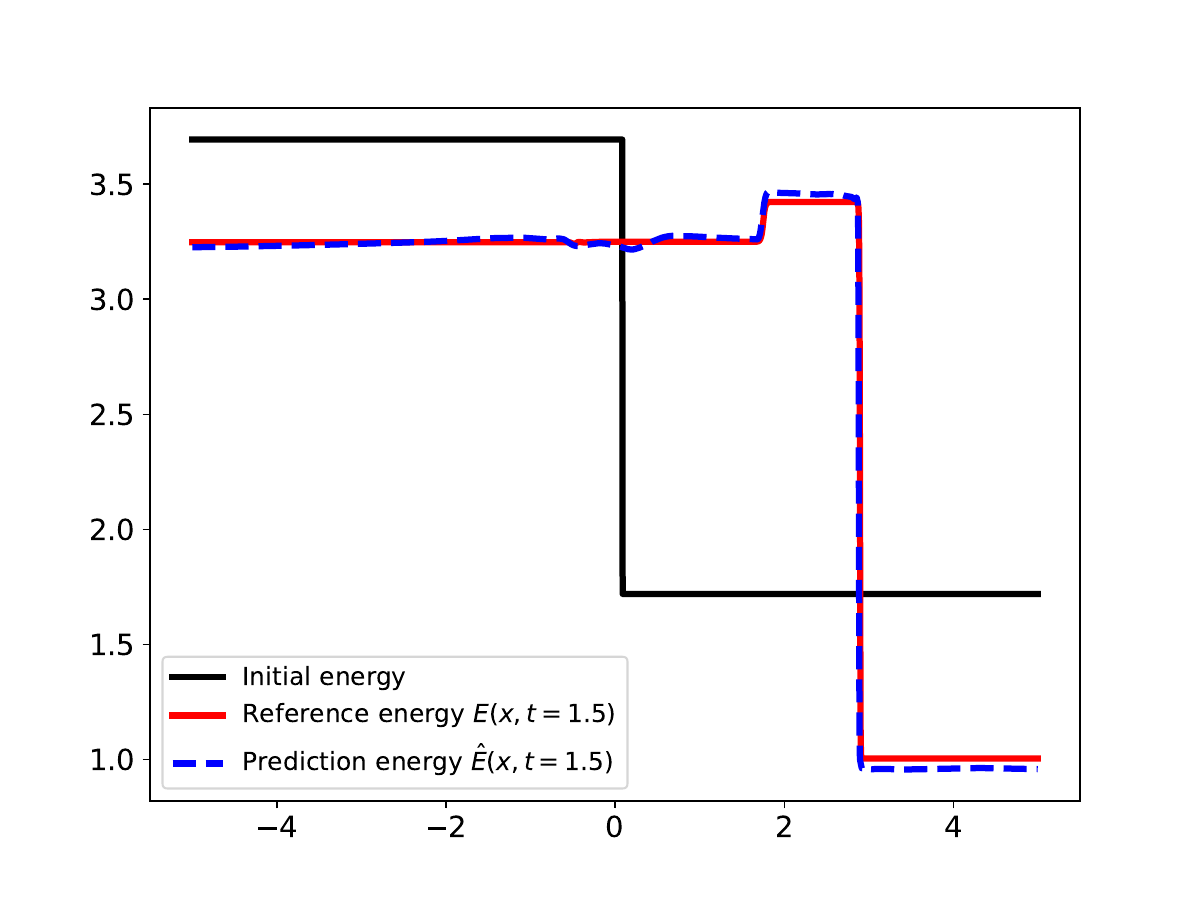}
\end{subfigure}
\caption{ShockTube: Predictions of the total energy functions at $t=1.5$ given by PiT for four different input functions, compared to the reference solutions. The initial conditions for density, momentum and energy are all step functions.}
\label{fig:ShockTube}
\end{figure}

In this operator learning task, the initial density, momentum and energy are step-shaped functions. The 1D compressible Euler equations evolve these initial conditions into various wave structures with discontinuities at $t=1.5$ (see Figure \ref{fig:ShockTube}). 
The results predicted by PiT show good agreement with the reference solutions.

In Table \ref{tab:2575}, we present the 25\% and 75\% quantiles of the relative $l_1$ errors  over the testing data for PiT in comparison with the baselines. For this benchmark, PiT achieves the lowest prediction error, surpassing all tested baselines.

%%%%%%%%%%%%%%%%%%%%%%%%%%%%%%%%%%%%%
\subsection{Additional Experimental Results on Benchmark 3: Darcy2D}
\label{section:sublinearly}
We have shown that PiT demonstrates exceptional efficiency in approximating the solution operator for the Darcy2D benchmark. Remarkably, a PiT trained on downscaled $43^2$ data can accurately predict solutions on the full $421^2$ grid, achieving only a 4.50\% relative error. 

\begin{figure}[ht!]
	\centering
	\includegraphics[width=1.0\linewidth]{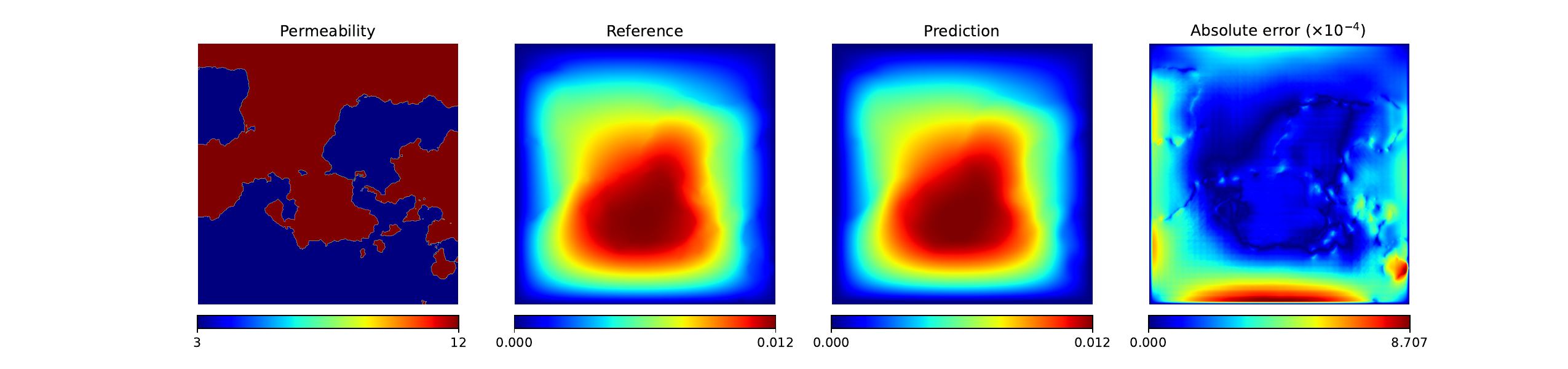}
	\includegraphics[width=1.0\linewidth]{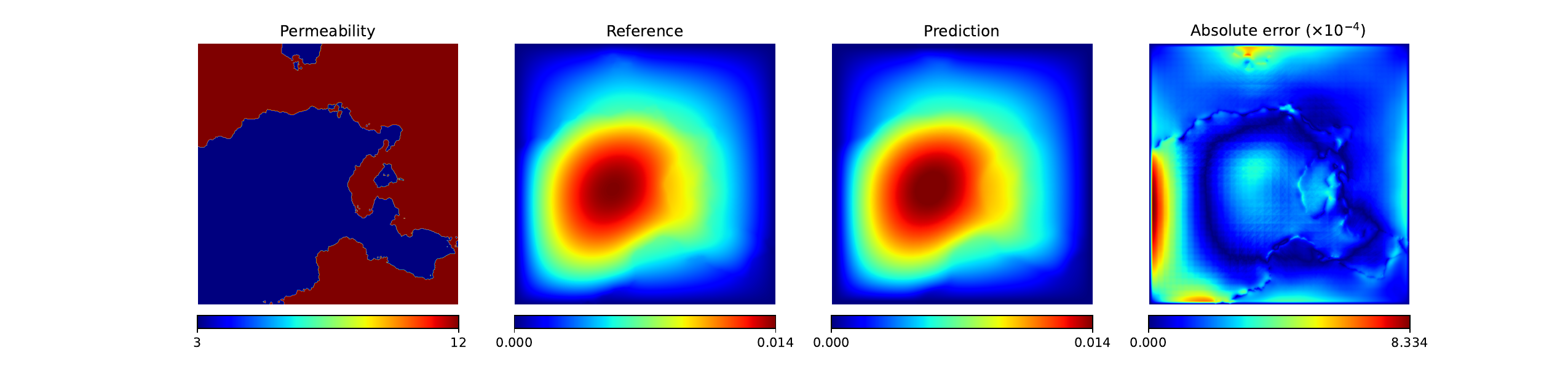}
	\includegraphics[width=1.0\linewidth]{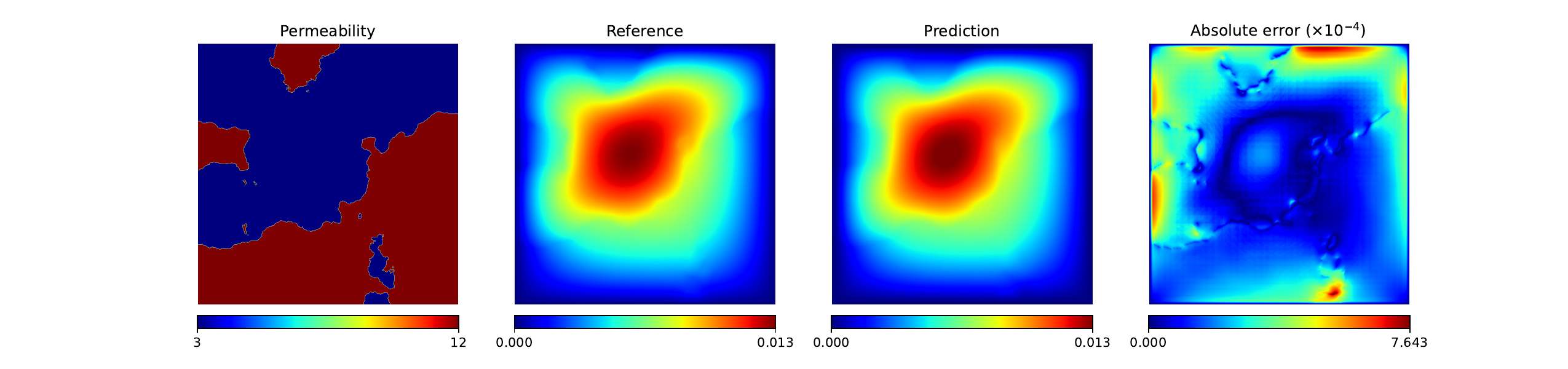}
	\includegraphics[width=1.0\linewidth]{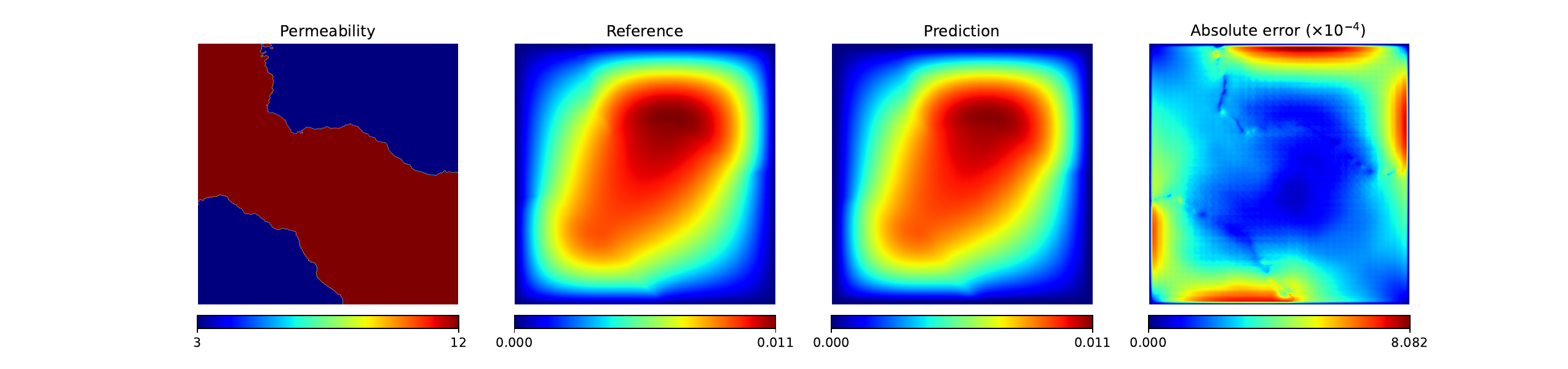}
	\caption{Darcy2D: The input functions (permeability), the referential and predicted pressure fields, and the absolute error (from left to right). The model is trained on $43^2$ mesh resolution, and then tested on $421^2$ mesh resolution.  From top to bottom: four testing examples with different input functions.} 
	\label{fig:Darcy2D}
\end{figure}

The latent resolution of PiT is fixed at $32^2$, regardless of the mesh resolution of the input functions. When trained with data at a finer resolution, PiT shows enhanced performance. We document the prediction error and the training cost for different datasets in Table \ref{tab:Darcy2Dmore}. Notably, the training time per epoch scales sub-linearly with the number of mesh points, denoted by $N$.  Furthermore, the prediction error rapidly reduces as $N$ increases. These results highlight PiT's effectiveness in learning large-scale operators and its remarkable discretization-convergent property under zero-shot super-resolution evaluation (see Figure \ref{fig:Darcy2D}).

\begin{table}[ht!]
	\caption{Darcy2D: Results of discretization-convergent experiments. Four PiT models are trained with data on different mesh resolutions, and then evaluated using the testing data on either the same mesh or the finer $421^2$ mesh. The prediction errors are measured by average relative $l_2$ errors.}
	\label{tab:Darcy2Dmore}
	\vskip 0.15in
	\begin{center}
	\resizebox{0.55\textwidth}{!}{
		\begin{tabular}{ccccccc}
			\toprule
			\multicolumn{1}{c}{Training resolution}            &$43^2$  &$85^2$ &$141^2$ &$211^2$ \\
			\midrule
			\multicolumn{1}{c}{\begin{tabular}{c}Training time\\(seconds/epoch)\end{tabular}} &$1.11$ &$2.41$  &$4.37$     &$14.7$ \\
			\midrule
			\multicolumn{1}{c}{\begin{tabular}{c}Prediction error on \\training resolution\end{tabular}} &$0.00974$ &$0.00578$  &$0.00558$     &$0.00485$ \\
			\midrule
			\multicolumn{1}{c}{\begin{tabular}{c}Prediction error on \\$421\times421$ resolution\end{tabular}}  &$0.0450$      &$0.0209$     &$0.0117$   &$0.00715$\\
			\bottomrule
		\end{tabular}
    }
	\end{center}
	\vskip -0.1in
\end{table}

\subsection{More Experimental Results on Benchmark 4:  Vorticity}
\begin{figure}[ht!]
\centering
    \hspace{-30pt}\includegraphics[width=0.45\linewidth]{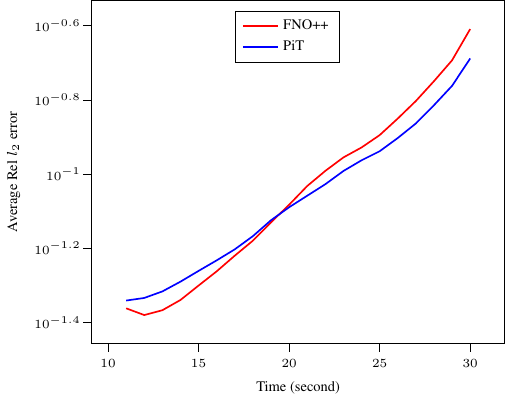}
    \caption{Vorticity: Evolution of prediction errors over time.}
\label{fig:Vorticity_error}
\end{figure}

\begin{figure}[ht!]
\centering
\begin{subfigure}[c]{.3\textwidth}
	\centering
	\hspace{-3pt}\raisebox{0mm}{\includegraphics{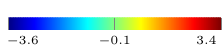}}
%	\raisebox{0mm}{\colorbarc{-3.612948}{3.4198112}{3.6}}
\end{subfigure}
\begin{subfigure}[c]{.3\textwidth}
	\centering
	\hspace{-3pt}\raisebox{0mm}{\includegraphics{Figures/vorticity.pdf}}
%	\raisebox{0mm}{\colorbarc{-3.612948}{3.4198112}{3.6}}
\end{subfigure}
\begin{subfigure}[c]{.3\textwidth}
	\centering
	\hspace{1pt}\raisebox{0mm}{\includegraphics{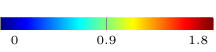}}
%	\raisebox{0mm}{\colorbarc{0}{1.7799222}{3.6}}
\end{subfigure}

\smallskip

\begin{subfigure}[c]{.3\textwidth}
	\centering
	\includegraphics[width=0.7\linewidth]{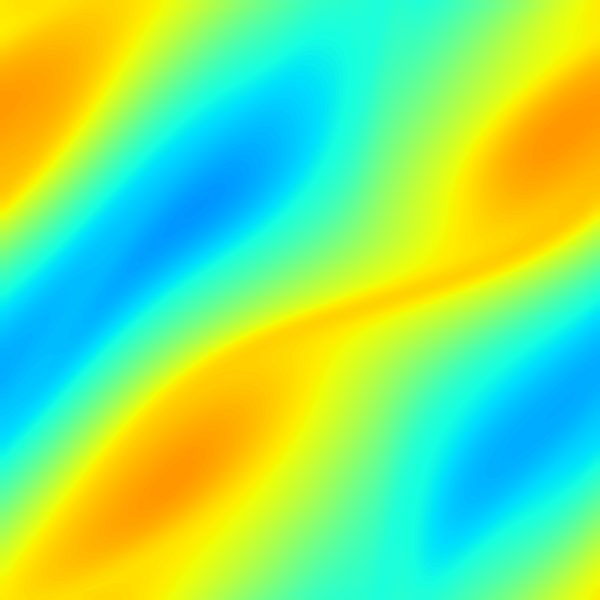}
\end{subfigure}
\begin{subfigure}[c]{.3\textwidth}
	\centering
	\includegraphics[width=0.7\linewidth]{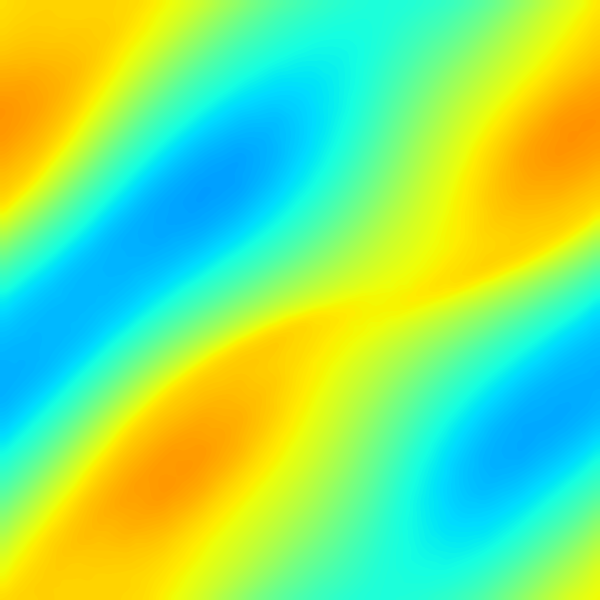}
\end{subfigure}
\begin{subfigure}[c]{.3\textwidth}
	\centering
	\includegraphics[width=0.7\linewidth]{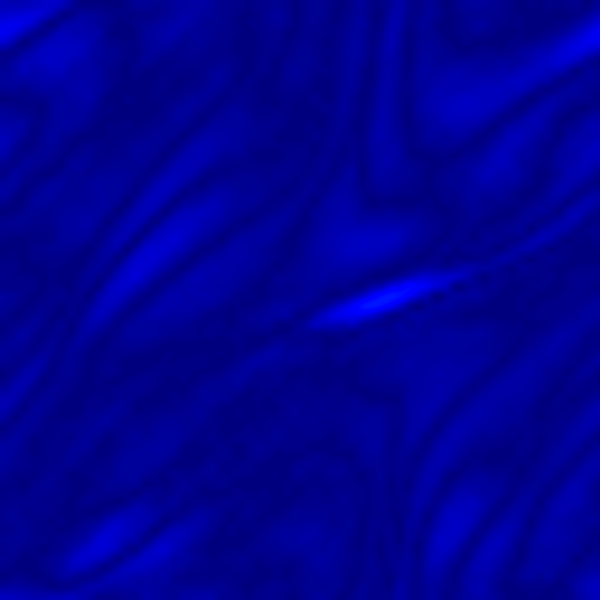}
\end{subfigure}

\smallskip

\begin{subfigure}[c]{.3\textwidth}
	\centering
	\includegraphics[width=0.7\linewidth]{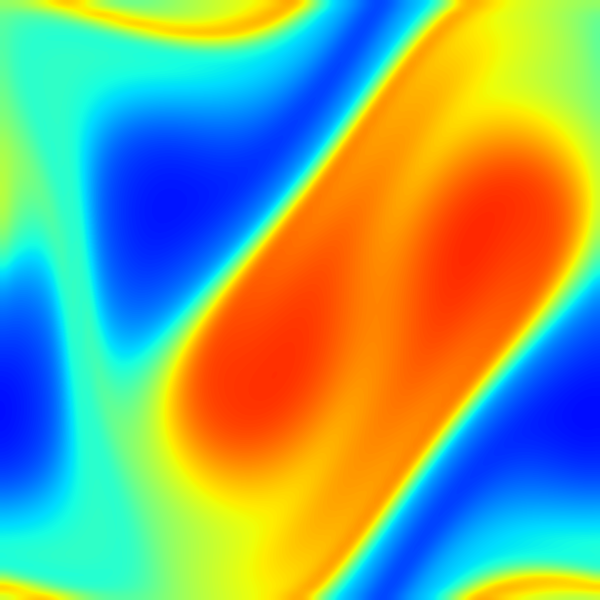}
\end{subfigure}
\begin{subfigure}[c]{.3\textwidth}
	\centering
	\includegraphics[width=0.7\linewidth]{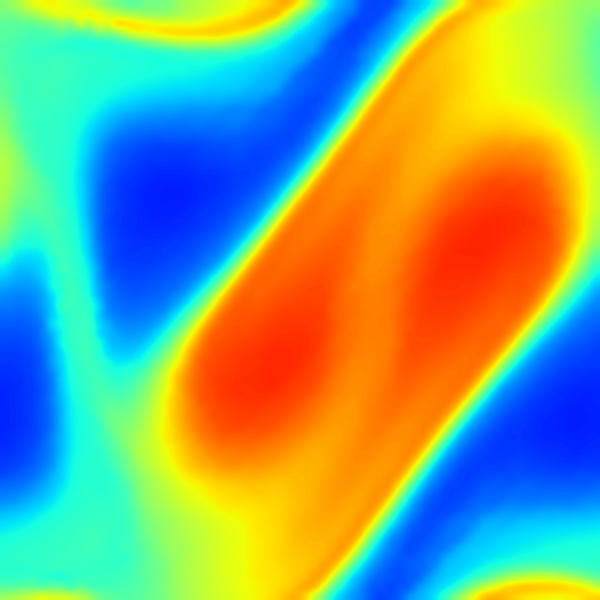}
\end{subfigure}
\begin{subfigure}[c]{.3\textwidth}
	\centering
	\includegraphics[width=0.7\linewidth]{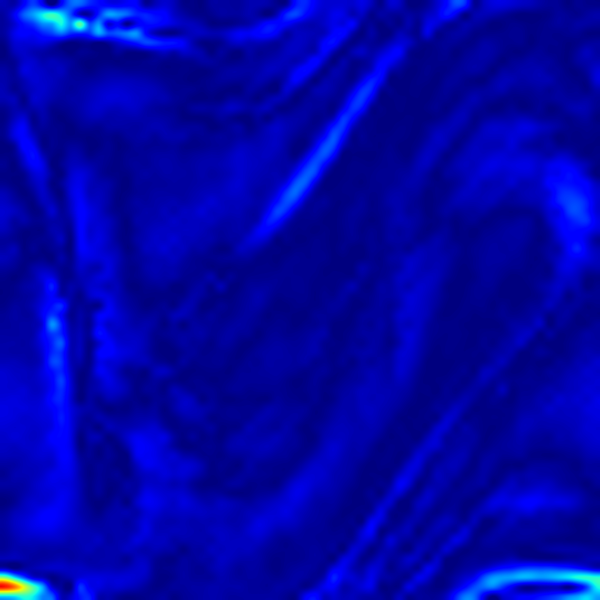}
\end{subfigure}

\smallskip

\begin{subfigure}[c]{.3\textwidth}
	\centering
	\includegraphics[width=0.7\linewidth]{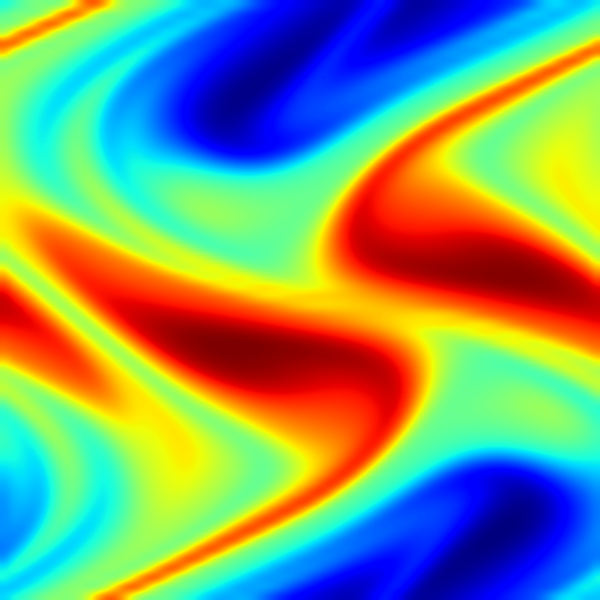}
\end{subfigure}
\begin{subfigure}[c]{.3\textwidth}
	\centering
	\includegraphics[width=0.7\linewidth]{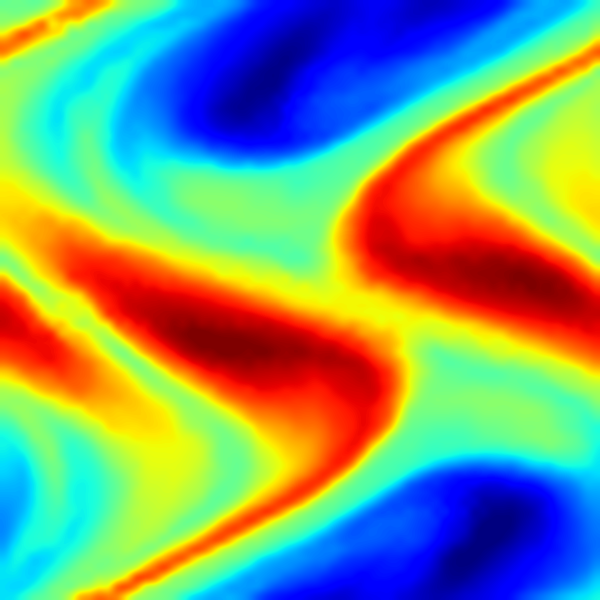}
\end{subfigure}
\begin{subfigure}[c]{.3\textwidth}
	\centering
	\includegraphics[width=0.7\linewidth]{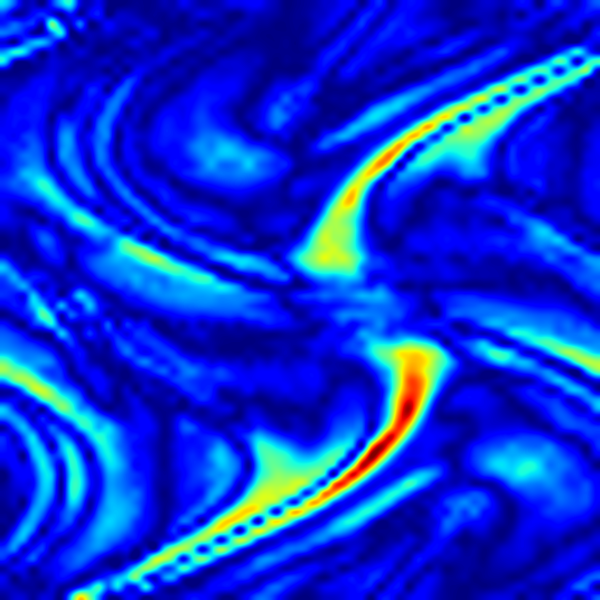}
\end{subfigure}
\caption{Vorticity benchmark: Evolution of the vorticity field at $t=11$, $20$, and $30$ (from top to bottom). Left: reference. Middle: prediction. Right: absolute error.}
\label{fig:Vorticity_fig}
\end{figure}

In Figure \ref{fig:Vorticity_error}, the growth of prediction error is plotted. We observe approximately an exponential trend, which is normal for data-driven evolution operators. In Figure \ref{fig:Vorticity_fig}, we present the evolution of the vorticity field. Although Vorticity is a hard task with turbulent flow pattern and scarce data, our method correctly captures the evolution pattern.

%%%%%%%%%%%%%%%%%%%%%%%%%%%%%%%%%%%%%%%%%%%

\subsection{Additional Experimental Results on Benchmark 5: Elasticity}
\begin{figure}[ht!]
\centering
\begin{subfigure}[c]{.05\textwidth}
	\centering
	\raisebox{0mm}{\includegraphics{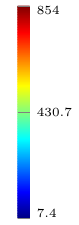}}
%	\raisebox{0mm}{\colorbard{7.35}{854}{3.6}}
\end{subfigure}\quad
\begin{subfigure}[c]{.25\textwidth}
	\centering
	\includegraphics[width=.9\linewidth]{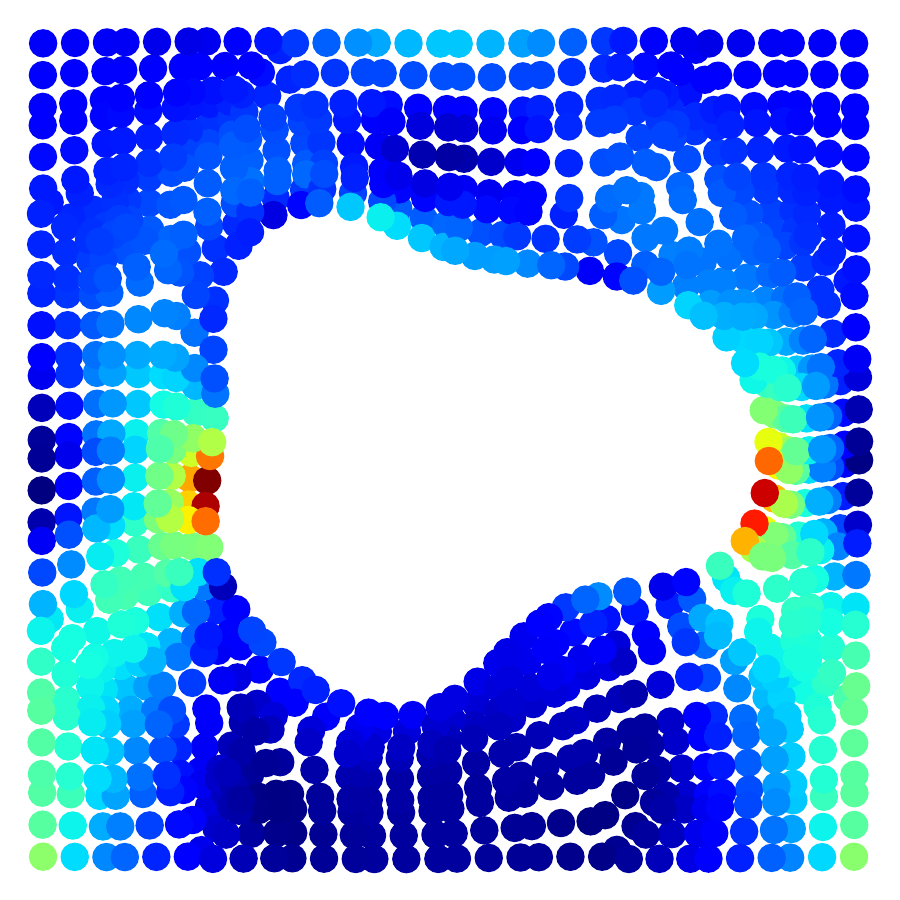}
\end{subfigure}
\begin{subfigure}[c]{.25\textwidth}
	\centering
	\includegraphics[width=.9\linewidth]{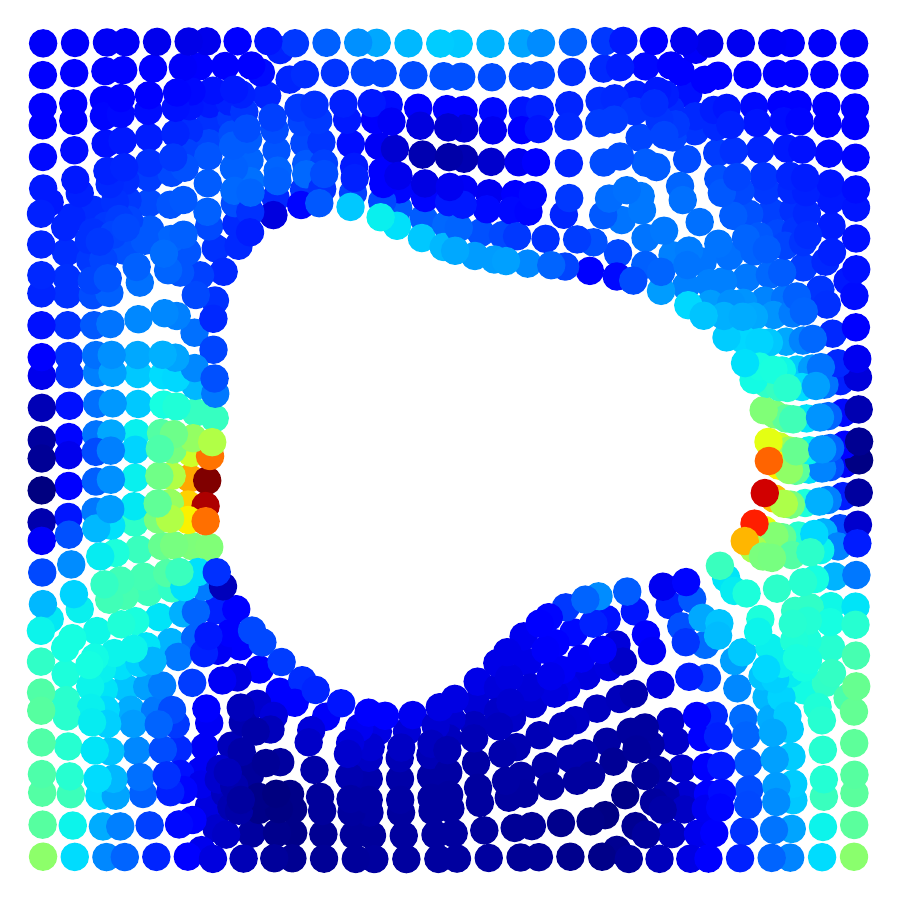}
\end{subfigure}
\begin{subfigure}[c]{.05\textwidth}
	\centering
	\raisebox{0mm}{\includegraphics{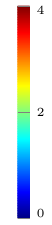}}
%	\raisebox{0mm}{\colorbard{0}{6.62}{3.6}}
\end{subfigure}
\begin{subfigure}[c]{.25\textwidth}
	\centering
	\includegraphics[width=.9\linewidth]{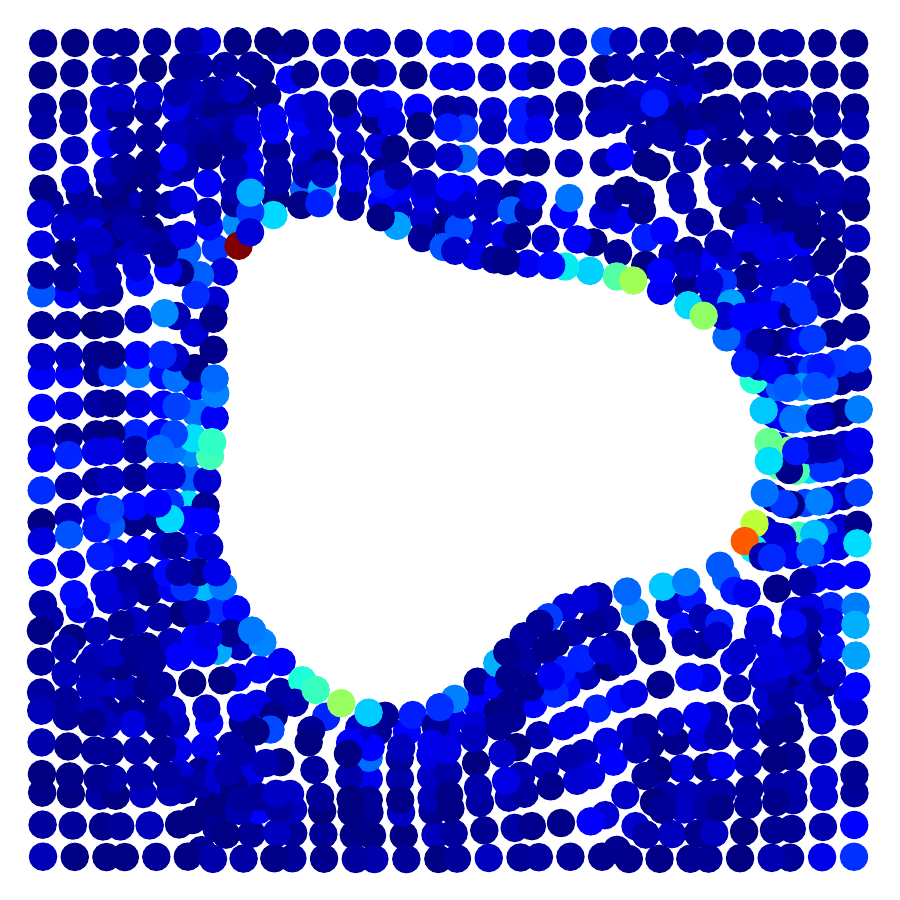}
\end{subfigure}

\smallskip

\begin{subfigure}[c]{.05\textwidth}
	\centering
	\raisebox{0mm}{\includegraphics{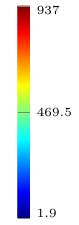}}
%	\raisebox{0mm}{\colorbard{1.94}{937}{3.6}}
\end{subfigure}\quad
\begin{subfigure}[c]{.25\textwidth}
	\centering
	\includegraphics[width=.9\linewidth]{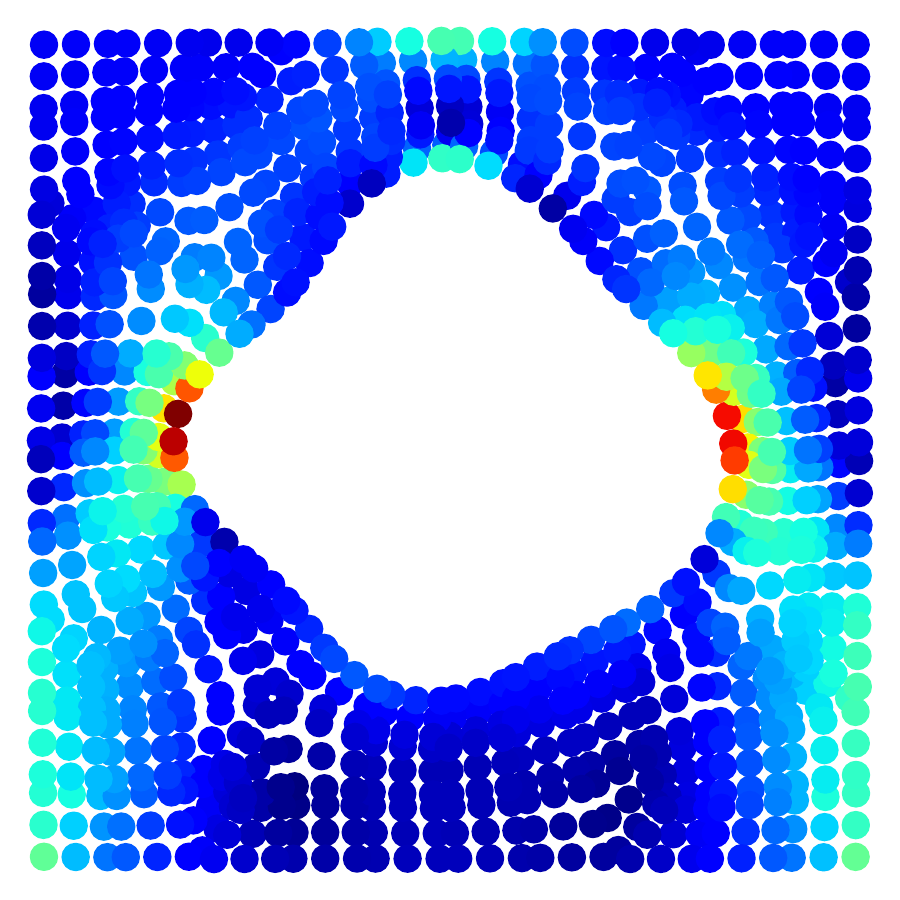}
\end{subfigure}
\begin{subfigure}[c]{.25\textwidth}
	\centering
	\includegraphics[width=.9\linewidth]{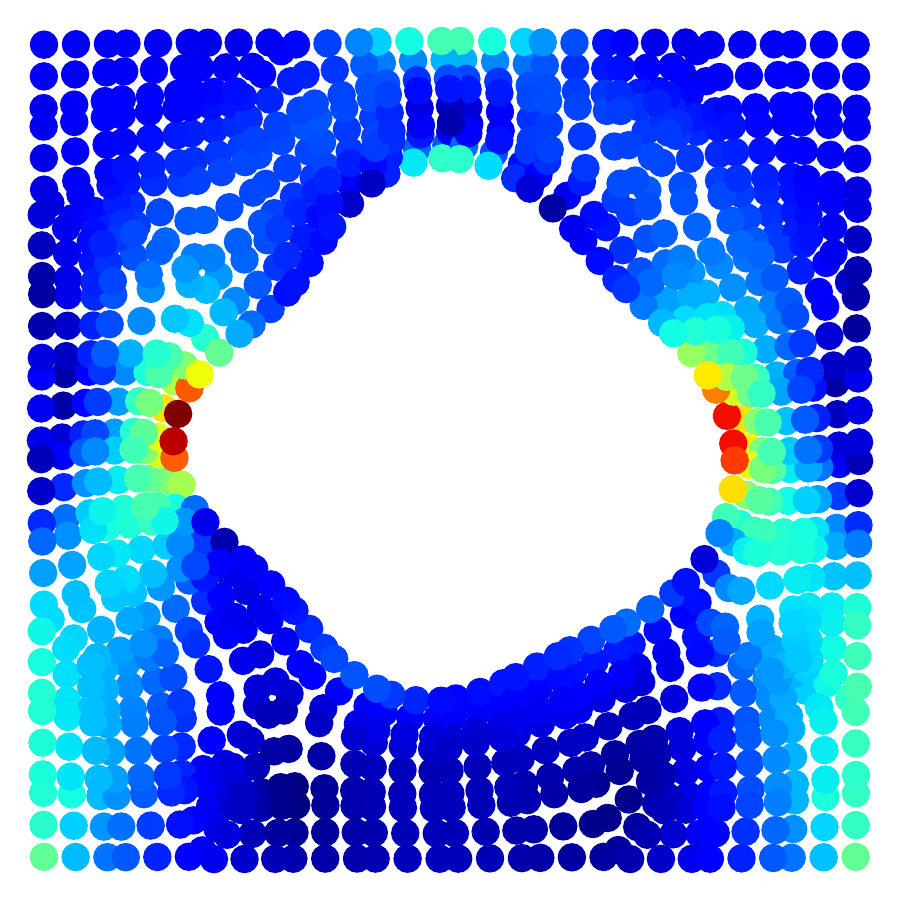}
\end{subfigure}
\begin{subfigure}[c]{.05\textwidth}
	\centering
	\raisebox{0mm}{\includegraphics{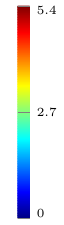}}
%	\raisebox{0mm}{\colorbard{0}{3.65}{3.6}}
\end{subfigure}
\begin{subfigure}[c]{.25\textwidth}
	\centering
	\includegraphics[width=.9\linewidth]{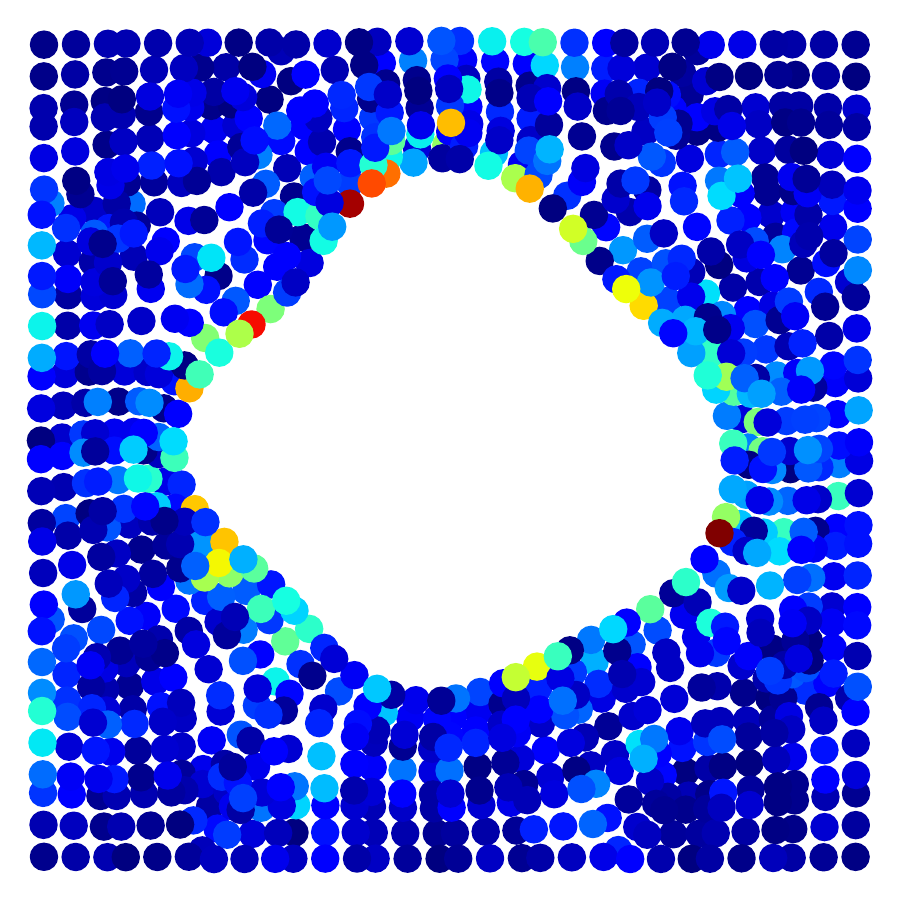}
\end{subfigure}

\smallskip

\begin{subfigure}[c]{.05\textwidth}
	\centering
	\raisebox{0mm}{\includegraphics{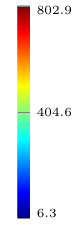}}
%	\raisebox{0mm}{\colorbard{1.94}{937}{3.6}}
\end{subfigure}\quad
\begin{subfigure}[c]{.25\textwidth}
	\centering
	\includegraphics[width=.9\linewidth]{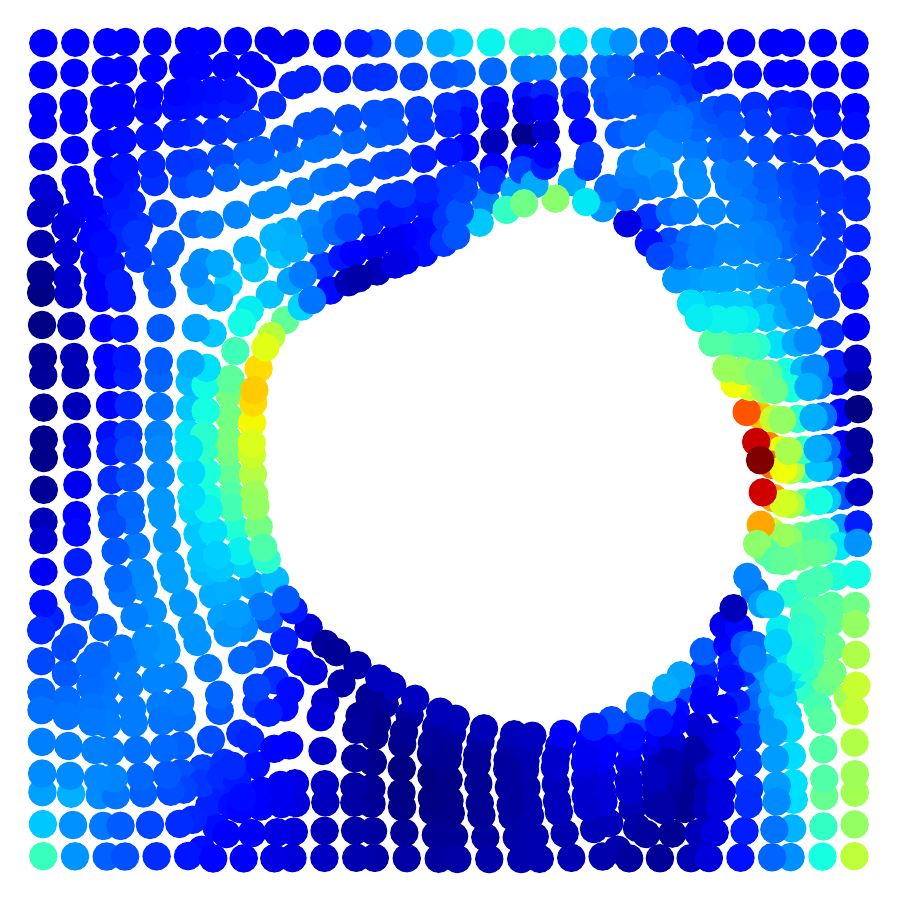}
\end{subfigure}
\begin{subfigure}[c]{.25\textwidth}
	\centering
	\includegraphics[width=.9\linewidth]{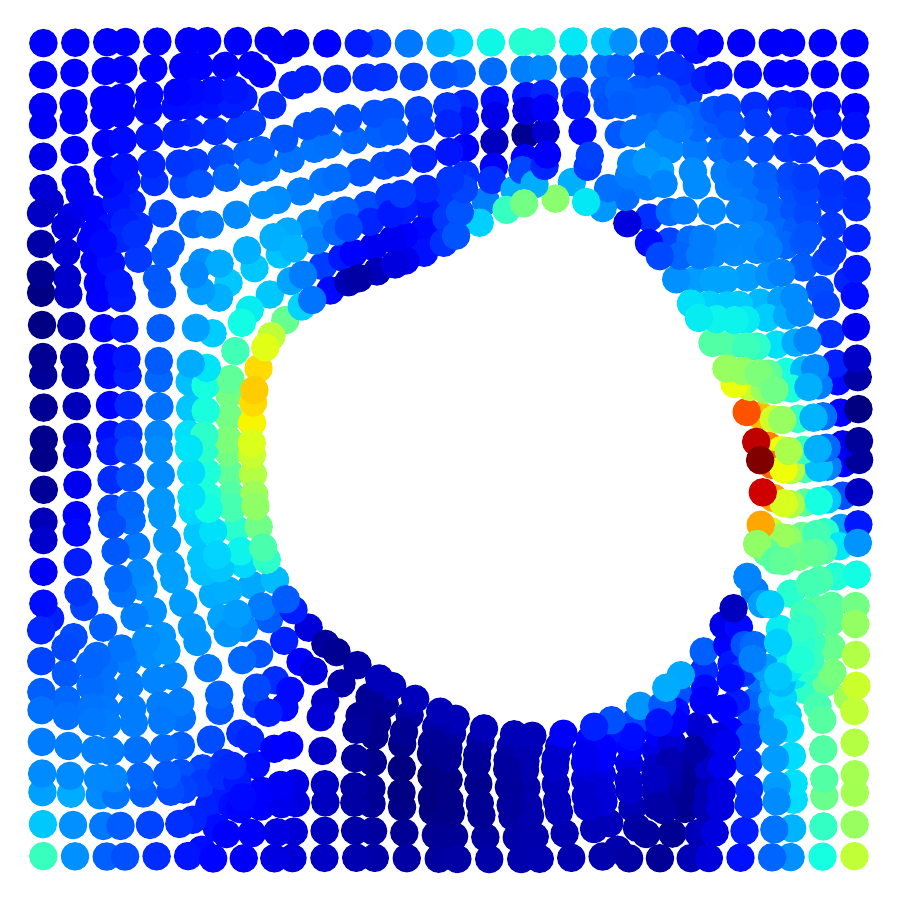}
\end{subfigure}
\begin{subfigure}[c]{.05\textwidth}
	\centering
	\raisebox{0mm}{\includegraphics{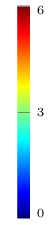}}
%	\raisebox{0mm}{\colorbard{0}{3.65}{3.6}}
\end{subfigure}
\begin{subfigure}[c]{.25\textwidth}
	\centering
	\includegraphics[width=.9\linewidth]{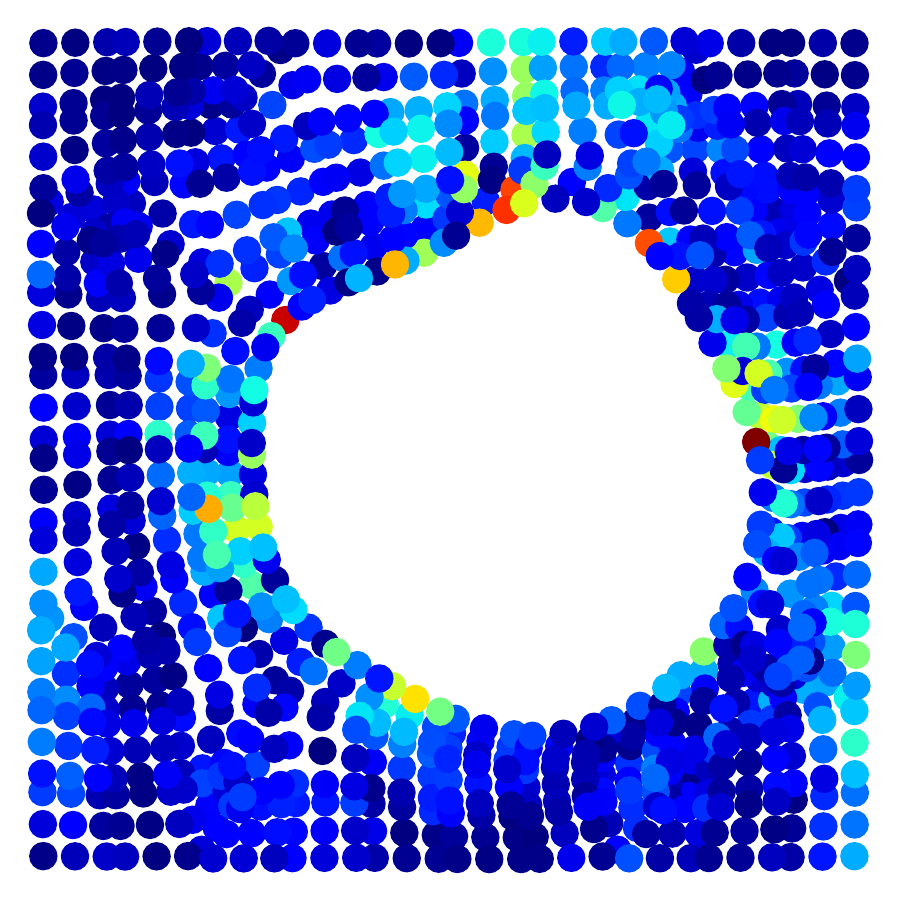}
\end{subfigure}

\smallskip

\begin{subfigure}[c]{.05\textwidth}
	\centering
	\raisebox{0mm}{\includegraphics{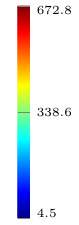}}
%	\raisebox{0mm}{\colorbard{1.94}{937}{3.6}}
\end{subfigure}\quad
\begin{subfigure}[c]{.25\textwidth}
	\centering
	\includegraphics[width=.9\linewidth]{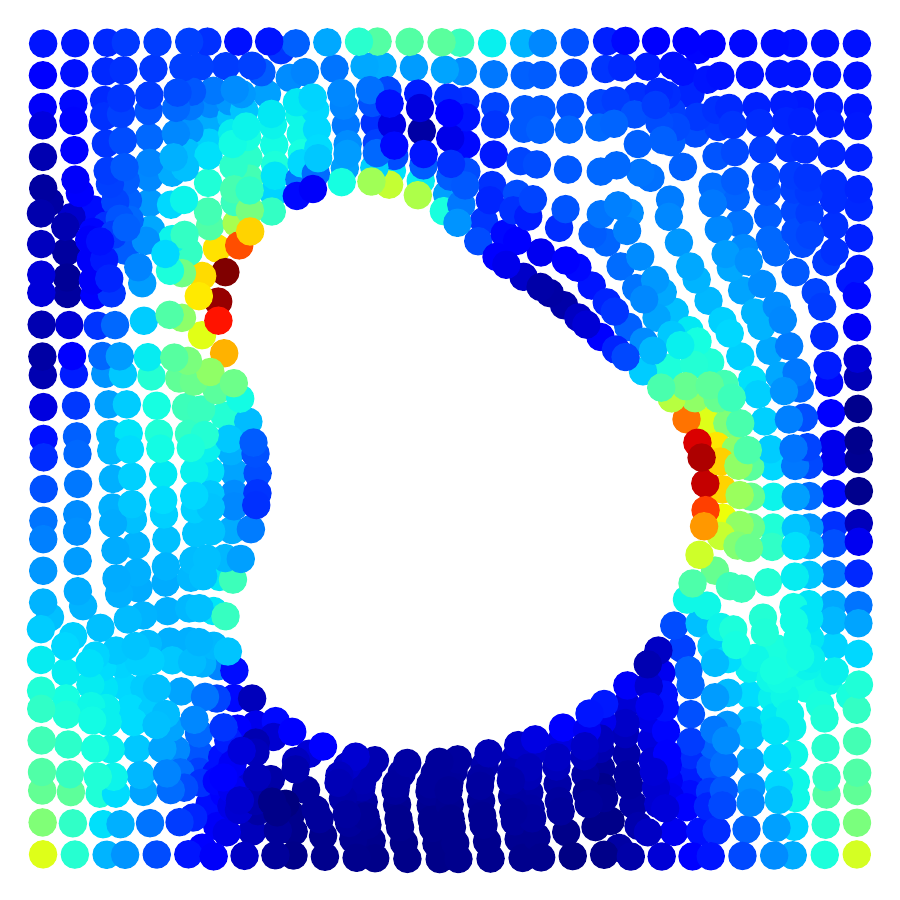}
\end{subfigure}
\begin{subfigure}[c]{.25\textwidth}
	\centering
	\includegraphics[width=.9\linewidth]{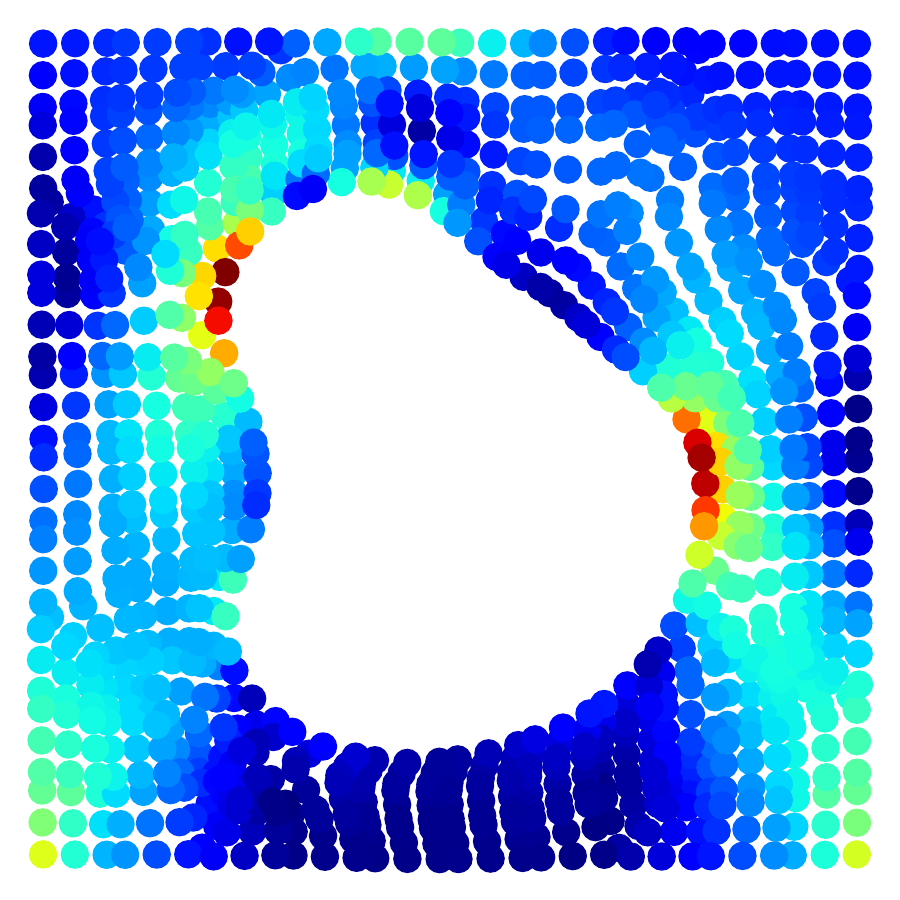}
\end{subfigure}
\begin{subfigure}[c]{.05\textwidth}
	\centering
	\raisebox{0mm}{\includegraphics{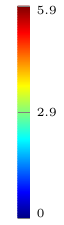}}
%	\raisebox{0mm}{\colorbard{0}{3.65}{3.6}}
\end{subfigure}
\begin{subfigure}[c]{.25\textwidth}
	\centering
	\includegraphics[width=.9\linewidth]{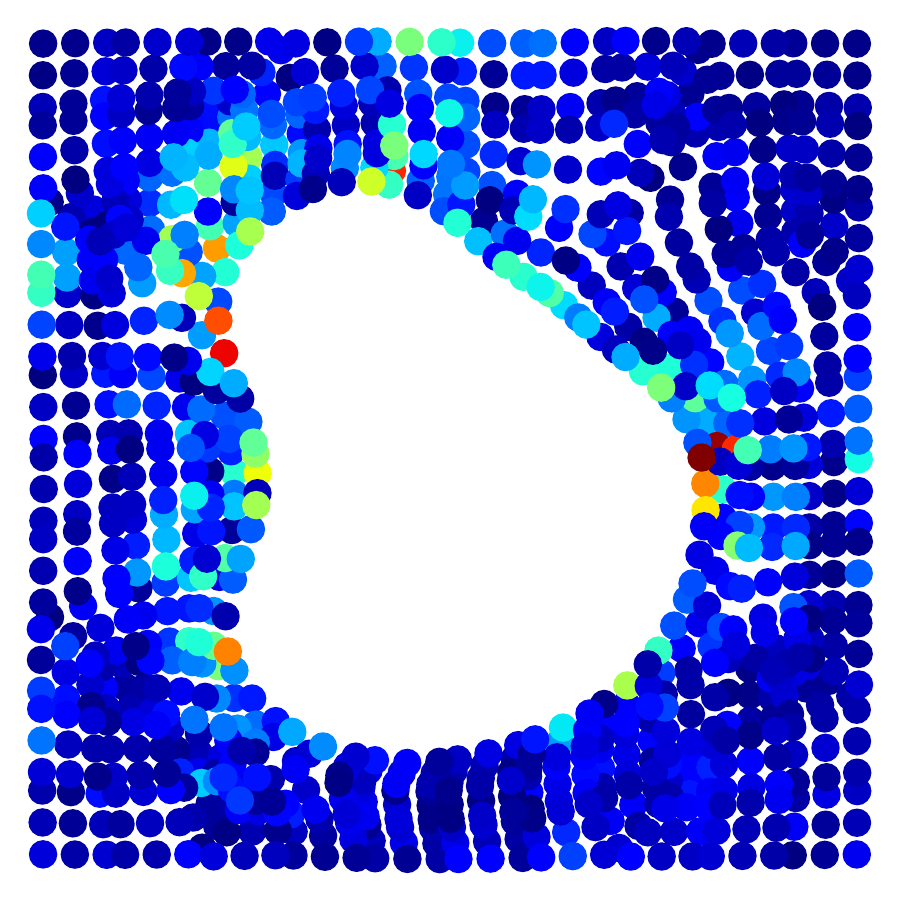}
\end{subfigure}
\caption{Stress field for elasticity. Left: reference. Middle: prediction by PiT. Right: absolute error.}
\label{fig:elasticity}
\end{figure}

The Elasticity benchmark presents a unique challenge as it comprises samples with cavities of varying shapes. Unlike other tasks that use a fixed grid for all data samples, the sampling points in the Elasticity dataset are body-fitted to the cavities. This feature makes Elasticity distinct. As further demonstrated in Figure \ref{fig:elasticity}, PiT effectively captures the stress concentration resulting from the irregular geometries of the cavities.

We have found that the parametrization of the cavity's shape is crucial for learning the target operator. Each sample in the dataset is characterized by a 42-dimensional vector, which is utilized to parametrize the shape of the cavity. 
We concatenate this vector with the mesh point coordinates to serve as the input for our model. In other words, the input is a tensor with 44 channels, of which 42 are constants across the domain. 
To improve the performance,  we apply the transformation $g(r)=5r-1$ to each channel of the shape parameters, effectively normalizing their values to the interval  $[0,1]$.  This normalization ensures that the shape features are comparable to the coordinate features, considering that the problem is defined within the unit square  $[0,1]^2$.

%%%%%%%%%%%%%%%%%%%%%%%%%%%%%%%%%%%%%%%%%%%%

\subsection{Additional Experimental Results on Benchmark 6: NACA}
The data for the NACA benchmark are sampled on C-grid meshes, which are locally refined near the surfaces of the airfoils. The mesh points cover a large domain encompassing the airfoil, where the chord length is set to $1$. 

\begin{figure}[ht!]
\centering
\begin{subfigure}[c]{.05\textwidth}
	\centering
	\raisebox{0mm}{\includegraphics{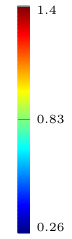}}
%	\raisebox{0mm}{\colorbard{0.25739902}{1.3959179}{3.85}}
\end{subfigure}\quad
\begin{subfigure}[c]{.25\textwidth}
	\centering
	\includegraphics[width=.9\linewidth]{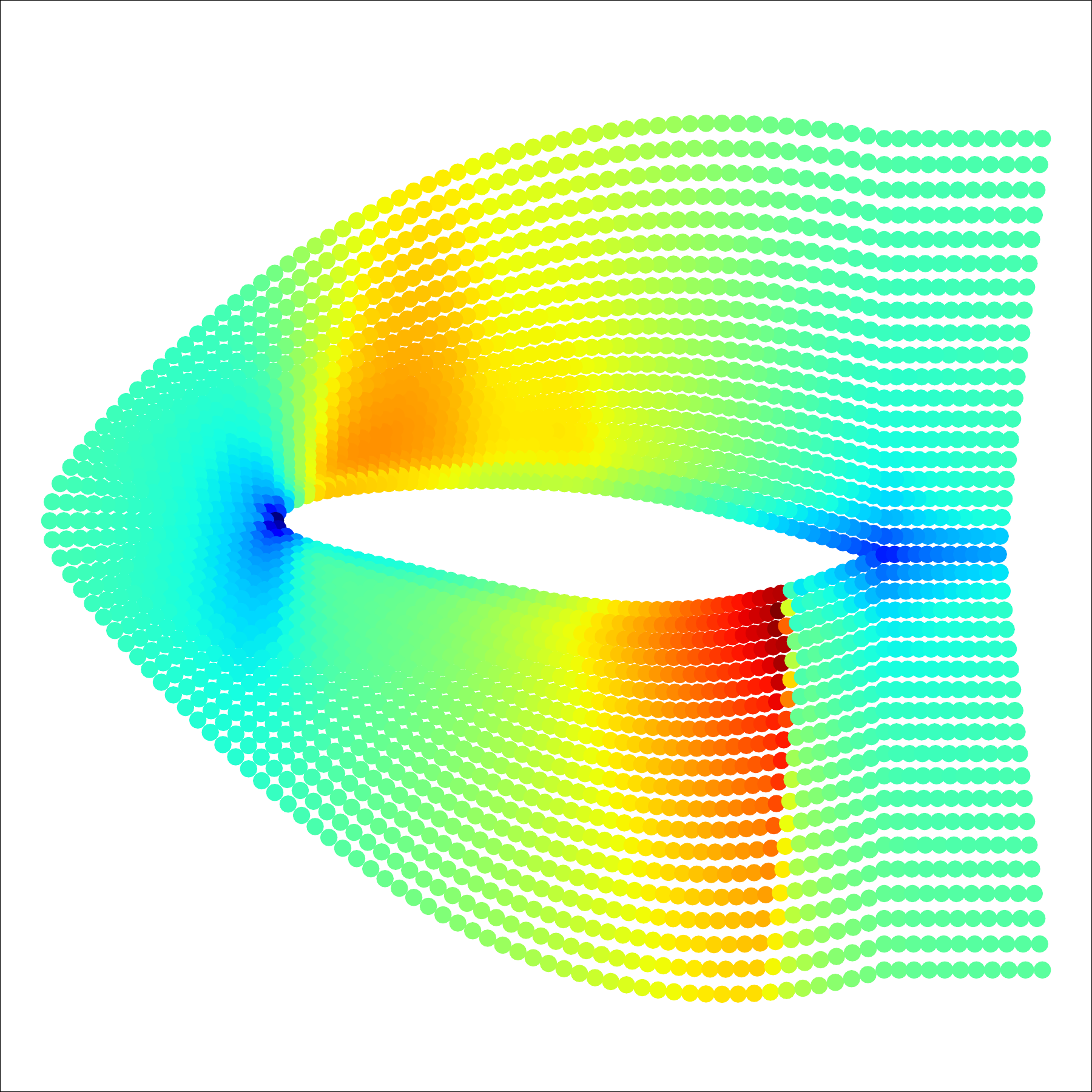}
\end{subfigure}
\begin{subfigure}[c]{.25\textwidth}
	\centering
	\includegraphics[width=.9\linewidth]{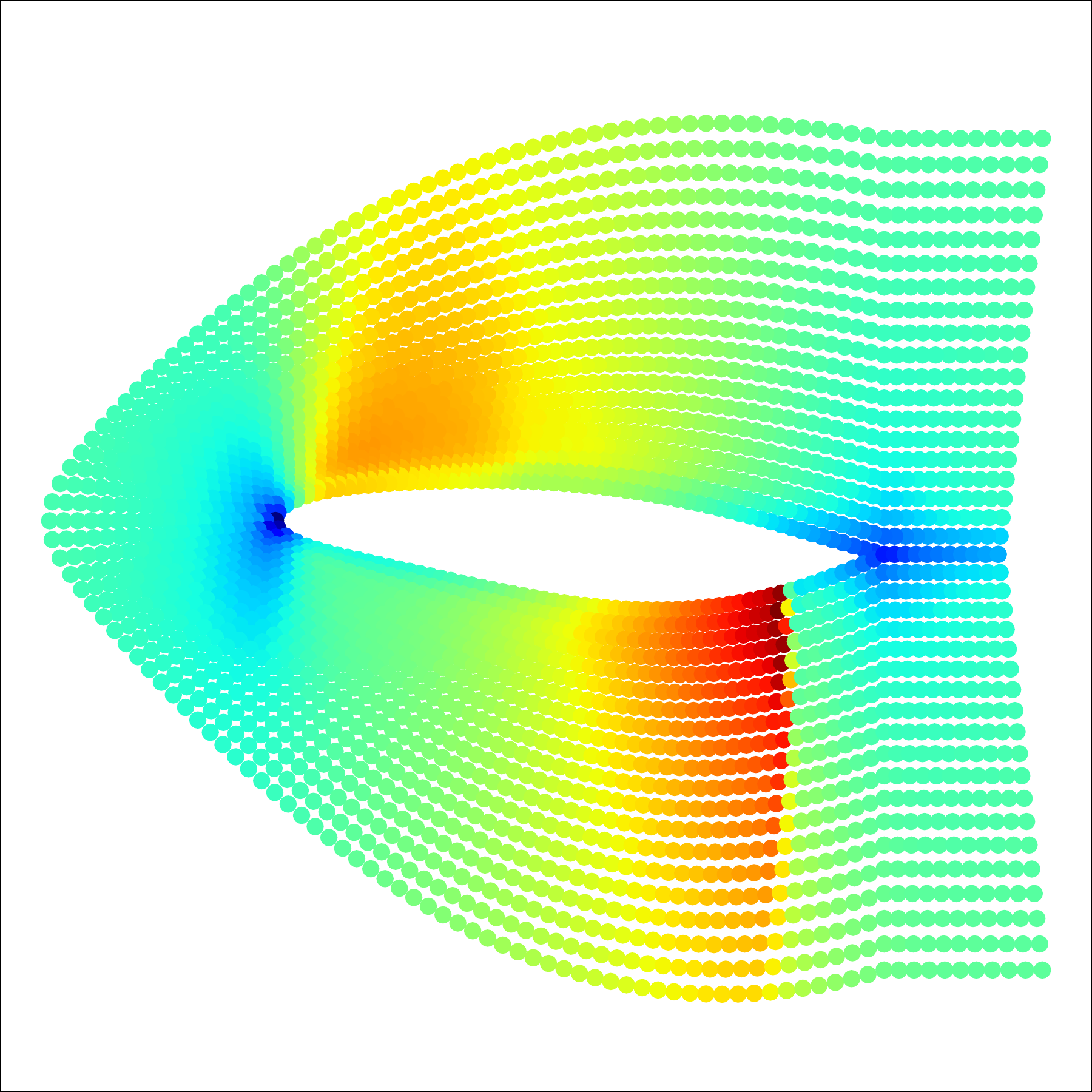}
\end{subfigure}
\begin{subfigure}[c]{.05\textwidth}
	\centering
	\raisebox{0mm}{\includegraphics{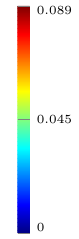}}
%	\raisebox{0mm}{\colorbard{0}{0.09376395}{3.85}}
\end{subfigure}\;
\begin{subfigure}[c]{.25\textwidth}
	\centering
	\includegraphics[width=.9\linewidth]{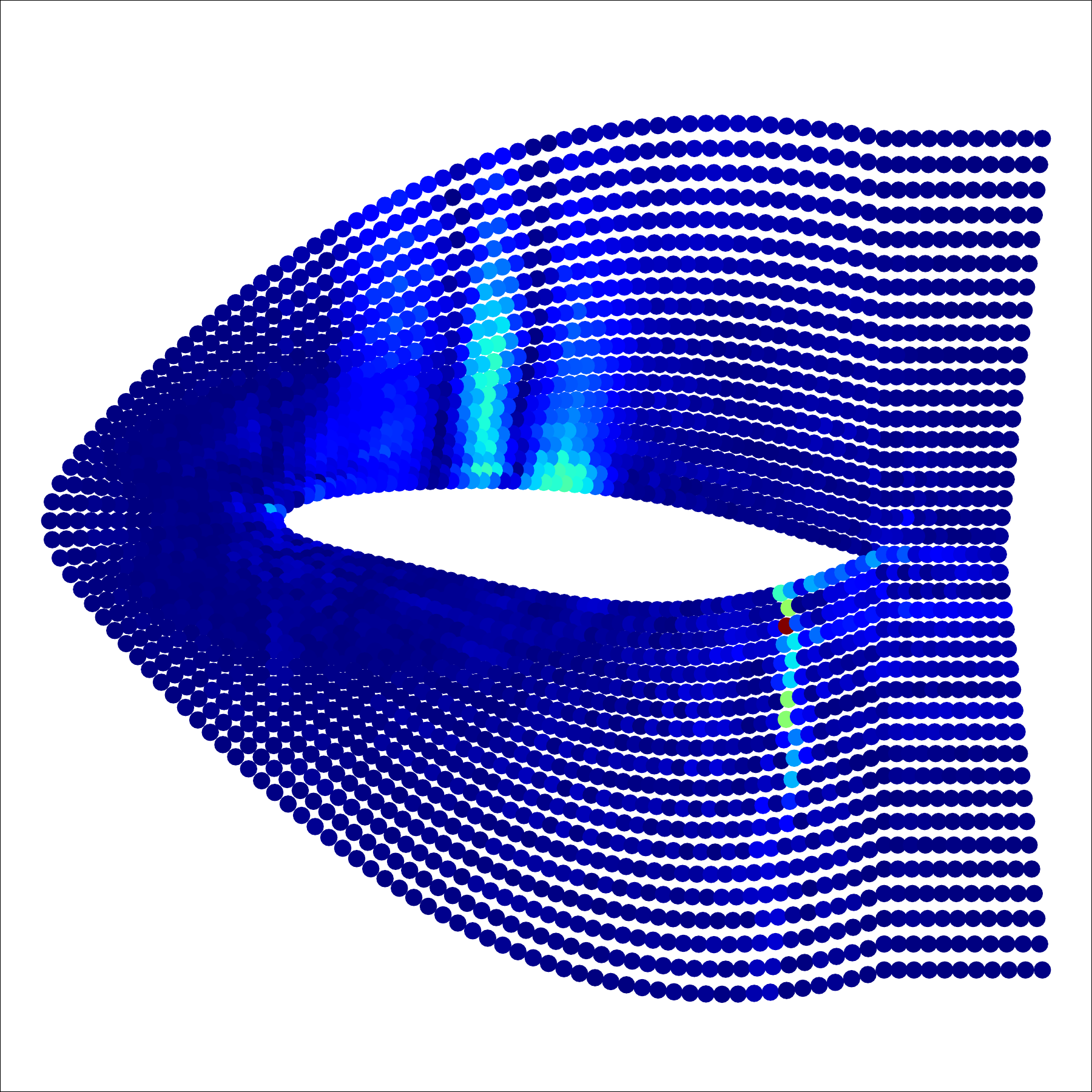}
\end{subfigure}

\smallskip

\begin{subfigure}[c]{.05\textwidth}
	\centering
	\raisebox{0mm}{\includegraphics{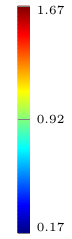}}
%	\raisebox{0mm}{\colorbard{0.25739902}{1.3959179}{3.85}}
\end{subfigure}\quad
\begin{subfigure}[c]{.25\textwidth}
	\centering
	\includegraphics[width=.9\linewidth]{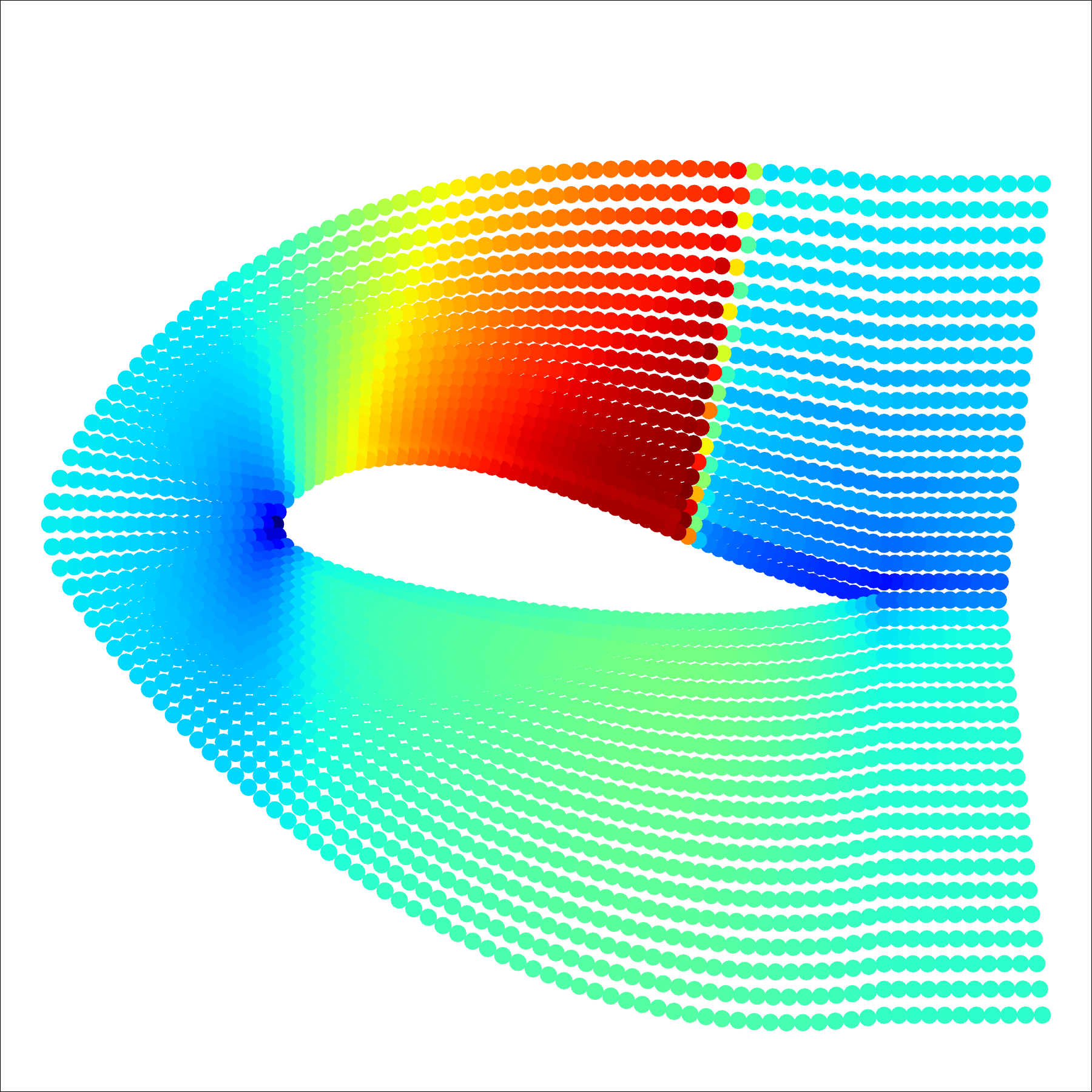}
\end{subfigure}
\begin{subfigure}[c]{.25\textwidth}
	\centering
	\includegraphics[width=.9\linewidth]{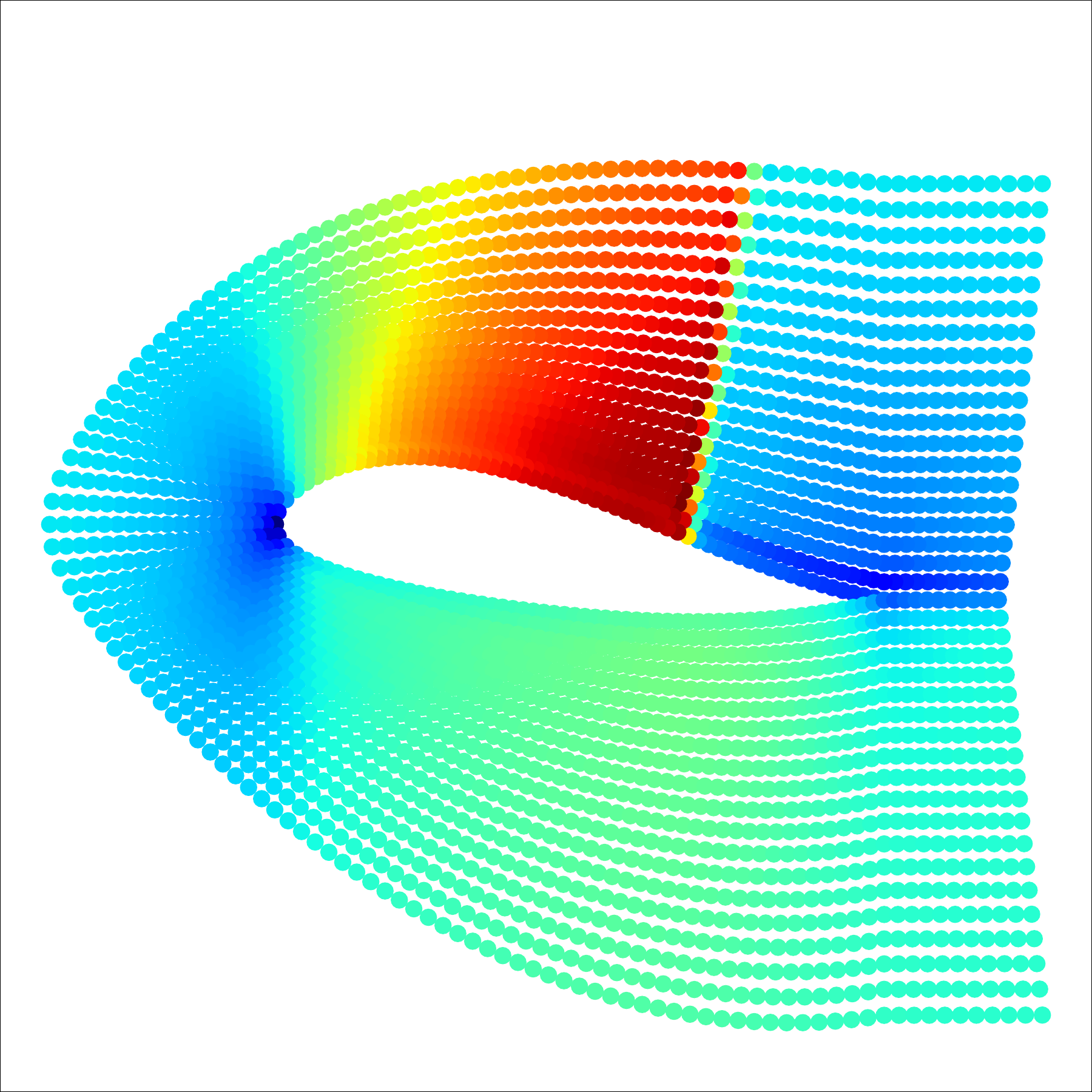}
\end{subfigure}
\begin{subfigure}[c]{.05\textwidth}
	\centering
	\raisebox{0mm}{\includegraphics{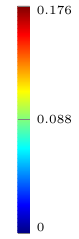}}
%	\raisebox{0mm}{\colorbard{0}{0.09376395}{3.85}}
\end{subfigure}\;
\begin{subfigure}[c]{.25\textwidth}
	\centering
	\includegraphics[width=.9\linewidth]{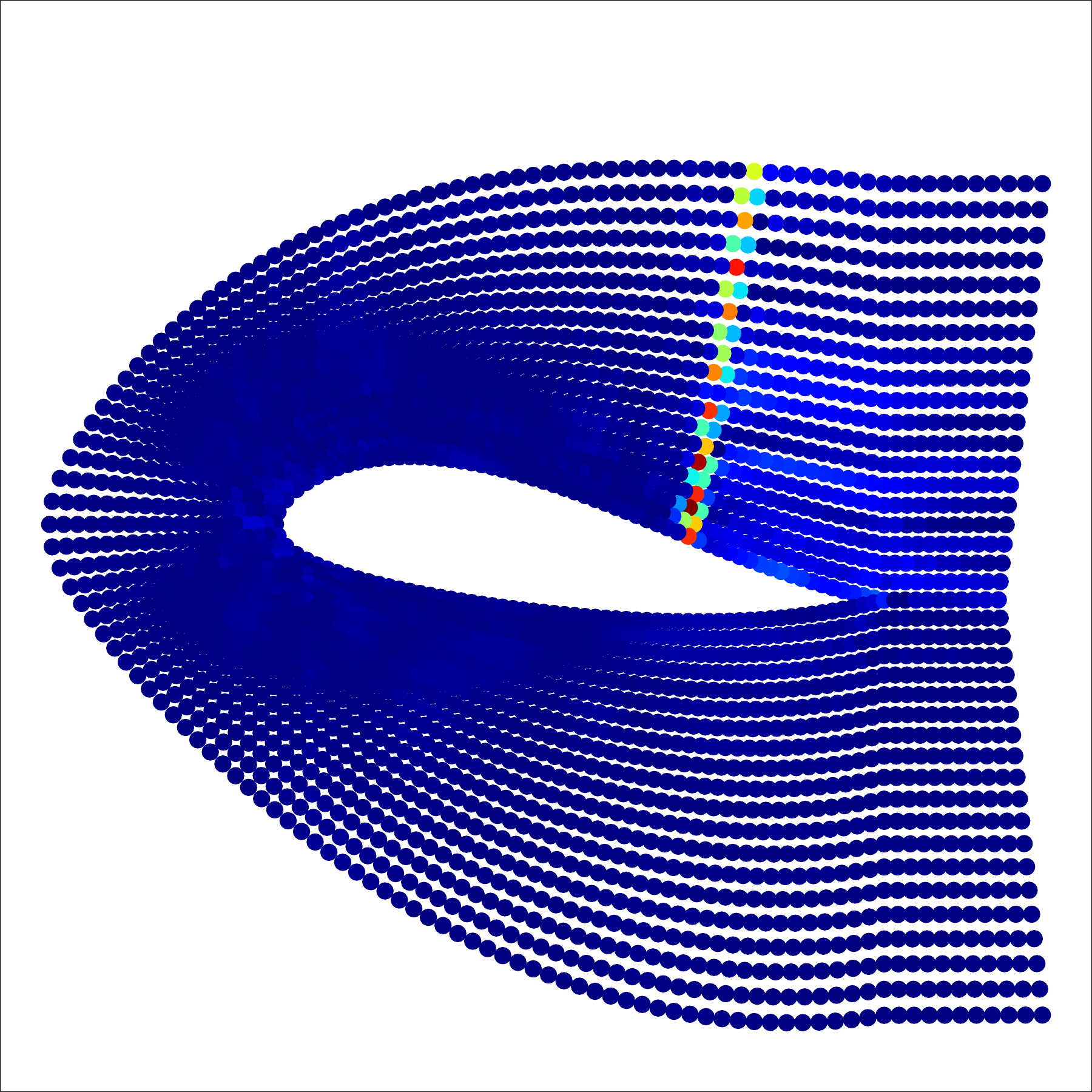}
\end{subfigure}

\smallskip

\begin{subfigure}[c]{.05\textwidth}
	\centering
	\raisebox{0mm}{\includegraphics{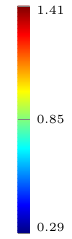}}
%	\raisebox{0mm}{\colorbard{0.25739902}{1.3959179}{3.85}}
\end{subfigure}\quad
\begin{subfigure}[c]{.25\textwidth}
	\centering
	\includegraphics[width=.9\linewidth]{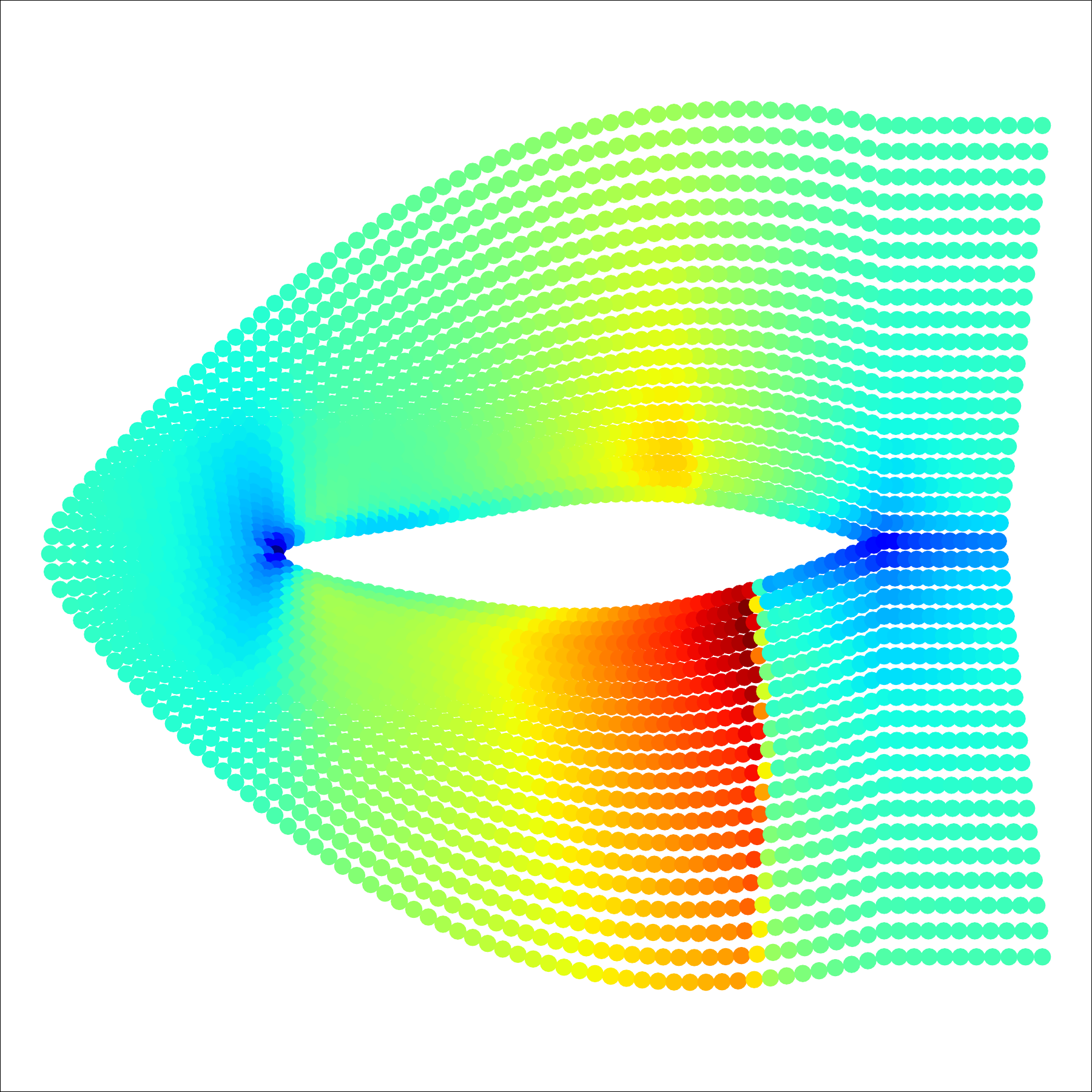}
\end{subfigure}
\begin{subfigure}[c]{.25\textwidth}
	\centering
	\includegraphics[width=.9\linewidth]{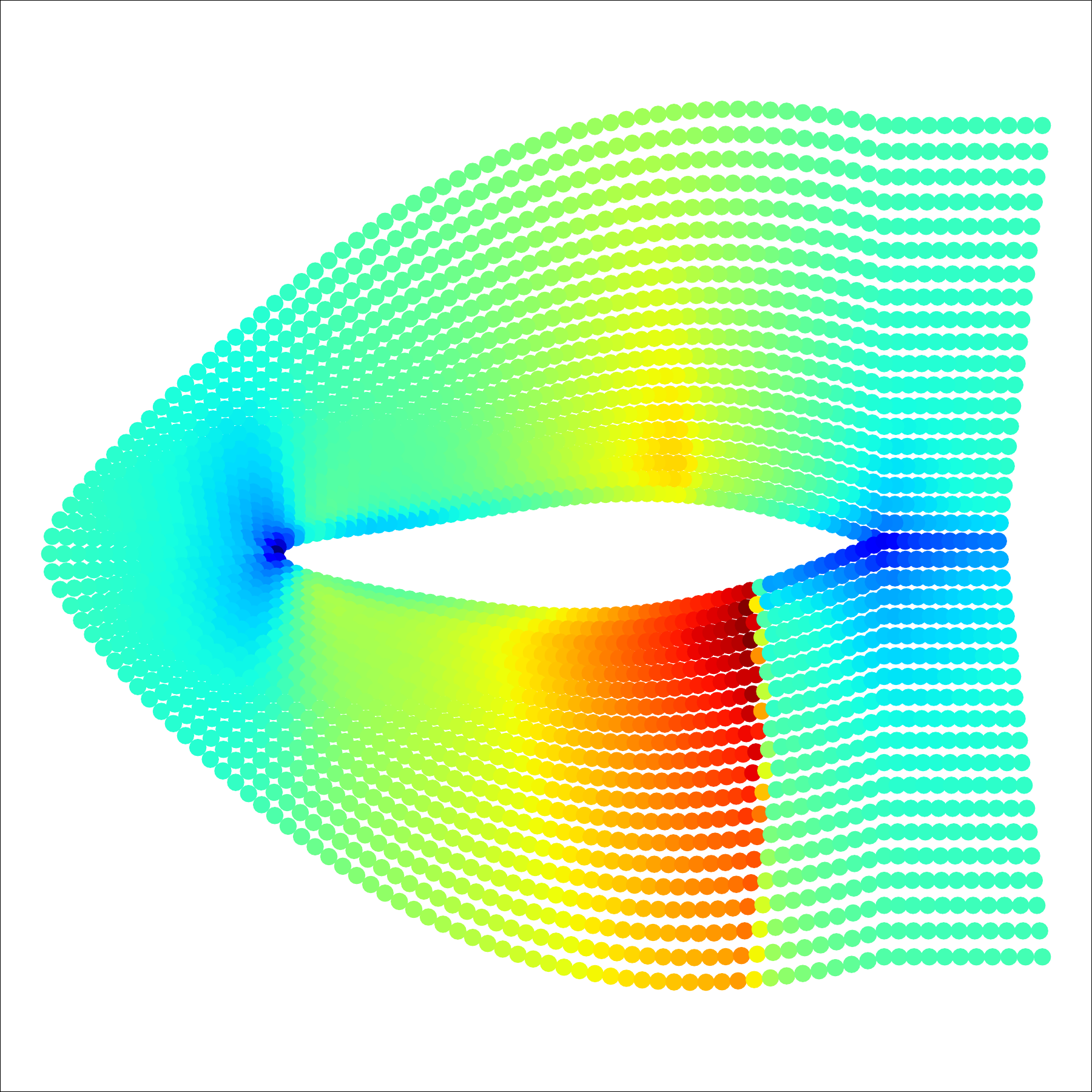}
\end{subfigure}
\begin{subfigure}[c]{.05\textwidth}
	\centering
	\raisebox{0mm}{\includegraphics{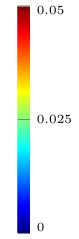}}
%	\raisebox{0mm}{\colorbard{0}{0.09376395}{3.85}}
\end{subfigure}\;
\begin{subfigure}[c]{.25\textwidth}
	\centering
	\includegraphics[width=.9\linewidth]{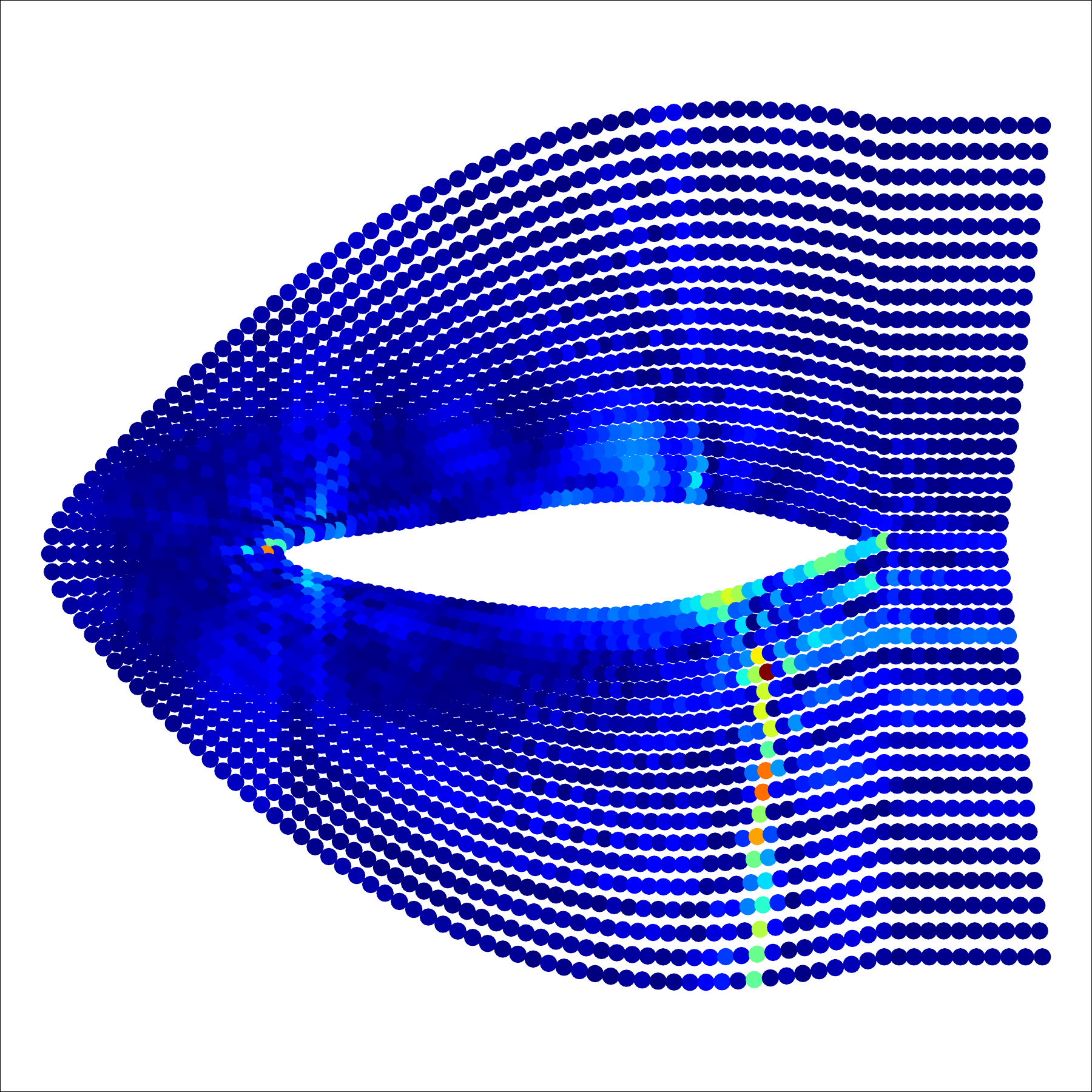}
\end{subfigure}

\smallskip

\begin{subfigure}[c]{.05\textwidth}
	\centering
	\raisebox{0mm}{\includegraphics{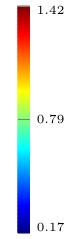}}
%	\raisebox{0mm}{\colorbard{0.25739902}{1.3959179}{3.85}}
\end{subfigure}\quad
\begin{subfigure}[c]{.25\textwidth}
	\centering
	\includegraphics[width=.9\linewidth]{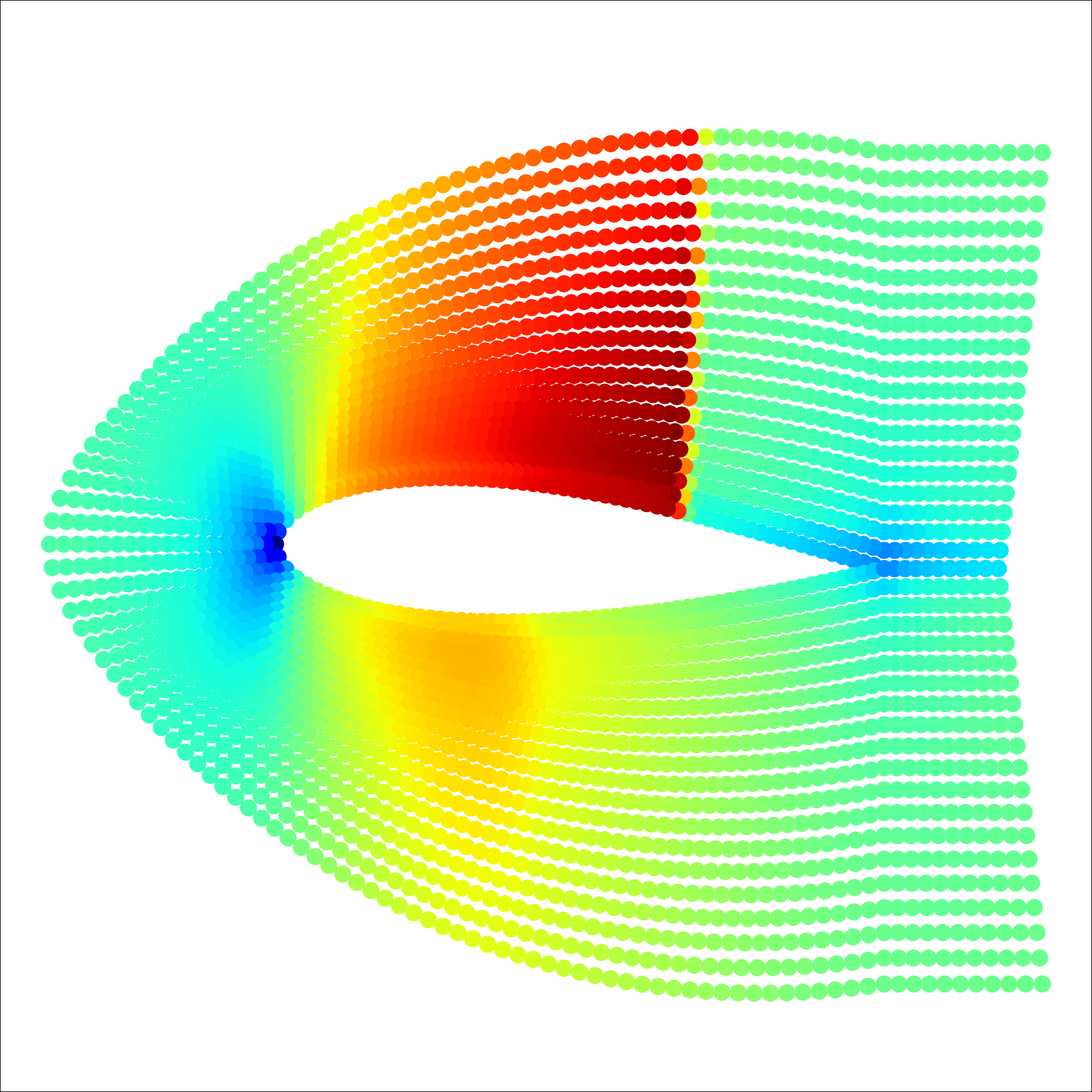}
\end{subfigure}
\begin{subfigure}[c]{.25\textwidth}
	\centering
	\includegraphics[width=.9\linewidth]{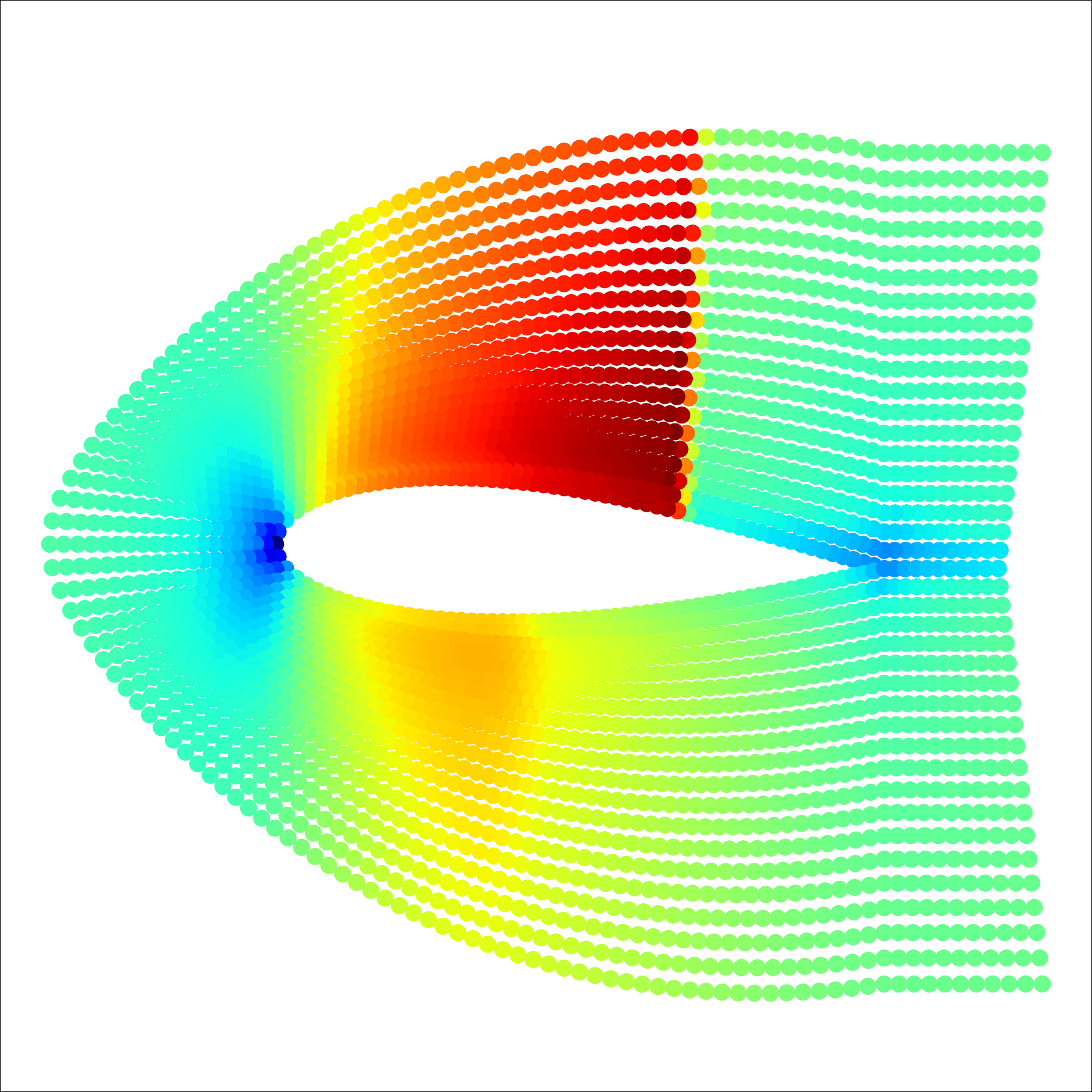}
\end{subfigure}
\begin{subfigure}[c]{.05\textwidth}
	\centering
	\raisebox{0mm}{\includegraphics{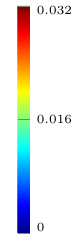}}
%	\raisebox{0mm}{\colorbard{0}{0.09376395}{3.85}}
\end{subfigure}\;
\begin{subfigure}[c]{.25\textwidth}
	\centering
	\includegraphics[width=.9\linewidth]{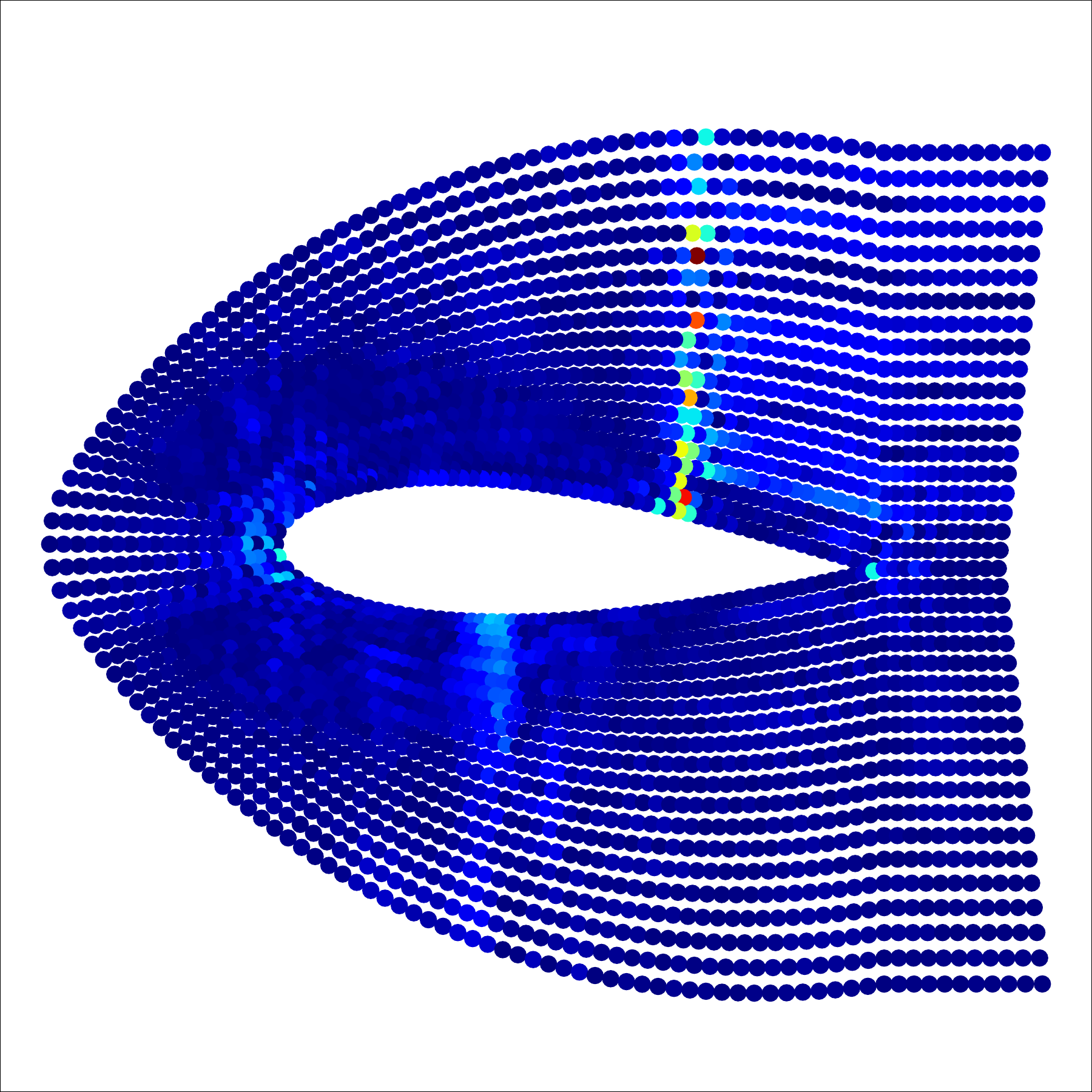}
\end{subfigure}
\caption{Fluid Mach numbers for NACA. Left: reference. Middle: prediction by PiT. Right: absolute error.}
\label{fig:NACA}
\end{figure}

Figure \ref{fig:NACA} displays a close-up view of the Mach field around the airfoils, produced by the learned operator of PiT. One can see that PiT accurately predicts the Mach field and effectively captures the shock wave structures. The predicted solutions are in good agreement with the reference solutions.

%%%%%%%%%%%%%%%%%%%%%%%%%%%%%%%%%%%%%

\end{document}